\documentclass[10pt]{article}

\usepackage[accepted]{tmlr}
\usepackage[T1]{fontenc}
\usepackage[utf8]{inputenc}
\usepackage{microtype}
\usepackage{graphicx}
\usepackage{booktabs}
\usepackage{subcaption}

\usepackage{amsmath,amssymb}
\usepackage{mathtools}
\usepackage{amsthm}
\usepackage{algorithm}
\usepackage{algorithmic}
\usepackage{multirow}
\usepackage[most]{tcolorbox}
\usepackage[framemethod=TikZ]{mdframed}
\usepackage{listings}
\usepackage{enumitem}
\usepackage{array}
\usepackage{colortbl}
\usepackage{url}
\usepackage{hyperref}
\hypersetup{hypertexnames=false}

\providecommand{\Description}[1]{}

\lstdefinestyle{promptjson}{
  basicstyle=\ttfamily\scriptsize,
  breaklines=true,
  breakatwhitespace=false,
  columns=fullflexible,
  keepspaces=true,
  showstringspaces=false
}

\newcommand{\meanstd}[2]{%
  #1\,{\text{\fontsize{4pt}{4pt}\selectfont(\textpm\,#2)}}%
}
\definecolor{litblue}{HTML}{DCEAFF}
\definecolor{sageorange}{HTML}{FFF0D9}
\definecolor{proofgreen}{HTML}{EAF7E8}
\newcommand{\litcell}{\cellcolor{litblue}}
\newcommand{\sagecell}{\cellcolor{sageorange}}

\usepackage[capitalize,noabbrev]{cleveref}

\theoremstyle{plain}
\newtheorem{theorem}{Theorem}[section]
\newtheorem{proposition}[theorem]{Proposition}

\theoremstyle{definition}
\newtheorem{definition}[theorem]{Definition}

\theoremstyle{remark}

\mdfdefinestyle{theorembox}{%
  linewidth=1pt,
  linecolor=blue!50!black,
  backgroundcolor=blue!5,
  roundcorner=4pt,
  skipabove=3pt,
  skipbelow=3pt,
  innertopmargin=3pt,
  innerbottommargin=2pt,
  splittopskip=4pt,
  splitbottomskip=4pt,
  innerleftmargin=8pt,
  innerrightmargin=8pt,
  nobreak=true,
}
\surroundwithmdframed[style=theorembox]{definition}

\title{LLM-Guided Evolution of Symbolic Acceleration Inequality Families for MILP}
\author{\name Milad Yazdani\thanks{Work conducted during an internship at Huawei Technologies Canada.} \email milad.yazdani@ece.ubc.ca \\
  \addr Department of Electrical and Computer Engineering\\
  University of British Columbia, Vancouver, BC, Canada
  \AND
  \name Mahdi Mostajabdaveh\thanks{Corresponding author.} \email mahdi.ms86@gmail.com \\
  \addr Huawei Technologies Canada, Burnaby, BC, Canada
  \AND
  \name Samin Aref \email samin.aref@utoronto.ca \\
  \addr Department of Mechanical and Industrial Engineering\\
  University of Toronto, Toronto, ON, Canada
  \AND
  \name Zirui Zhou \email zirui.zhou@huawei.com \\
  \addr Huawei Technologies Canada, Burnaby, BC, Canada}

\def\month{10} %
\def\year{2026}
\def\openreview{\url{https://openreview.net/forum?id=rxJvINT074}}

\begin{document}
\maketitle
\let\AND\relax
\begin{abstract}
Integer programming (IP) is central to many combinatorial optimization tasks but remains challenging due to its NP-hard nature. A practical way to improve IP solvers is to manually design inequality families that exploit problem structure. However, this creative process requires deep expertise and has therefore been difficult to automate. Our proposed framework, Symbolic Acceleration via Generative Evolution (\textsc{SAGE}), automates symbolic discovery of acceleration inequality families at the modeling level. It reasons over a symbolic IP formulation and a natural language description of the problem to discover a reusable set of acceleration inequality families that can be instantiated for each problem instance. These acceleration inequality families are selected through empirical admissibility and impact screening rather than mathematical proofs. \textsc{SAGE} (i) initializes a population of candidate acceleration inequality families via an initializer agent that uses an LLM; (ii) it screens candidates for admissibility on a small development admissibility set for solution feasibility and the separation of the relaxation solution; and (iii) it iteratively refines the population of inequalities through evolutionary crossover and mutation agents. Across seven MILP benchmarks, \textsc{SAGE} reduces optimality gaps by up to \(62.1\%\) relative to baseline MILP formulations solved with a fixed time budget and reaches target gaps up to \(5.51\times\) faster according to shifted geometric mean speedup. Comprehensive ablations show its robustness across different LLM backends and across solvers and internal cut settings.
\href{https://github.com/milad1378yz/sage}{Code}

\end{abstract}

\begin{figure}[htbp]
    \centering
    \includegraphics[width=0.82\textwidth]{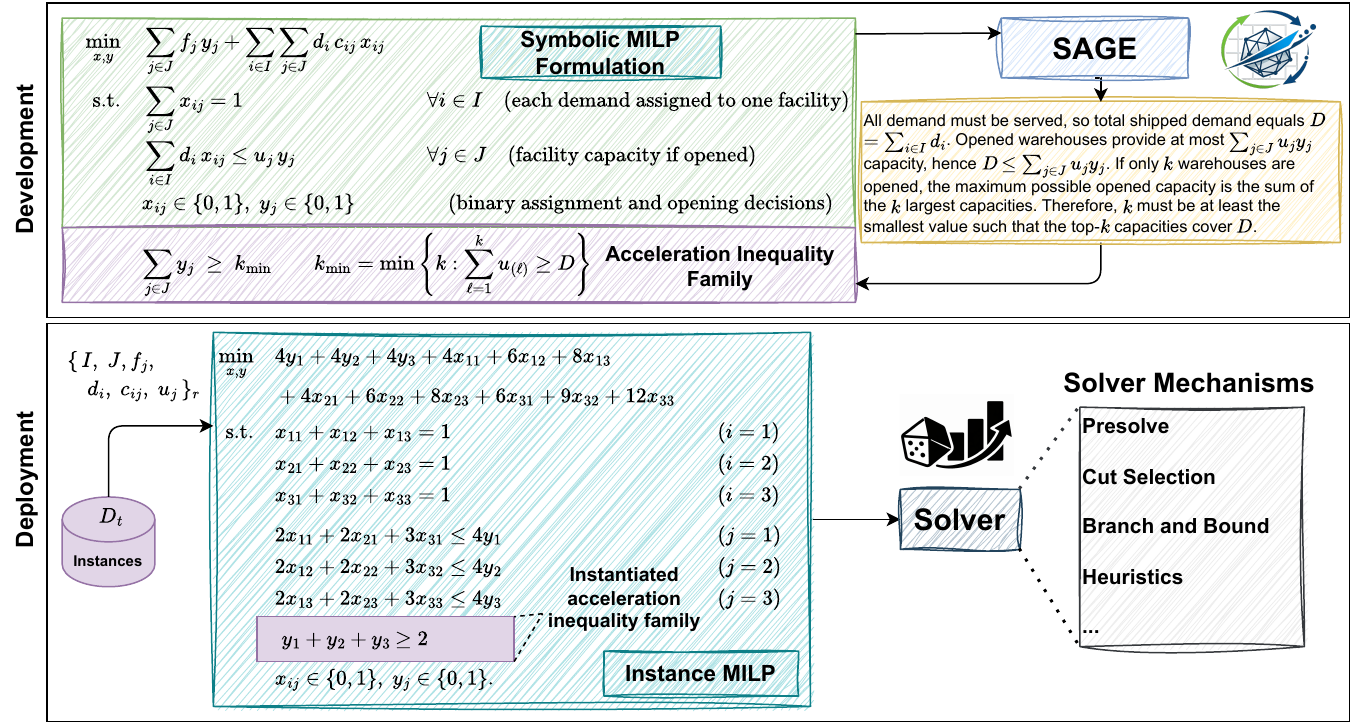}
    \caption{View of \textsc{SAGE} in two stages: during development, \textsc{SAGE} learns a reusable set of acceleration inequality families for a symbolic MILP formulation using a natural language description. During deployment, the families are instantiated for each problem instance and appended to its instantiated MILP before solving.}
    \Description{A two-stage workflow showing development of reusable inequality families and deployment on instantiated MILPs.}
    \label{fig:evocut_symbolic_overview}
\end{figure}

\section{Introduction}
Operations Research (OR) workflows usually consist of two main phases: formulating an optimization model and solving it. The first phase involves understanding the problem, identifying decision variables, objectives, and constraints, and building a mathematical formulation. The second phase uses numerical algorithms and generic optimization solvers to find exact or approximate solutions. As problems become more complex in practice, using artificial intelligence (AI) to improve OR has become increasingly important. Recent advances have used large language models (LLMs) to automate mathematical modeling, converting natural language descriptions directly into mathematical formulations~\cite{ramamonjison2023nl4opt,jiang2025llmopt,huang2025orlm,mostajabdaveh2025evaluating}. While there have been promising results in generating correct models, improving the efficiency of those models remains challenging. Parallel efforts have applied AI to the solving phase, including AI for heuristic design~\cite{liu2024evolution,ye2024reevo} and machine learning techniques for accelerating generic optimization solvers~\cite{alvarez2017machine,li2023learning}. However, the crucial feedback loop between how a problem is formulated and how efficiently it can be solved remains largely unexplored. Optimization problems can often be formulated in multiple ways, with solver performance varying substantially depending on the formulation and its interaction with the solver~\cite{klotz2024converting}.

\noindent In this study, we focus on static model augmentation before calling the solver for Mixed Integer Linear Programming (MILP). Starting from a base MILP formulation, we generate additional linear inequality families, called acceleration inequality families, intended to improve solver efficiency. Many widely studied problem classes admit problem-specific inequalities derived in the OR literature. However, we aim for a general pipeline that operates directly on a symbolic MILP formulation shared across instances. We refer to this shared formulation as the symbolic MILP formulation and to its specialization with instance data as the instantiated MILP. Our proposed method, Symbolic Acceleration via Generative Evolution (\textsc{SAGE}), uses LLM agents guided by evolutionary search to discover and refine such families by reasoning over a symbolic MILP formulation and a natural language description of the problem. In particular, \textsc{SAGE} learns a reusable set of acceleration inequality families during development and instantiates them for each problem instance at deployment time (Fig.~\ref{fig:evocut_symbolic_overview}). \textsc{SAGE} is a general framework for symbolic discovery at the modeling level and integrates into standard solver pipelines across different MILP models.

\noindent Other methods also augment or transform a MILP: manual formulation strengthening and solver presolve act before branch-and-bound begins, and solver cuts and learned runtime policies act during it. Table~\ref{tab:mechanisms} separates \textsc{SAGE} from these mechanisms on five axes: the stage at which each acts, whether it uses problem-specific structure, whether it operates on the symbolic MILP formulation or on the instantiated MILP, whether it is manual or automated, and what it returns. Formulation strengthening uses problem-specific structure at the level of the symbolic formulation, is derived manually by an expert, and returns a tighter equivalent formulation of the problem class, established through a proof of validity or of equivalence~\mbox{\cite{vielma2015mixed,klotz2024converting}}. \textsc{SAGE} works at the same level and with the same kind of structure, but derives a reusable acceleration inequality family automatically, screens it empirically, and instantiates it for each instance of the class. Solver presolve applies generic, correctness-preserving reductions, such as bound tightening and probing, to the instantiated MILP inside the solver and returns a reduced model for that one instance~\mbox{\cite{savelsbergh1994preprocessing,achterberg2020presolve}}. Solver cuts and learned runtime policies also act inside the solver, during branch-and-bound, on the instantiated MILP (Section~\ref{sec:related_work}). \textsc{SAGE} augments the symbolic formulation before submission, so presolve and the runtime mechanisms run on the augmented instantiated MILP, and the mechanisms are complementary.
\begin{table}[t]
  \centering
  \footnotesize
  \caption{\textsc{SAGE} compared with other mechanisms that augment or transform a MILP: the stage at which each acts, whether it uses problem-specific structure, what it operates on, whether it is manual or automated, and what it returns.}
  \label{tab:mechanisms}
  \hypersetup{hidelinks}
  \setlength{\tabcolsep}{2pt}
  \setlength{\aboverulesep}{4pt}
  \setlength{\belowrulesep}{4pt}
  \renewcommand{\arraystretch}{1.1}
  \begin{tabular}{@{}
    >{\raggedright\arraybackslash}m{\dimexpr0.22\linewidth-\tabcolsep\relax}
    >{\raggedright\arraybackslash}m{\dimexpr0.154\linewidth-2\tabcolsep\relax}
    >{\raggedright\arraybackslash}m{\dimexpr0.159\linewidth-2\tabcolsep\relax}
    >{\raggedright\arraybackslash}m{\dimexpr0.134\linewidth-2\tabcolsep\relax}
    >{\raggedright\arraybackslash}m{\dimexpr0.145\linewidth-2\tabcolsep\relax}
    >{\raggedright\arraybackslash}m{\dimexpr0.188\linewidth-\tabcolsep\relax}@{}}
    \toprule
    \textbf{Mechanism} & \textbf{Stage} & \textbf{Problem-specific structure} & \textbf{Operates on} & \textbf{Derivation} & \textbf{Output} \\
    \midrule
    \textbf{\textsc{SAGE}} & Model design,\newline before the solver\newline is called & Yes, for a\newline problem class & Symbolic MILP\newline formulation & Automated, by LLM agents with evolutionary search & A reusable acceleration inequality family, instantiated for each instance \\
    \midrule[0.25pt]
    \textbf{Formulation\newline strengthening}\par\vspace{2pt}\citep{vielma2015mixed,klotz2024converting} & Model design,\newline before the solver\newline is called & Yes, for a\newline problem class & Symbolic MILP\newline formulation & Manual, by an expert & A tighter equivalent formulation of the problem class \\
    \midrule[0.25pt]
    \textbf{Solver presolve}\par\vspace{2pt}\citep{savelsbergh1994preprocessing,achterberg2020presolve} & Inside the solver,\newline before\newline branch-and-bound & No, generic\newline reductions & Instantiated\newline MILP & Automated, by solver routines & A reduced model of that one instance \\
    \midrule[0.25pt]
    \textbf{Solver cuts and learned\newline runtime policies}\par\vspace{2pt}(\citealp{turner2023adaptive};\newline\citealp{gasse2019exact};\newline\citealp{li2023learning}) & Inside the solver,\newline during\newline branch-and-bound & No, generic cuts or learned policies & Instantiated\newline MILP & Automated, by solver routines or trained policies & Cuts and search decisions during the solve \\
    \bottomrule
  \end{tabular}
\end{table}

\noindent Together, this paper makes three contributions. First, we introduce \textsc{SAGE}, a general framework for \emph{symbolic discovery of empirically screened acceleration inequality families at the modeling level}. Given a symbolic MILP formulation and a small admissibility set, \textsc{SAGE} discovers and selects a \emph{reusable} set of acceleration inequality families that can be instantiated for each problem instance. Second, we evaluate \textsc{SAGE} on seven diverse MILP problems and show consistent solver speedups on unseen instances. The largest row-average checkpoint gap reduction is \(62.1\%\), and the largest row-average target-gap speedup is \(5.51\times\). On the four benchmarks with published valid inequalities, the discovered families match or exceed the mean speedups from those inequalities under the same protocol (Section~\ref{sec:literature_cut_comparison}). Third, we provide an ablation and robustness analysis of \textsc{SAGE}: (a) data efficiency by varying the sizes of the impact-assessment and admissibility sets (Section~\ref{sec:sensitivity_training_size}), (b) LLM sensitivity across DeepSeek-R1, Gemini 3 Pro, and GPT-5.1 (Table~\ref{tab:post_ga_results}), (c) solver robustness by transferring the same discovered acceleration inequality families across solvers (Section~\ref{sec:cross_solver_transfer}), and (d) interaction with solver internal cut generation by comparing Gurobi with internal cuts disabled vs.\ default settings (Section~\ref{sec:internal_cut_interaction}).

\section{Related Work}
\label{sec:related_work}

Recent work has explored using LLMs to automate various stages of OR workflows, from problem understanding to algorithm generation~\cite{liu2026systematic,huang2024when}. A central line of research focuses on converting natural language descriptions into optimization formulations, initiated by the NL4Opt competition~\cite{ramamonjison2023nl4opt} and advanced by systems such as OptiMUS~\cite{ahmaditeshnizi2024optimus}, ORLM~\cite{huang2025orlm}, and LM4Opt~\cite{ahmed2024lm4opt}. Recent frameworks with multiple agents incorporate dedicated verification agents to ensure correctness before generating code that can be passed directly to a solver, including the chain of experts~\cite{xiao2023chain} and the staged agent architecture of~\cite{mostajabdaveh2024optimization}. These efforts focus on modeling fidelity. In contrast, \textsc{SAGE} targets efficiency after a base formulation already exists: it proposes additional acceleration inequality families, screens them empirically on held-out instances, and ranks them by measured effect on time-limited solver progress. To our knowledge, prior research has not studied the automatic discovery of \emph{reusable} acceleration inequality families that augment the symbolic model before the solver is called. Another emerging direction uses LLMs within evolutionary frameworks to generate heuristics for combinatorial optimization. These methods aim to boost heuristic algorithm performance by modifying parts of the algorithm code (e.g., heuristic rules)~\cite{liu2024evolution,ye2024reevo}. FunSearch~\cite{romera2024mathematical} couples a frozen LLM with evolutionary search and an evaluator to discover heuristics, while the evolution of heuristics~\cite{liu2024evolution} refines natural language "thoughts" and executable code using evolutionary strategies. ReEvo~\cite{ye2024reevo} introduces reflective evolution with pairwise comparisons and reflections over longer horizons to improve code generated by LLMs iteratively. While these works evolve heuristics or code, \textsc{SAGE} evolves reusable acceleration inequality families. More broadly, machine learning for MILP solving has targeted decisions \emph{inside} branch-and-bound: learned branching approximates strong branching or variable-selection policies~\cite{alvarez2017machine,khalil2016learning,gasse2019exact}; learned cut selection chooses which generated cuts to apply~\cite{tang2020reinforcement,turner2023adaptive,paulus2022learning,wang2023learning}; separator-configuration methods such as L2Sep, cold-start LLM configuration, and dynamic separator configuration decide which solver separators to activate during the solve~\cite{li2023learning,lawless2025llms,ye2025dynamic}; learned or optimized cut-generation mechanisms select cut-generating functions~\cite{chetelat2023continuous,cheng2024learning}; and learned primal heuristics aim to find incumbents early~\cite{shen2021learning}. The central distinction is that these methods modify runtime policies within the solver's branch-and-bound process, whereas \textsc{SAGE} operates one level earlier: it augments the symbolic MILP formulation before the solver is called, producing reusable acceleration inequality families that can then be solved with any branching, cut-selection, separator-configuration, cut-generation, or primal-heuristic policy. Thus our contribution is not a replacement for ML solver policies but an orthogonal modeling-level layer; the same learned runtime policy can be run on the baseline formulation or on the \textsc{SAGE}-augmented formulation.

\section{Preliminaries}
To motivate an automatic yet reliable generator of acceleration inequality families, we first recall the baseline MILP and define the acceleration inequality families that \textsc{SAGE} discovers. We then relate them to the linear relaxation and to the classical notions of valid and optimality-preserving inequalities. Our goal is to discover useful acceleration inequality families, including valid inequalities when possible, and to retain them through empirical screening and measured solver progress.

\noindent We consider an MILP of the form
{\small
\begin{equation}
    \max\bigl\{\, c^\top x + h^\top y \,\colon\; A x + G y \le b,\; x \in \mathbb{Z}_{\ge 0}^n,\; y \in \mathbb{R}_{\ge 0}^p \bigr\},
    \label{eq:MILP}
\end{equation}
}
with integer variables \(x\in\mathbb{Z}_{\ge 0}^n\) and continuous variables \(y\in\mathbb{R}_{\ge 0}^p\). Its feasible set is
\[
S \;=\; \{\,(x,y) \in \mathbb{Z}_{\ge 0}^n \times \mathbb{R}_{\ge 0}^p \;:\; A x + G y \le b \,\}.
\]

\noindent Throughout the paper we use the term additional inequality family to mean a set of linear inequalities appended to the MILP. We write such a family in the form
\[
  C := \bigl\{\,A_r' x + G_r' y + H_r' z \le \delta_r\,\bigr\}_{r\in R},
\]
where \(R\) is an index set and \(z\) denotes auxiliary variables introduced by the added inequalities. Including \(z\) allows the inequality family to encode additional logical relations. In practice, one may add inequality families only to speed up solving. We call any such added inequality family, tailored to a particular problem class, an acceleration inequality family.
\begin{definition}[Acceleration inequality family]
An additional linear inequality family added to an MILP with the explicit aim of reducing solution time. Such a family need not be composed only of proven valid inequalities or proven optimality-preserving inequalities; in this work, it is retained only after empirical admissibility checks that (i) stored reference solutions remain feasible on a small admissibility set and (ii) at least one stored LP-relaxation solution is separated, with quality assessed by solver progress on an impact-assessment set.
\end{definition}

Relaxing integrality yields the linear programming (LP) relaxation
{\small
\begin{equation}
    \max \bigl\{\,c^\top x + h^\top y \,:\, A x + G y \le b,\, x \in \mathbb{R}_{\ge 0}^n,\, y \in \mathbb{R}_{\ge 0}^p \bigr\},
    \label{eq:LP_relaxation}
\end{equation}
}
\noindent whose feasible set (the linear relaxation set) is
\[
P \,=\, \{\, (x,y) \in \mathbb{R}_{\ge 0}^{n} \times \mathbb{R}_{\ge 0}^{p} \,:\, A x + G y \le b \,\}.
\]
Although Eq.~\eqref{eq:LP_relaxation} can be solved in polynomial time, its optimum is generally fractional (i.e., not in \(S\)). This motivates cutting-plane and branch-and-cut methods, in which a solver adds valid inequalities dynamically during the solve. By contrast, \textsc{SAGE} produces acceleration inequality families at the user or model level that are appended to the MILP before calling the solver.

\noindent Formally, an inequality
\[
    w^\top (x,y) \;\le\; \delta
\]
is valid for a set \(Q\) if and only if \(Q \subseteq \{(x,y) : w^\top(x,y)\le \delta\}\). Designing strong valid inequalities tailored to a specific problem class typically requires nontrivial proofs and insight into the combinatorial structure of the problem.

\noindent Separately, an inequality is optimality-preserving for Eq.~\eqref{eq:MILP} if it does not change the optimal value \(Z^*=\max\{c^\top x+h^\top y:(x,y)\in S\}\), i.e.,
\[
  \max\{c^\top x+h^\top y:(x,y)\in S,\; w^\top(x,y)\le\delta\}=Z^*,
\]
even though it may remove other feasible solutions~\cite{laporte1999optimality,da2015optimality}. We reserve the term optimality-preserving for inequalities that are proved to satisfy this. Appendix~\ref{app:preliminaries} reviews the geometry of valid and optimality-preserving inequalities and their effect on the linear relaxation set (Appendix~\ref{app:geometric_intuition}), polyhedra and facets (Appendix~\ref{app:polyhedra_facets}), and facet-defining valid inequalities (Appendix~\ref{app:facet_defining}).

\noindent \textsc{SAGE}'s acceleration inequality families differ from several classical objects. A valid inequality is proved to contain all feasible integer points in \(S\). A formulation strengthening is a reformulation or augmentation with established validity for the model class. Benders cuts are derived from a decomposition subproblem and added through a master-subproblem scheme. By contrast, a \textsc{SAGE} acceleration inequality family is proposed before solving and kept because it passes empirical optimal solution screening (OSS) and LP-separation checks and improves solver progress. OSS checks whether the family preserves stored reference solutions on sampled instances. No global proof of validity or optimality preservation is assumed.

\noindent Modern MILP solvers routinely inject cuts that are general purpose during solving and that apply to many formulations~\citep{chvatal1973edmonds,balas1979disjunctive}. Commercial solvers such as Gurobi enable these cut-generation routines by default or through solver parameters~\citep{gurobi2026parameters}. Separately, problem-specific structure often gives rise to inequalities that improve solver speed in vehicle routing, network design, scheduling, and graph models~\citep{toth2002vehicle,lysgaard2004new,costa2009benders,queyranne1991single,aref2020modeling}. Given a symbolic MILP formulation, \textsc{SAGE} automatically proposes, empirically screens, and evolves acceleration inequality families to improve solve time.

\section{Proposed Method}
\label{sec:solution_method}

\begin{figure}[t]
  \centering
  \includegraphics[width=0.8\linewidth]{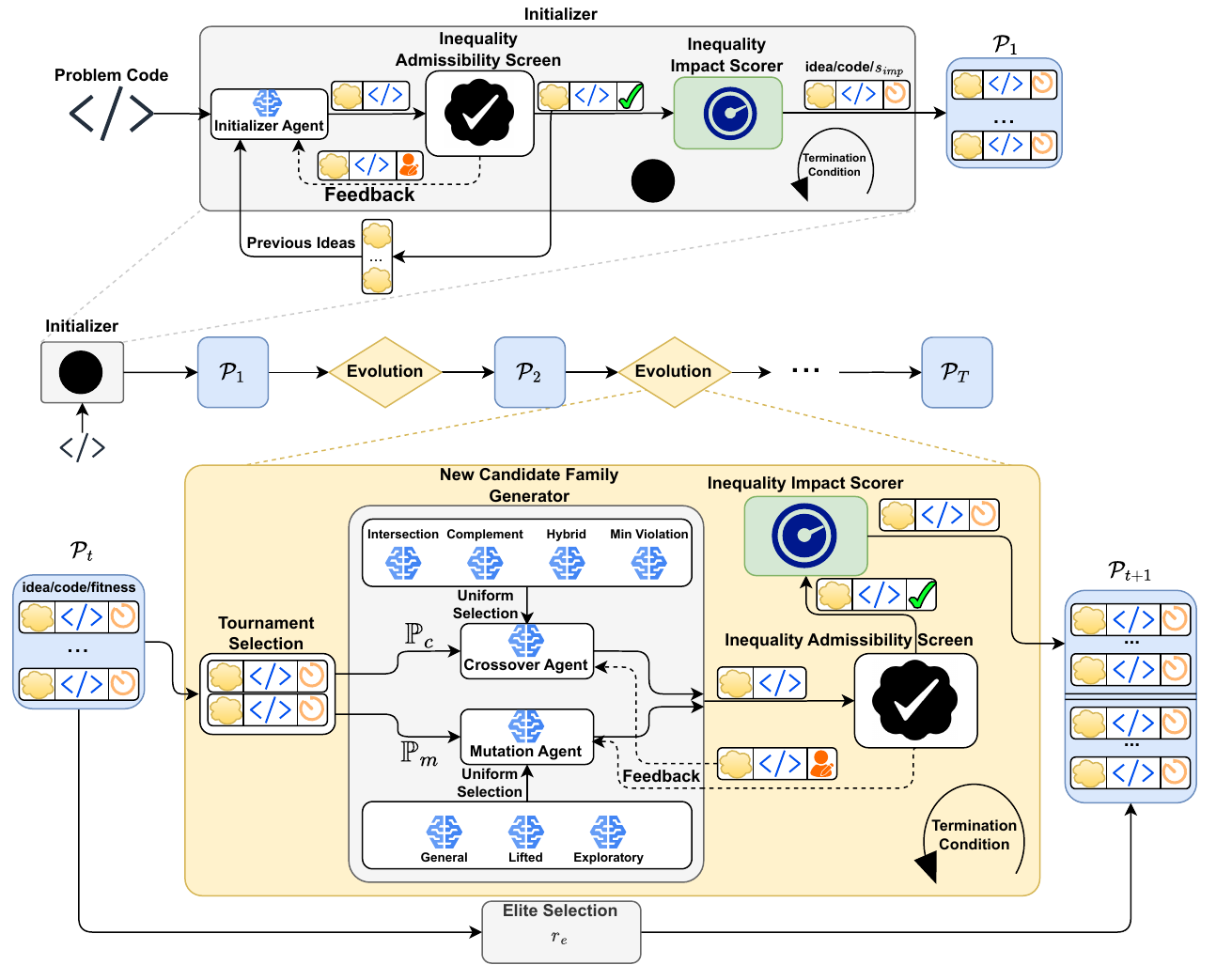}
  \caption{\textsc{SAGE} workflow: an initializer agent proposes a population of candidate acceleration inequality families. Each candidate passes through the Inequality Admissibility Screen for code validity, OSS on the admissibility set \(D_{adm}\), and an LP-separation check, then through the Inequality Impact Scorer on \(D_{imp}\) to obtain an inequality impact score based on the optimality gap. Tournament/elite selection and crossover/mutation driven by LLM agents generate the next population.}
  \Description{A workflow diagram showing initialization, admissibility screening, impact scoring, selection, crossover, and mutation in SAGE.}
  \label{fig:flowchart_placeholder}
\end{figure}

\noindent We propose \textsc{SAGE}, an evolutionary algorithm powered by multiple LLM agents to iteratively generate and refine acceleration inequality families that speed up solving across instances of a given MILP model. Candidate acceptance is determined by empirical admissibility and impact screens, not by a proof of global validity. Fig.~\ref{fig:flowchart_placeholder} illustrates the flow diagram of \textsc{SAGE} at a high level. In addition, Appendix~\ref{sec:llm_evocut} (Algorithm~\ref{alg:evocut}) provides the pseudocode and additional details for the overall procedure. Our proposed method can be summarized in three main phases: (1) data preprocessing, (2) population initialization (driven by the initializer agent), and (3) evolution (crossover and mutation driven by LLM agents). Next, we describe these phases, the Inequality Admissibility Screen, and the Inequality Impact Scorer.

\noindent We begin by splitting available development instances for a given MILP into an impact-assessment set $D_{imp}$ and an admissibility set $D_{adm}$. We use $D_{adm}$ only as a development screen for OSS and LP-separation checks, and $D_{imp}$ to score candidate family quality. For every instance $i\in D_{imp}$, we run the baseline MILP under a fixed computational budget. We use Gurobi's \texttt{WorkLimit} parameter, set to 10~\citep{gurobi2026parameters}, and record its terminal optimality gap, $\mathrm{gap}_{\mathrm{ref}}(i)$. We construct $D_{imp}$ so that $\mathrm{gap}_{\mathrm{ref}}(i)>0$ for all $i\in D_{imp}$. This \textit{reference gap} serves as the baseline for assessing the effect of candidate acceleration inequality families on solver performance within \textsc{SAGE}. For the smaller subset $D_{adm}$, we also store (i) a solver solution $(\hat x_i^{*}, \hat y_i^{*})$ from solving the MILP to an optimality gap tolerance of $10^{-5}$ (for the OSS check) and (ii) the LP relaxation optimum $(x_{i}^{\mathrm{LP}},y_{i}^{\mathrm{LP}})$ (for the LP-separation check). Thus, $D_{imp}$ stores $\mathrm{gap}_{\mathrm{ref}}(i)$, while $D_{adm}$ stores $(\hat x_i^{*}, \hat y_i^{*})$, $(x_{i}^{\mathrm{LP}},y_{i}^{\mathrm{LP}})$, and $\mathrm{gap}_{\mathrm{ref}}(i)$ for each instance $i$.

\noindent With these datasets in place, \textsc{SAGE} seeds an initial population of candidate acceleration inequality families using an initializer agent. At each iteration, the LLM agent receives three inputs: (1) the complete MILP formulation code with a natural language description of model components (Fig.~\ref{fig:evocut_symbolic_overview}), (2) the set of acceleration inequality families generated so far, and (3) instructions to propose novel acceleration inequality families intended to separate poor LP-relaxation behavior in Eq.~\eqref{eq:LP_relaxation}. The LLM response includes a brief idea for the proposed family together with executable code (e.g., Pyomo~\cite{hart2011pyomo}). Each LLM-proposed candidate acceleration inequality family is first passed through the Inequality Admissibility Screen, which checks whether the candidate can be implemented, preserves the stored reference solutions in the admissibility set \(D_{adm}\), and separates at least one LP-relaxation solution. We log each family idea to prevent duplication. This loop continues until the population reaches its predefined size, at which point the admissible and distinct candidates enter the evolutionary search phase (Fig.~\ref{fig:flowchart_placeholder}).

\begin{figure}[bt]
    \centering
    \includegraphics[width=0.7\columnwidth]{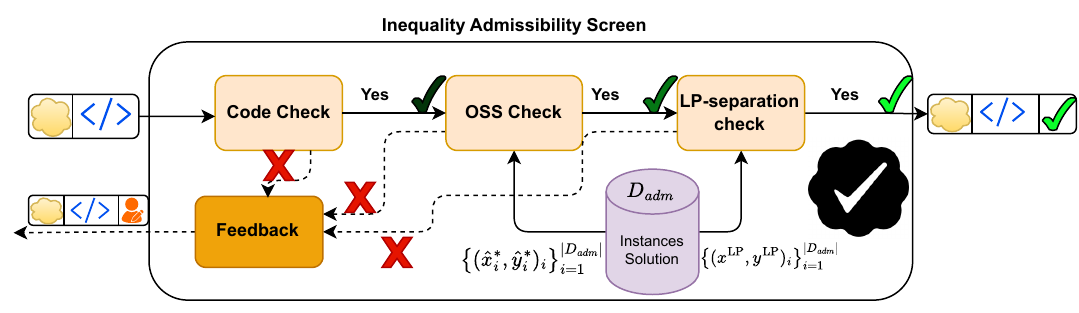}
    \caption{\textsc{SAGE} Inequality Admissibility Screen: candidate family code must compile (code check), pass OSS check on the admissibility set \(D_{adm}\) (i.e., not exclude the recorded reference solution), and separate at least one stored fractional LP optimum (LP-separation check). Failures trigger feedback and regeneration.}
    \Description{An admissibility-screen flow diagram with code, optimal solution screening, and LP-separation checks.}
    \label{fig:verifier_placeholder}
\end{figure}

\noindent To enter the population, each newly proposed candidate acceleration inequality family \(C\) must pass three admissibility checks (Fig.~\ref{fig:verifier_placeholder}): (i) a code check, (ii) an OSS check, and (iii) an LP-separation check. Whenever a check fails, the agent receives the diagnostic as feedback and may revise the candidate up to a preset retry limit (Algorithm~\ref{alg:evocut} in Appendix~\ref{sec:llm_evocut}). First, we run the code check by parsing and compiling the code snippet that implements \(C\). If a syntax or runtime error occurs, $C$ fails the code check. Next, we perform the OSS check to ensure that adding \(C\) does not exclude the recorded reference solutions on \(D_{adm}\). Concretely, for each \(i\in D_{adm}\), we ensure that the recorded solution \((\hat x_i^{*}, \hat y_i^{*})\) remains feasible after appending \(C\) to the baseline MILP (Eq.~\eqref{eq:MILP}). We fix \((x,y)=(\hat x_i^{*}, \hat y_i^{*})\) and solve the resulting reduced model to test feasibility. Because \(C\) may introduce auxiliary variables, we cannot simply evaluate the inequalities on the fixed solution. Instead, we ask whether there exists \(z\) that makes the added inequalities feasible. This solve is efficient because all original variables are fixed. If the reduced model is infeasible, \(C\) fails the OSS check. Formally, for instance \(i\), \(C\) passes the check iff there exists \(z\) such that
\[
  A_r'\,\hat x_i^{*} + G_r'\,\hat y_i^{*} + H_r'\,z \;\le\; \delta_r \quad \forall r \in R
\]
is feasible. Otherwise, \(C\) fails the OSS check for instance \(i\).

\noindent If \(C\) passes the OSS check, we finally run the LP-separation check by discarding any candidate acceleration inequality family that does not separate any stored LP optimum in \(D_{adm}\). For each \(i\in D_{adm}\), we relax all integrality constraints, fix \((x,y)=\bigl(x_i^{\mathrm{LP}},y_i^{\mathrm{LP}}\bigr)\), and test feasibility with \(C\) appended. We retain \(C\) iff there exists \(i\in D_{adm}\) such that \(\not\exists z\) with
\[
  A_r'\,x_i^{\mathrm{LP}} + G_r'\,y_i^{\mathrm{LP}} + H_r'\,z \;\le\; \delta_r \quad \forall r \in R
\]
(i.e., the reduced LP becomes infeasible). Otherwise, \(C\) is discarded. This LP-separation check is a minimal check: passing it is sufficient to show that \(C\) removes at least one sampled point of the LP relaxation. It does not imply that all alternative LP optima are removed, that the root bound improves, or that branch-and-bound search improves. These effects are assessed by the impact scorer on \(D_{imp}\) and by the test-set solve metrics.

\noindent Candidate families that pass this screen are then processed by the Inequality Impact Scorer. We append \(C\) to the baseline MILP and solve every impact-assessment instance \(i\in D_{imp}\) under the same budget, record the resulting optimality gap \(\mathrm{gap}_{\mathrm{aug}}(i)\), and compare it to \(\mathrm{gap}_{\mathrm{ref}}(i)\). The key intuition is that a candidate family should receive a higher inequality impact score when it reduces the gap on average, and a lower score when it increases it. We implement this by averaging the signed relative gap change over \(D_{imp}\) and applying a monotone mapping, which gives the score
\[
d(i)=\frac{\mathrm{gap}_{\mathrm{aug}}(i)-\mathrm{gap}_{\mathrm{ref}}(i)}
{\mathrm{gap}_{\mathrm{ref}}(i)},
\qquad
\bar d=\frac{1}{|D_{imp}|}\sum_{i\in D_{imp}}d(i),
\qquad
s_{imp}(C)=10e^{-\bar d}.
\]
A beneficial family has \(\bar d<0\), so larger gap reductions yield exponentially higher inequality impact scores \(s_{imp}(C)\) (Appendix~\ref{app:impact_score}).

\noindent With the initial population of admissible and scored candidate families ready, the \textsc{SAGE} algorithm evolves the population for \(T\) generations. The main steps of the evolution are as follows: \textbf{(i)} Elitism: The top \(r_e\) fraction of families with the highest inequality impact scores carry over to the next generation. \textbf{(ii)} Selection: Parent families are picked via tournament selection based on inequality impact score. \textbf{(iii)} Crossover: With probability \(\mathbb{P}_c\), two parents are passed to a Crossover LLM, which attempts to produce a new child family \(C_o\) by merging parent features. \textbf{(iv)} Mutation: With probability \(\mathbb{P}_m\), a single parent undergoes a Mutation LLM, which alters coefficients or terms to produce a variant \(C_m\). \textbf{(v)} Admissibility Screening \& Impact Scoring: Each newly created candidate \((C_o\text{ or }C_m)\) is checked by the Inequality Admissibility Screen. Failed candidates are discarded, and a feedback prompt is provided to the LLM (up to a predefined maximum retry limit). Admissible candidates are then forwarded to the Inequality Impact Scorer. Their inequality impact scores are computed and then they are added to the next generation. At each iteration, reproduction (i.e., the generation of a new candidate family) continues until the next population reaches its required size. This evolutionary process then repeats for \(T\) generations. In each reproduction step, \textsc{SAGE} chooses whether to perform crossover (with probability \(\mathbb{P}_c\)) or mutation (with probability \(\mathbb{P}_m\)). Once the operation type is chosen, one of the corresponding agents is selected uniformly at random.

\noindent \textsc{SAGE} draws on seven agents, three for mutation and four for crossover. Each mutation agent rewrites a single parent family. General Mutation modifies its structure or its coefficients. Lifted Mutation strengthens the parent in the style of lifting~\citep{balas1975facets,desrochers1991improvements}, by assigning coefficients to variables its inequalities leave out, or by adding auxiliary structure. Exploratory Mutation moves away from the parent to reach a distinct family. Each crossover agent combines two parent families. Intersection Crossover keeps useful structural elements from each parent. Complementary Crossover looks for structure that neither parent covers. Hybrid Crossover takes structure from one parent and numerical or conditional features from the other. Min Violation Crossover merges the parents so that the child keeps the recorded reference solutions feasible, which is what the OSS check requires. Appendix~\ref{app:agents_prompts} gives the full prompts.

\section{Experiments and Results}
We structure our evaluation around three questions. RQ1 asks whether \textsc{SAGE} improves solver efficiency on unseen instances. We measure checkpoint optimality gaps and time to target gaps. RQ2 asks how sensitive \textsc{SAGE} is to the sizes of the impact assessment set \(D_{imp}\) and admissibility set \(D_{adm}\). We vary \(|D_{imp}|\) and \(|D_{adm}|\) and report test set gap metrics for the selected acceleration inequality family. RQ3 asks how the evolutionary search shapes the acceleration inequality families that the LLMs generate. We compare the acceptance rates of the parent-based operators with those of the one-shot initializer and track inequality impact scores across generations. We also compare \textsc{SAGE} with a token matched initializer only control on JSSP.

\noindent The subsections answer these questions in order. Sections~\ref{sec:post_ga_generalization} and~\ref{sec:time_to_target} answer RQ1 with checkpoint gaps and time to target, and Section~\ref{sec:deployment_time_sanity} then checks that the families measured there exclude no proven optimal solution. Sections~\ref{sec:literature_cut_comparison} to~\ref{sec:solver_robustness_ablations} test the RQ1 result against published valid inequalities, learned solver policies, a second solver, and two internal cut settings. Section~\ref{sec:sensitivity_training_size} answers RQ2, Section~\ref{sec:evo_diagnostics} answers RQ3, and Section~\ref{sec:shuc_case_study} interprets one selected family.
\subsection{Experimental Setup}
\label{sec:experimental_setup}
\noindent We evaluate \textsc{SAGE} by comparing the baseline MILP to the baseline augmented by the selected \textsc{SAGE} acceleration inequality family under fixed solver budgets, across seven NP-hard MILP benchmarks (Table~\ref{tab:post_ga_results}). As the MILP solver, we use the Gurobi Optimizer 10.0.0 (build v10.0.0rc2, Linux 64-bit), which is widely regarded as one of the fastest commercial MIP solvers~\cite{rehfeldt2023faster}. We access Gurobi via the Pyomo interface~\cite{hart2011pyomo} using eight threads. Unless otherwise stated, solver parameters are defaults except for \texttt{MIPGap}=\texttt{1e-5} and the parameters stated here. We fix \texttt{Method}=4 (deterministic concurrent algorithm for continuous models and the initial root relaxation). We set Gurobi's \texttt{WorkLimit} parameter to 10~\citep{gurobi2026parameters}, for \textsc{SAGE}'s internal impact-assessment runs (Section~\ref{sec:solution_method}). We set \texttt{TimeLimit}=2000\,s for test set benchmarking. The work limit controls the development budget for scoring candidates and can therefore affect impact scores, rankings, and the selected family. The time limit defines the test horizon over which we measure terminal gaps, target counts, and speedups. %
In \textsc{SAGE}, we run \(T=20\) generations with a population size of \(|\mathcal{P}|=8\), using a mutation probability of \(\mathbb{P}_m = 0.3\), an elitism ratio of \(r_e = 0.2\), and a crossover probability of \(\mathbb{P}_c = 0.7\). Agents may retry candidate family generation up to three times if admissibility checks fail. We cap one complete search at 20 hours, and every search finished within this cap. The runtime varies with the LLM inference provider, its latency and rate limits, and the solver time spent scoring candidates. We use Gemini 3 Pro~\citep{googledeepmind2025gemini} as the main LLM in \textsc{SAGE} (Table~\ref{tab:llm_params} in Appendix~\ref{app:llm_params} gives the full API settings and approximate costs). Unless otherwise stated, the main results use the family selected from the Gemini 3 Pro \textsc{SAGE} run for each benchmark. To assess sensitivity to the LLM, we also run \textsc{SAGE} with DeepSeek-R1~\cite{guo2025deepseek} and GPT-5.1~\cite{openai2025gpt51} (Table~\ref{tab:post_ga_results}).

Our four core benchmarks include%
: Traveling Salesman Problem (TSP), Multi-Commodity Network Design (MCND), Capacitated Warehouse Location Problem (CWLP), and Job Shop Scheduling Problem (JSSP). Additionally, we report results on the Pickup and Delivery Problem with Time Windows (PDPTW), International Mathematical Olympiad 2025 Problem~6 (IMO6), and a unit commitment benchmark with a horizon under one hour (SHUC) (Table~\ref{tab:post_ga_results}). IMO6 is a newly formulated MILP benchmark of a 2025 IMO problem, so LLMs are less likely to have been exposed to optimization literature specific to it. PDPTW and SHUC are practically motivated models with many constraints, and some of our instances are included in MIPLIB~\cite{koch2011miplib}. The test set sizes are 54 (TSP), 81 (MCND), 36 (CWLP), 80 (JSSP), 165 (PDPTW), 34 (IMO6), and 56 (SHUC). Appendix~\ref{app:benchmarks} provides a dedicated subsection for each benchmark problem, including its formulation, instance source, and dataset details: TSP (Appendix~\ref{app:tsp}), MCND (Appendix~\ref{app:mcnd}), CWLP (Appendix~\ref{app:cwlp}), JSSP (Appendix~\ref{app:jssp}), PDPTW (Appendix~\ref{app:pdptw}), IMO6 (Appendix~\ref{app:imo6}), and SHUC (Appendix~\ref{app:shuc}). Tables~\ref{tab:benchmark_scale} and~\ref{tab:dev_set_scale} in Appendix~\ref{app:benchmarks} summarize the held-out test set scales and the synthetic development source scales, respectively. During evolution, \textsc{SAGE} relies on relatively small development datasets $D_{imp}$ (default size 10) and $D_{adm}$ (default size 2), as detailed in Section~\ref{sec:solution_method}. The default \(|D_{adm}|=2\) is used only for candidate screening during development. We do not use benchmark test instances during development. For all problems except IMO6, both $D_{imp}$ and $D_{adm}$ are drawn from synthetic generators released alongside our code. The test set $D_t$ is separate and consists of public benchmark instances. For IMO6, all instances are generated and split into $D_{imp}$, $D_{adm}$, and $D_t$.

\subsection{Generalization to Unseen Instances}
\label{sec:post_ga_generalization}

We compare the baseline MILP to the baseline MILP augmented with the selected \textsc{SAGE} acceleration inequality family on unseen test instances in \(D_t\). Each instance is solved twice under a 2000 s wall clock limit: once as the baseline MILP (reference) and once with the selected acceleration inequality family. At each of the seven checkpoints (5 s, 10 s, 50 s, 150 s, 300 s, 1000 s, 2000 s), we record the optimality gap \(g\) reported by the solver. We then compute the stabilized relative gap improvement:
\[
\Delta_g(i,t)=
\frac{g_{\mathrm{ref}}(i,t)-g_{\mathrm{aug}}(i,t)}
     {\max\{g_{\mathrm{ref}}(i,t),\epsilon\}},
\quad \epsilon=10^{-5}.
\]
This avoids division by very small reference gaps. We report \(\bar{\Delta}_g\), the average of \(\Delta_g\) over all instances in \(D_t\). At a checkpoint, \(\Delta_g>0\) iff \(g_{\mathrm{aug}}<g_{\mathrm{ref}}\). We summarize checkpoint performance in Table~\ref{tab:post_ga_results}, reporting \(\bar{\Delta}_g\ {\scriptstyle\pm\sigma}\ (\%)\) across the test set at each checkpoint for each benchmark and LLM.
\begin{table}[htb]
		  \centering
		  \scriptsize
		  \caption{Test set mean stabilized relative gap improvement $\bar{\Delta}_g$ (\%) from adding the selected \textsc{SAGE} acceleration inequality family (Gurobi, three LLMs). Mean $\pm\sigma$ is reported at each checkpoint. Positive values correspond to $g_{\mathrm{aug}}<g_{\mathrm{ref}}$ at that checkpoint. Average reports the row average of checkpoint means.}
		  \label{tab:post_ga_results}
		  \setlength{\tabcolsep}{3pt}
		  \begin{tabular}{llccccccc@{\hspace{6pt}}|@{\hspace{6pt}}c}
	    \toprule
	    Problem & LLM &
	    5 s & 10 s & 50 s & 150 s & 300 s & 1000 s & 2000 s & Average \\
	    \midrule
		    \multirow{3}{*}{TSP}
		      & DeepSeek-R1
		        & \meanstd{8.2}{26.0}
	        & \meanstd{9.1}{26.9}
	        & \meanstd{11.9}{32.2}
	        & \meanstd{18.1}{30.9}
		        & \meanstd{18.5}{33.8}
		        & \meanstd{12.9}{31.3}
		        & \meanstd{20.7}{30.7} & 14.2 \\
		      & Gemini 3 Pro
		        & \meanstd{13.1}{27.5}
		        & \meanstd{15.0}{27.9}
		        & \meanstd{22.6}{31.3}
	        & \meanstd{34.2}{30.3}
		        & \meanstd{27.6}{33.1}
		        & \meanstd{26.4}{33.3}
		        & \meanstd{37.0}{31.1} & 25.1 \\
		      & GPT-5.1
		        & \meanstd{12.1}{28.4}
		        & \meanstd{8.8}{29.7}
		        & \meanstd{15.6}{33.2}
	        & \meanstd{17.4}{34.1}
		        & \meanstd{11.3}{35.0}
		        & \meanstd{16.7}{33.8}
		        & \meanstd{18.2}{31.6} & 14.3 \\
		    \midrule
		    \multirow{3}{*}{MCND}
	      & DeepSeek-R1
	        & \meanstd{3.3}{17.8}
        & \meanstd{4.3}{19.9}
        & \meanstd{11.7}{27.0}
        & \meanstd{13.0}{28.2}
	        & \meanstd{14.7}{28.9}
	        & \meanstd{18.9}{32.4}
		        & \meanstd{16.7}{29.7} & 11.8 \\
		      & Gemini 3 Pro
		        & \meanstd{6.8}{17.7}
		        & \meanstd{8.2}{19.9}
		        & \meanstd{15.1}{25.8}
	        & \meanstd{17.4}{26.9}
		        & \meanstd{17.8}{25.5}
		        & \meanstd{23.2}{27.9}
			        & \meanstd{15.8}{28.6} & 14.9 \\
	      & GPT-5.1
	        & \meanstd{25.5}{30.1}
	        & \meanstd{22.3}{29.6}
	        & \meanstd{16.5}{27.7}
        & \meanstd{6.9}{30.0}
	        & \meanstd{15.6}{28.9}
	        & \meanstd{18.1}{29.8}
		        & \meanstd{20.4}{25.2} & 17.9 \\
	    \midrule
	    \multirow{3}{*}{CWLP}
	      & DeepSeek-R1
	        & \meanstd{17.8}{21.4}
        & \meanstd{18.6}{22.4}
        & \meanstd{25.2}{26.7}
        & \meanstd{36.3}{26.6}
	        & \meanstd{34.1}{34.7}
	        & \meanstd{33.8}{29.9}
		        & \meanstd{32.5}{30.5} & 28.3 \\
	      & Gemini 3 Pro
	        & \meanstd{43.8}{20.7}
	        & \meanstd{49.4}{22.3}
	        & \meanstd{68.8}{28.6}
        & \meanstd{76.3}{23.8}
	        & \meanstd{64.8}{23.0}
	        & \meanstd{35.2}{39.3}
		        & \meanstd{30.1}{38.9} & 52.6 \\
	      & GPT-5.1
	        & \meanstd{35.9}{20.6}
	        & \meanstd{41.3}{22.2}
	        & \meanstd{64.7}{28.6}
        & \meanstd{73.7}{23.7}
	        & \meanstd{57.5}{23.0}
	        & \meanstd{31.9}{39.2}
		        & \meanstd{18.3}{38.8} & 46.2 \\
	    \midrule
		    \multirow{3}{*}{JSSP}
		      & DeepSeek-R1
		        & \meanstd{41.7}{18.5}
		        & \meanstd{38.2}{19.6}
		        & \meanstd{27.5}{22.8}
	        & \meanstd{37.5}{25.6}
		        & \meanstd{28.9}{24.1}
		        & \meanstd{39.8}{22.7}
			        & \meanstd{41.5}{20.7} & 36.4 \\
		      & Gemini 3 Pro
		        & \meanstd{64.1}{9.9}
		        & \meanstd{63.7}{10.7}
		        & \meanstd{63.3}{14.4}
        & \meanstd{61.8}{16.6}
		        & \meanstd{63.2}{16.5}
		        & \meanstd{59.4}{21.2}
			        & \meanstd{59.2}{20.6} & 62.1 \\
		      & GPT-5.1
		        & \meanstd{49.0}{14.8}
		        & \meanstd{46.5}{16.0}
		        & \meanstd{39.2}{19.5}
	        & \meanstd{38.0}{23.0}
		        & \meanstd{37.4}{24.6}
		        & \meanstd{42.8}{23.8}
			        & \meanstd{56.7}{20.3} & 44.2 \\
		    \midrule
		    \multirow{3}{*}{PDPTW}
		      & DeepSeek-R1
		        & \meanstd{0.2}{4.8}
	        & \meanstd{-0.6}{5.3}
	        & \meanstd{5.4}{9.2}
	        & \meanstd{12.8}{14.9}
		        & \meanstd{17.1}{19.6}
		        & \meanstd{18.3}{21.5}
			        & \meanstd{24.0}{19.9} & 11.0 \\
		      & Gemini 3 Pro
		        & \meanstd{46.9}{32.4}
	        & \meanstd{47.5}{32.1}
	        & \meanstd{54.3}{29.2}
        & \meanstd{62.6}{28.6}
		        & \meanstd{55.8}{35.1}
		        & \meanstd{64.2}{31.2}
			        & \meanstd{70.7}{29.4} & 57.4 \\
		      & GPT-5.1
		        & \meanstd{7.5}{5.2}
	        & \meanstd{8.2}{5.6}
	        & \meanstd{9.7}{15.7}
	        & \meanstd{12.9}{16.8}
		        & \meanstd{14.3}{21.9}
		        & \meanstd{18.4}{17.2}
			        & \meanstd{27.8}{25.5} & 14.1 \\
		    \midrule
		    \multirow{3}{*}{IMO6}
	      & DeepSeek-R1
	        & \meanstd{12.2}{13.0}
        & \meanstd{14.3}{14.9}
        & \meanstd{23.7}{20.7}
        & \meanstd{26.4}{19.6}
	        & \meanstd{31.0}{18.7}
	        & \meanstd{42.0}{14.8}
		        & \meanstd{53.6}{11.1} & 29.0 \\
	      & Gemini 3 Pro
	        & \meanstd{13.6}{11.2}
	        & \meanstd{15.7}{13.5}
	        & \meanstd{24.6}{18.4}
        & \meanstd{25.0}{21.4}
	        & \meanstd{28.9}{23.0}
	        & \meanstd{24.3}{26.8}
	        & \meanstd{19.7}{27.8} & 21.7 \\
	      & GPT-5.1
	        & \meanstd{14.0}{14.1}
	        & \meanstd{16.6}{16.2}
	        & \meanstd{26.9}{22.0}
        & \meanstd{29.5}{25.2}
	        & \meanstd{34.1}{25.0}
	        & \meanstd{43.6}{20.8}
	        & \meanstd{53.3}{17.8} & 31.1 \\
	    \midrule
	    \multirow{3}{*}{SHUC}
	      & DeepSeek-R1
	        & \meanstd{44.8}{37.9}
	        & \meanstd{43.9}{38.3}
	        & \meanstd{27.3}{37.0}
        & \meanstd{18.3}{31.8}
	        & \meanstd{9.7}{29.4}
	        & \meanstd{7.6}{28.7}
		        & \meanstd{6.2}{27.6} & 22.5 \\
	      & Gemini 3 Pro
	        & \meanstd{51.2}{35.1}
	        & \meanstd{50.6}{34.8}
	        & \meanstd{32.2}{37.7}
        & \meanstd{6.3}{29.4}
	        & \meanstd{3.7}{24.9}
	        & \meanstd{2.6}{24.2}
		        & \meanstd{4.3}{25.0} & 21.6 \\
	      & GPT-5.1
	        & \meanstd{34.2}{36.8}
	        & \meanstd{34.8}{37.6}
	        & \meanstd{25.2}{35.7}
        & \meanstd{6.9}{29.6}
	        & \meanstd{4.0}{26.4}
	        & \meanstd{6.0}{24.4}
		        & \meanstd{7.0}{24.4} & 16.9 \\
	    \bottomrule
	  \end{tabular}
		\end{table}
\noindent In Table~\ref{tab:post_ga_results}, the mean relative gap improvement is positive at 2000\,s for all 21 benchmark/LLM pairs. At 2000\,s, the mean ranges from 4.3\% on SHUC with Gemini 3 Pro to 70.7\% on PDPTW with Gemini 3 Pro. The row-average checkpoint improvement is positive for all 21 rows. Gemini 3 Pro has the largest row average on TSP (25.1\%), CWLP (52.6\%), JSSP (62.1\%), and PDPTW (57.4\%). The largest single checkpoint mean is 76.3\% on CWLP with Gemini 3 Pro at 150\,s. GPT-5.1 has the largest row average on MCND (17.9\%) and IMO6 (31.1\%). DeepSeek-R1 has the largest row average on SHUC (22.5\%). Averaging the 7 row results by LLM gives aggregate improvements of 36.5\% for Gemini 3 Pro, 26.4\% for GPT-5.1, and 21.9\% for DeepSeek-R1. Fig.~\ref{fig:gap_rel_panel} in Appendix~\ref{app:gap_improvement_curves} illustrates the full 2000\,s time profiles. For SHUC with Gemini 3 Pro, the checkpoint means move from 51.2\% at 5\,s to 3.7\% at 300\,s and 4.3\% at 2000\,s.

\noindent Appendix~\ref{app:additional_experiments} also reports trajectory checks for representative instances. Appendix~\ref{app:tsp_examples} (Fig.~\ref{fig:tsp_examples}) compares four TSPLIB instances under the same 2000\,s limit, and the augmented gap curve is below the baseline curve for most of the horizon in all four panels. Appendix~\ref{app:tsp_best_bound_exp} (Fig.~\ref{fig:tsp_best_bound_panel}) uses the same four instances and shows higher best-bound curves for the augmented model. Appendix~\ref{app:tsp_pdi_exp} (Fig.~\ref{fig:tsp_pdi_panel}) reports roughly \(75\!-\!95\%\) lower PDI for the augmented model on those instances. Appendix~\ref{app:mcnd_gap_failure} (Fig.~\ref{fig:mcnd_gap_failure}) shows four MCND instances where the selected acceleration inequality family slightly increases the gap, illustrating that the added inequalities can negatively impact some individual cases.

\noindent As an additional control for prior exposure, we created the unpublished Reconfigurable Network Topology Design (RNTD) problem. The SAGE agents received only a Pyomo formulation with generic graph, node, link, and commodity terms, without the problem name or application description. Appendix~\ref{app:rntd} gives the formulation, selected family, generator, and data split. Appendix~\ref{app:rntd_control} gives the anonymized prompt and full results. Tables~\ref{tab:rntd_gap} and~\ref{tab:rntd_time} in Appendix~\ref{app:rntd_control} show positive mean gap improvement at six of seven checkpoints, including $23\pm19\%$ at 2000\,s, and a mean SGM speedup of $3.3\times$.

\subsection{Time to Target Optimality Gap}
\label{sec:time_to_target}
To complement the checkpoint gap analysis (Section~\ref{sec:post_ga_generalization}), we measure how much \textsc{SAGE} shortens the time to reach target optimality gaps \(g\in\{10^{-5},10^{-4},10^{-3},10^{-2},10^{-1}\}\) on the unseen test sets. For each instance and method (baseline vs.\ augmented), we compute the time to reach gap \(t(g)\) by linearly interpolating the solver trace. If a run never reaches the target gap, we assign a penalized time \(t(g)=2T_{\max}\). We aggregate using the shifted geometric mean (SGM, with shift \(s{=}1\)):
\[
\mathrm{SGM}(t)=\bigl(\prod_{i\in\mathcal{I}_g}(t_i+s)\bigr)^{1/|\mathcal{I}_g|}-s,
\]
where \(\mathcal{I}_g\) is the set of instances for which at least one method reaches \(g\). Table~\ref{tab:post_ga_time} reports \(\mathrm{SGM}(t_{\mathrm{ref}})/\mathrm{SGM}(t_{\mathrm{aug}})\) and solved counts (B/A).
\begin{table}[t]
  \centering
  \scriptsize
  \caption{Time to reach target gaps \(g\): baseline vs.\ \textsc{SAGE}. B/A: instances reaching \(g\) within 2000\,s (Baseline/\textsc{SAGE}). SGM: shifted geometric mean speedup (\(>1\) better). Average reports the row average over all target gaps.}
  \label{tab:post_ga_time}
  {\setlength{\tabcolsep}{5pt}%
  \renewcommand{\arraystretch}{1.0}%
  \begin{tabular}{@{}l*{5}{cc}@{\hspace{6pt}}|@{\hspace{6pt}}cc@{}}
    \toprule
    & \multicolumn{2}{c}{$g=10^{-5}$}
    & \multicolumn{2}{c}{$g=10^{-4}$}
    & \multicolumn{2}{c}{$g=10^{-3}$}
    & \multicolumn{2}{c}{$g=10^{-2}$}
    & \multicolumn{2}{c@{\hspace{6pt}}|@{\hspace{6pt}}}{$g=10^{-1}$}
    & \multicolumn{2}{c@{}}{Average} \\
    \cmidrule(lr){2-3}\cmidrule(lr){4-5}\cmidrule(lr){6-7}\cmidrule(lr){8-9}\cmidrule(lr){10-11}\cmidrule(lr){12-13}
    Problem
      & B/A & SGM
      & B/A & SGM
      & B/A & SGM
      & B/A & SGM
      & B/A & SGM
      & B/A & SGM \\
    \midrule
    TSP
      & 26/35 & 3.21
      & 27/37 & 3.13
      & 27/37 & 3.16
      & 31/44 & 3.00
      & 44/47 & 1.83
      & 31.0/40.0 & 2.87 \\
    MCND
      & 56/57 & 1.04
      & 62/62 & 1.04
      & 62/62 & 1.03
      & 69/69 & 1.04
      & 76/74 & 0.92
      & 65.0/64.8 & 1.01 \\
    CWLP
      & 26/23 & 2.44
      & 35/36 & 4.24
      & 35/36 & 4.42
      & 35/36 & 5.19
      & 35/36 & 3.83
      & 33.2/33.4 & 4.02 \\
    JSSP
      & 7/8 & 2.18
      & 7/8 & 2.31
      & 7/8 & 2.42
      & 7/8 & 2.55
      & 10/19 & 7.20
      & 7.6/10.2 & 3.33 \\
    PDPTW
      & 0/5 & 5.18
      & 0/5 & 5.18
      & 0/5 & 5.19
      & 0/5 & 5.25
      & 0/20 & 6.77
      & 0.0/8.0 & 5.51 \\
    IMO6
      & 7/8 & 1.23
      & 7/8 & 1.23
      & 7/8 & 1.23
      & 7/8 & 1.24
      & 7/8 & 1.39
      & 7.0/8.0 & 1.26 \\
    SHUC
      & 2/2 & 1.47
      & 6/6 & 1.28
      & 17/17 & 1.26
      & 39/39 & 1.25
      & 53/54 & 1.34
      & 23.4/23.6 & 1.32 \\
    \bottomrule
  \end{tabular}
  }
\end{table}
\noindent In Table~\ref{tab:post_ga_time}, \(\mathrm{SGM}(t_{\mathrm{ref}})/\mathrm{SGM}(t_{\mathrm{aug}}) > 1\) for all problems at \(g \in \{10^{-5},10^{-4},10^{-3},10^{-2}\}\), and for six of seven problems at \(g=10^{-1}\). The row-average SGM speedup is largest on PDPTW (\(5.51\times\)), followed by CWLP (\(4.02\times\)), JSSP (\(3.33\times\)), and TSP (\(2.87\times\)). The largest single target-gap SGM speedup is \(7.20\times\) on JSSP at \(g=10^{-1}\). MCND is nearly neutral overall (\(1.01\times\)) and is the only benchmark below 1 at \(g=10^{-1}\) (\(0.92\times\)). For PDPTW, the baseline reaches none of the target gaps within 2000\,s (0 instances at each target), while the augmented model reaches targets on a subset of instances (5 instances for \(g\le 10^{-2}\) and 20 at \(g=10^{-1}\)).

Table~\ref{tab:post_ga_time} reports deployment solve time. We separately measure discovery cost, deployment setup, and payback on the test sets listed in Appendix~\ref{app:payback}. Payback uses the target $g\le 10^{-3}$ and caps solver time to target at $T_{\max}$.

\noindent Table~\ref{tab:payback} in Appendix~\ref{app:payback} shows that the mean solver time with \textsc{SAGE} is lower on all six benchmarks. The estimated payback ranges from 238 solves on TSP to 5276 solves on MCND, and the mean added setup time is below 1 second on five of the six benchmarks and 10.55 seconds on IMO6.

\noindent Table~\ref{tab:selected_family_reporting} in Appendix~\ref{app:model_size_lp} reports model sizes before solver presolve and LP relaxation results for the selected families. Only JSSP and SHUC introduce auxiliary variables. Among benchmarks with proven objectives, the mean per-instance gap closure is smallest on MCND (2.81\%, mean LP gap 25.45\% to 24.62\%) and largest on CWLP (98.59\%, mean LP gap 24.93\% to 0.38\%).

\subsection{Post-hoc Screening Against Proven Optimal Solutions}
\label{sec:deployment_time_sanity}
\noindent This check applies to the same selected families evaluated in Sections~\ref{sec:post_ga_generalization} and~\ref{sec:time_to_target}. \textsc{SAGE} screens candidate acceleration inequality families using the small admissibility set \(D_{adm}\) during development (Section~\ref{sec:solution_method}). In the default setting, this admissibility set has only two instances. After selecting an acceleration inequality family, we run an additional offline check on the held-out test sets to answer the following question: Does the selected \textsc{SAGE} acceleration inequality family ever exclude a proven optimal solution that we can otherwise compute? This post-hoc check is not an extra step required by \textsc{SAGE} at deployment time.
\noindent For each test instance in every benchmark, we run the baseline MILP for 48\,h and record the instances where the solver proves optimality. Across the full test sets, proven optima are obtained for CWLP at the rate of 36/36. Similarly, the rates are 58/80 for JSSP, 77/81 for MCND, 54/54 for TSP, 9/34 for IMO6, and 53/56 for SHUC. For each such instance, we fix the original decision variables to the proven optimal solution, append the selected \textsc{SAGE} acceleration inequality family, and solve only for any auxiliary variables introduced by the family to test feasibility. In all cases, the feasibility check succeeds. We find no case in which the selected \textsc{SAGE} acceleration inequality family excludes a proven optimal solution in our test set.
\noindent For PDPTW, none of the 48\,h baseline runs proves optimality (0/165), so the above check was not possible. We therefore report the following bound consistency results: using the 2000\,s augmented run from Table~\ref{tab:post_ga_results}, (i) the 2000\,s lower bound does not exceed the 48\,h upper bound and (ii) the 48\,h lower bound does not exceed the 2000\,s upper bound. This condition holds for PDPTW at the rate of 165/165.
\noindent Across the instances for which this check is possible, we detect zero exclusions of the proven optimum. This test set result provides empirical evidence of optimality preservation for the selected families on the evaluated instances.

\noindent Appendix~\ref{app:heldout_lp_separation} reports a related held-out LP separation check. For each test instance, we fix the original variables to the baseline LP optimum, append the selected acceleration inequality family, and solve only for auxiliary variables. Table~\ref{tab:heldout_lp_separation} in Appendix~\ref{app:heldout_lp_separation} shows 100.0\% LP optimum separation on TSP, CWLP, JSSP, PDPTW, and IMO6, 88.9\% on MCND, and 85.7\% on SHUC. This test complements the preservation check by measuring separation of one baseline LP optimum per test instance.

\subsection{Comparison with Static Literature Augmentation}
\label{sec:literature_cut_comparison}
For the four benchmarks with well-known published valid inequalities, we also compare against adding these inequalities up front, using the same time-to-target protocol as in Table~\ref{tab:post_ga_time}. The detailed formulas and sources are given in Appendix~\ref{app:literature_cuts}. Averaged over the five target gaps \(g\in\{10^{-5},10^{-4},10^{-3},10^{-2},10^{-1}\}\), the static literature valid inequalities achieve mean SGM speedups of \(3.01\times\), \(1.91\times\), \(1.01\times\), and \(2.57\times\) on CWLP, JSSP, MCND, and TSP, respectively. On the same benchmarks, the \textsc{SAGE} families achieve \(4.02\times\), \(3.33\times\), \(1.01\times\), and \(2.87\times\). Thus, relative to these instantiated static literature valid inequalities, \textsc{SAGE} is higher by about \(34\%\) on CWLP, \(74\%\) on JSSP, and \(12\%\) on TSP, while MCND is unchanged. The main gains come from CWLP and JSSP. TSP shows a smaller gain, and MCND shows no measurable difference.

\subsection{Complementarity with Learned Solver-Layer Policies}
\label{sec:ml_policy_complementarity}
To check how \textsc{SAGE} interacts with two learned runtime methods that act inside branch-and-bound, we run a factorial CWLP experiment on the held-out 36-instance OR-Library benchmark, with one random seed and a 2000\,s SCIP~10.0 time limit. The two methods are the Turner et al.\ adaptive cut selector~\cite{turner2023adaptive} and the Lawless et al.\ cold-start LLM separator configuration~\cite{lawless2025llms}. The factorial crosses \textsc{SAGE} (off / on, at the model layer) with the ML policy (off / Turner / Lawless-LLM(0) / Lawless-LLM(5), at the solver layer). Turner changes SCIP's cut selector by scoring generated cuts before SCIP decides which cuts to keep. The Turner column uses the reinforcement learning setup from~\cite{turner2023adaptive}. Lawless changes SCIP's separator settings before solving by choosing configurations generated by GPT-4o. The Lawless columns use 100 separator configurations generated by GPT-4o, clustered into LLM(0) (largest-cluster medoid) and LLM(5) (k-medoids with $k{=}5$), exactly as in~\cite{lawless2025llms}. We report the shifted-geometric-mean speedup (SGM, shift $s{=}1$, PAR-2 penalty at 2000\,s) defined in Section~\ref{sec:time_to_target} at the five target gaps used in Table~\ref{tab:post_ga_time}, and the row-averaged speedup. The reference cell is plain SCIP defaults without \textsc{SAGE}.

\begin{table}[htbp]
  \centering
  \small
  \caption{CWLP factorial with learned runtime policies, held-out OR-Library test set, SCIP~10.0, 2000\,s time limit, single seed. The reported value is the average over the five target gaps $g\in\{10^{-5},10^{-4},10^{-3},10^{-2},10^{-1}\}$ of the time-to-target SGM speedup vs the reference cell (\textsc{SAGE} off, plain SCIP defaults), following the protocol of Table~\ref{tab:post_ga_time}. Values $>1$ are faster. The reference cell is 1.00$\times$ by construction.}
  \label{tab:complementarity_ml_policies}
  {%
  \setlength{\tabcolsep}{6pt}
  \renewcommand{\arraystretch}{1.1}
  \begin{tabular}{@{}llc@{}}
    \toprule
    \textsc{SAGE} & Solver-layer policy & Avg SGM speedup \\
    \midrule
    Off & SCIP default                       & 1.00 \\
    Off & Turner et al.                            & 2.12 \\
    Off & Lawless et al.\ LLM(0)                    & 2.58 \\
    Off & Lawless et al.\ LLM(5)                    & 2.60 \\
    \midrule
    On  & SCIP default                       & 3.13 \\
    On  & Turner et al.                             & 3.45 \\
    On  & Lawless et al.\ LLM(0)                    & \textbf{3.51} \\
    On  & Lawless et al.\ LLM(5)                    & 3.43 \\
    \bottomrule
  \end{tabular}
  }
\end{table}

\noindent In Table~\ref{tab:complementarity_ml_policies}, each learned solver-layer method gives a speedup on its own, 2.12$\times$ for Turner, 2.58$\times$ for Lawless et al.\ LLM(0), and 2.60$\times$ for Lawless et al.\ LLM(5), while \textsc{SAGE} alone reaches 3.13$\times$. Combining \textsc{SAGE} with a learned solver-layer policy gives 3.45$\times$ with Turner, 3.51$\times$ with Lawless et al.\ LLM(0), and 3.43$\times$ with Lawless et al.\ LLM(5), so all three combined cells are above \textsc{SAGE} alone and above the matching cells without \textsc{SAGE}.

\subsection{Solver Robustness Ablations}
\label{sec:solver_robustness_ablations}
\subsubsection{Transfer Across Solvers}
\label{sec:cross_solver_transfer}
This cross-solver robustness ablation checks whether the observed gains are specific to a particular solver or persist under a different branch-and-bound implementation and default cut setting. We keep the same acceleration inequality families discovered with \texttt{Gemini 3 Pro} and solve the test instances with two different MILP solvers, Gurobi and HiGHS. Each solver uses its default settings unless specified in Section~\ref{sec:experimental_setup}, and each run uses the same 2000\,s time limit and seven checkpoints as Table~\ref{tab:post_ga_results}. Table~\ref{tab:gemini_two_solvers} reports \(\bar{\Delta}_g\) under each solver.

\begin{table}[t]
  \centering
  \scriptsize
\caption{\texttt{Gemini 3 Pro} selected acceleration inequality family: test stabilized $\bar{\Delta}_g$ (\%) with Gurobi vs.\ HiGHS. Mean $\pm\sigma$ is reported at each checkpoint. Average reports the row average of checkpoint means.}
  \label{tab:gemini_two_solvers}
  \setlength{\tabcolsep}{3pt}
  \begin{tabular}{llccccccc@{\hspace{6pt}}|@{\hspace{6pt}}c}
    \toprule
    Problem & Solver &
    5 s & 10 s & 50 s & 150 s & 300 s & 1000 s & 2000 s & Average \\
    \midrule
	    \multirow{2}{*}{TSP}
      & Gurobi
        & \meanstd{13.1}{27.5}
        & \meanstd{15.0}{27.9}
        & \meanstd{22.6}{31.3}
        & \meanstd{34.2}{30.3}
	        & \meanstd{27.6}{33.1}
	        & \meanstd{26.4}{33.3}
		        & \meanstd{37.0}{31.1} & 25.1 \\
	      & HiGHS
	        & \meanstd{2.5}{13.8}
	        & \meanstd{2.9}{14.9}
	        & \meanstd{11.3}{23.3}
        & \meanstd{18.6}{26.5}
	        & \meanstd{18.5}{27.2}
	        & \meanstd{21.4}{28.1}
		        & \meanstd{21.9}{28.7} & 13.9 \\
	    \midrule
		    \multirow{2}{*}{MCND}
	      & Gurobi
	        & \meanstd{6.8}{17.7}
	        & \meanstd{8.2}{19.9}
	        & \meanstd{15.1}{25.8}
	        & \meanstd{17.4}{26.9}
		        & \meanstd{17.8}{25.5}
		        & \meanstd{23.2}{27.9}
			        & \meanstd{15.8}{28.6} & 14.9 \\
	      & HiGHS
	        & \meanstd{6.0}{15.1}
	        & \meanstd{7.0}{16.9}
	        & \meanstd{6.6}{22.5}
        & \meanstd{1.4}{27.7}
	        & \meanstd{5.5}{28.4}
	        & \meanstd{2.0}{29.3}
		        & \meanstd{5.7}{29.0} & 4.9 \\
	    \midrule
	    \multirow{2}{*}{CWLP}
      & Gurobi
        & \meanstd{43.8}{20.7}
        & \meanstd{49.4}{22.3}
        & \meanstd{68.8}{28.6}
        & \meanstd{76.3}{23.8}
	        & \meanstd{64.8}{23.0}
	        & \meanstd{35.2}{39.3}
		        & \meanstd{30.1}{38.9} & 52.6 \\
	      & HiGHS
	        & \meanstd{25.5}{24.6}
	        & \meanstd{25.3}{24.6}
	        & \meanstd{23.0}{24.8}
        & \meanstd{28.1}{29.8}
	        & \meanstd{39.1}{36.2}
	        & \meanstd{62.3}{32.3}
		        & \meanstd{64.2}{31.9} & 38.2 \\
	    \midrule
		    \multirow{2}{*}{JSSP}
	      & Gurobi
	        	& \meanstd{64.1}{9.9}
		        & \meanstd{63.7}{10.7}
		        & \meanstd{63.3}{14.4}
	        & \meanstd{61.8}{16.6}
		        & \meanstd{63.2}{16.5}
		        & \meanstd{59.4}{21.2}
			        & \meanstd{59.2}{20.6} & 62.1 \\
		      & HiGHS
		        & \meanstd{12.8}{16.5}
		        & \meanstd{14.1}{17.2}
		        & \meanstd{18.9}{22.4}
	        & \meanstd{24.6}{24.8}
		        & \meanstd{21.0}{25.6}
		        & \meanstd{26.7}{26.5}
			        & \meanstd{27.3}{27.5} & 20.8 \\
		    \midrule
		    \multirow{2}{*}{PDPTW}
		      & Gurobi
		        & \meanstd{46.9}{32.4}
	        & \meanstd{47.5}{32.1}
	        & \meanstd{54.3}{29.2}
	        & \meanstd{62.6}{28.6}
		        & \meanstd{55.8}{35.1}
		        & \meanstd{64.2}{31.2}
			        & \meanstd{70.7}{29.4} & 57.4 \\
		      & HiGHS
		        & \meanstd{-0.4}{4.1}
		        & \meanstd{0.3}{4.6}
		        & \meanstd{3.1}{12.8}
	        & \meanstd{7.5}{21.7}
		        & \meanstd{4.2}{26.9}
		        & \meanstd{9.8}{28.4}
			        & \meanstd{12.3}{21.9} & 5.3 \\
		    \midrule
		    \multirow{2}{*}{IMO6}
		      & Gurobi
		        & \meanstd{13.6}{11.2}
	        & \meanstd{15.7}{13.5}
	        & \meanstd{24.6}{18.4}
        & \meanstd{25.0}{21.4}
		        & \meanstd{28.9}{23.0}
		        & \meanstd{24.3}{26.8}
		        & \meanstd{19.7}{27.8} & 21.7 \\
		      & HiGHS
		        & \meanstd{5.3}{12.9}
		        & \meanstd{6.7}{14.5}
		        & \meanstd{12.1}{19.4}
	        & \meanstd{14.8}{21.3}
		        & \meanstd{16.0}{22.7}
		        & \meanstd{13.5}{25.6}
		        & \meanstd{18.6}{24.8} & 12.4 \\
	    \midrule
	    \multirow{2}{*}{SHUC}
	      & Gurobi
	        & \meanstd{51.2}{35.1}
	        & \meanstd{50.6}{34.8}
	        & \meanstd{32.2}{37.7}
        & \meanstd{6.3}{29.4}
	        & \meanstd{3.7}{24.9}
	        & \meanstd{2.6}{24.2}
	        & \meanstd{4.3}{25.0} & 21.6 \\
	      & HiGHS
	        & \meanstd{1.9}{9.2}
	        & \meanstd{1.8}{9.7}
	        & \meanstd{-2.7}{11.4}
        & \meanstd{3.8}{23.0}
	        & \meanstd{17.8}{32.3}
	        & \meanstd{24.3}{28.2}
	        & \meanstd{20.4}{30.5} & 9.6 \\
    \bottomrule
  \end{tabular}
\end{table}
\noindent In Table~\ref{tab:gemini_two_solvers}, the HiGHS 2000\,s means are positive for 7/7 benchmarks and range from 5.7\% (MCND) to 64.2\% (CWLP). At 2000\,s, HiGHS is larger than Gurobi on CWLP (64.2\% vs.\ 30.1\%) and SHUC (20.4\% vs.\ 4.3\%) and smaller than Gurobi on the other 5 benchmarks.

\subsubsection{Interaction with Internal Solver Cuts}
\label{sec:internal_cut_interaction}
This check shows how much of \textsc{SAGE}'s benefit remains when the solver's own cutting-plane machinery is reduced and whether the interaction between instantiated acceleration inequality families and solver-generated cuts changes the net improvement. We solve the test instances under Gurobi with internal cuts disabled and under default internal cuts, using the same test sets, 2000\,s limit, and seven checkpoints as Table~\ref{tab:post_ga_results}. Table~\ref{tab:gemini_internal_cuts} reports \(\bar{\Delta}_g\) under each cut setting.

\begin{table}[t]
  \centering
  \scriptsize
\caption{\texttt{Gemini 3 Pro} selected acceleration inequality family: test stabilized $\bar{\Delta}_g$ (\%) with Gurobi internal cuts disabled vs.\ default. Mean $\pm\sigma$ is reported at each checkpoint. Average reports the row average of checkpoint means.}
  \label{tab:gemini_internal_cuts}
  \setlength{\tabcolsep}{2pt}
  \begin{tabular}{llccccccc@{\hspace{6pt}}|@{\hspace{6pt}}c}
    \toprule
    Problem & Solver &
    5 s & 10 s & 50 s & 150 s & 300 s & 1000 s & 2000 s & Average \\
    \midrule
    \multirow{2}{*}{TSP}
      & No internal cuts
        & \meanstd{34.9}{25.1}
        & \meanstd{37.8}{25.5}
        & \meanstd{53.5}{23.3}
        & \meanstd{61.6}{18.1}
        & \meanstd{64.2}{16.5}
        & \meanstd{64.0}{16.8}
        & \meanstd{62.0}{20.2} & 54.0 \\
      & Default internal cuts
        & \meanstd{13.1}{27.5}
        & \meanstd{15.0}{27.9}
        & \meanstd{22.6}{31.3}
        & \meanstd{34.2}{30.3}
        & \meanstd{27.6}{33.1}
        & \meanstd{26.4}{33.3}
        & \meanstd{37.0}{31.1} & 25.1 \\
    \midrule
    \multirow{2}{*}{MCND}
      & No internal cuts
        & \meanstd{2.5}{10.0}
        & \meanstd{2.9}{11.0}
        & \meanstd{4.3}{14.8}
        & \meanstd{3.4}{15.3}
        & \meanstd{1.7}{17.1}
        & \meanstd{4.1}{17.2}
        & \meanstd{5.1}{13.6} & 3.4 \\
	      & Default internal cuts
	        & \meanstd{6.8}{17.7}
	        & \meanstd{8.2}{19.9}
	        & \meanstd{15.1}{25.8}
	        & \meanstd{17.4}{26.9}
	        & \meanstd{17.8}{25.5}
	        & \meanstd{23.2}{27.9}
	        & \meanstd{15.8}{28.6} & 14.9 \\
    \midrule
    \multirow{2}{*}{CWLP}
      & No internal cuts
        & \meanstd{53.6}{19.8}
        & \meanstd{60.6}{21.2}
        & \meanstd{86.7}{19.8}
        & \meanstd{84.3}{28.3}
        & \meanstd{94.8}{8.8}
        & \meanstd{94.5}{8.9}
        & \meanstd{93.9}{9.2} & 81.2 \\
      & Default internal cuts
        & \meanstd{43.8}{20.7}
        & \meanstd{49.4}{22.3}
        & \meanstd{68.8}{28.6}
        & \meanstd{76.3}{23.8}
        & \meanstd{64.8}{23.0}
        & \meanstd{35.2}{39.3}
        & \meanstd{30.1}{38.9} & 52.6 \\
    \midrule
	    \multirow{2}{*}{JSSP}
	      & No internal cuts
	        & \meanstd{64.2}{14.6}
		      & \meanstd{62.8}{16.1}
		      & \meanstd{67.5}{20.3}
	        & \meanstd{71.7}{23.5}
		      & \meanstd{64.9}{22.7}
		      & \meanstd{69.3}{23.9}
		      & \meanstd{63.6}{19.7} & 66.3 \\
	      & Default internal cuts
	        & \meanstd{64.1}{9.9}
		      & \meanstd{63.7}{10.7}
		      & \meanstd{63.3}{14.4}
        & \meanstd{61.8}{16.6}
	      & \meanstd{63.2}{16.5}
	      & \meanstd{59.4}{21.2}
	      & \meanstd{59.2}{20.6} & 62.1 \\
    \midrule
	    \multirow{2}{*}{PDPTW}
	      & No internal cuts
	        & \meanstd{1.7}{6.2}
	        & \meanstd{2.9}{6.9}
	        & \meanstd{12.4}{16.8}
	        & \meanstd{22.5}{28.9}
	        & \meanstd{16.0}{33.5}
	        & \meanstd{27.8}{34.2}
	        & \meanstd{34.2}{32.6} & 16.8 \\
	      & Default internal cuts
	        & \meanstd{46.9}{32.4}
	        & \meanstd{47.5}{32.1}
	        & \meanstd{54.3}{29.2}
        & \meanstd{62.6}{28.6}
        & \meanstd{55.8}{35.1}
        & \meanstd{64.2}{31.2}
        & \meanstd{70.7}{29.4} & 57.4 \\
    \midrule
	    \multirow{2}{*}{IMO6}
	      & No internal cuts
	        & \meanstd{11.9}{13.6}
	        & \meanstd{13.7}{15.1}
	        & \meanstd{21.5}{19.2}
	        & \meanstd{23.4}{21.1}
	        & \meanstd{26.8}{22.5}
	        & \meanstd{20.6}{26.3}
	        & \meanstd{24.3}{26.7} & 20.3 \\
	      & Default internal cuts
	        & \meanstd{13.6}{11.2}
	        & \meanstd{15.7}{13.5}
	        & \meanstd{24.6}{18.4}
        & \meanstd{25.0}{21.4}
        & \meanstd{28.9}{23.0}
        & \meanstd{24.3}{26.8}
			        & \meanstd{19.7}{27.8} & 21.7 \\
    \midrule
	    \multirow{2}{*}{SHUC}
	      & No internal cuts
	        & \meanstd{44.3}{36.8}
	        & \meanstd{41.6}{35.9}
	        & \meanstd{28.2}{38.4}
	        & \meanstd{10.7}{31.6}
	        & \meanstd{8.2}{28.9}
	        & \meanstd{4.6}{27.8}
	        & \meanstd{3.4}{26.9} & 20.1 \\
	      & Default internal cuts
	        & \meanstd{51.2}{35.1}
	        & \meanstd{50.6}{34.8}
	        & \meanstd{32.2}{37.7}
        & \meanstd{6.3}{29.4}
        & \meanstd{3.7}{24.9}
        & \meanstd{2.6}{24.2}
        & \meanstd{4.3}{25.0} & 21.6 \\
    \bottomrule
  \end{tabular}
\end{table}
\noindent In Table~\ref{tab:gemini_internal_cuts}, disabling internal cuts gives larger \(\bar{\Delta}_g\) than default settings for TSP and CWLP at 7/7 checkpoints. MCND and PDPTW have smaller values with internal cuts disabled at 7/7 checkpoints. At 2000\,s, the no-internal-cuts and default values are 62.0\% vs.\ 37.0\% for TSP, 93.9\% vs.\ 30.1\% for CWLP, 5.1\% vs.\ 15.8\% for MCND, and 34.2\% vs.\ 70.7\% for PDPTW. JSSP, IMO6, and SHUC have mixed ordering across the 7 checkpoints. The row averages remain positive for all seven benchmarks in both settings: with cuts disabled they range from 3.4\% (MCND) to 81.2\% (CWLP), and with default cuts they range from 14.9\% (MCND) to 62.1\% (JSSP). This ablation tests cut-policy interaction: \textsc{SAGE} acceleration inequality families are added before Gurobi is called, while the cut parameter changes solver-generated cuts inside Gurobi.

\subsubsection{Sensitivity to the Continuous Solution Method}
\label{sec:method_sensitivity}
We also test sensitivity to the continuous solution method. We fix the selected TSP family and compare the automatic method with the deterministic concurrent method on TSP test instances under a limit of 300 seconds. Appendix~\ref{app:method_sensitivity} gives the protocol.

\noindent Table~\ref{tab:method_sensitivity} in Appendix~\ref{app:method_sensitivity} shows positive mean improvement under both methods. The mean is $34.3\%$ with the automatic method and $27.6\%$ with the deterministic concurrent method.

\subsection{Sensitivity to \texorpdfstring{\(|D_{imp}|\) and \(|D_{adm}|\)}{|Dimp| and |Dadm|}}
\label{sec:sensitivity_training_size}
We vary \(\lvert D_{imp}\rvert\) and \(\lvert D_{adm}\rvert\) on the JSSP benchmark using DeepSeek-R1. We run two experiments. In the first, we fix \(\lvert D_{imp}\rvert{=}10\) and vary \(\lvert D_{adm}\rvert\) over \(\{2,5,10\}\). In the second, we hold \(\lvert D_{adm}\rvert{=}10\) constant and vary \(\lvert D_{imp}\rvert\) over the same three values. These settings are independent \textsc{SAGE} reruns. In every setting, the selected acceleration inequality family found by \textsc{SAGE} is evaluated on the held-out test set \(D_t\). Following the post-hoc screening protocol in Section~\ref{sec:deployment_time_sanity}, we also check whether any generated acceleration inequality family excludes a proven-optimal JSSP solution. Table~\ref{tab:train_size} reports test-set \(\bar{\Delta}_g\ {\scriptstyle\pm\sigma}\) at the 2000\,s checkpoint for the selected acceleration inequality family found under each setting.
\begin{table}[t]
    \centering
    \caption{JSSP: test stabilized \(\bar{\Delta}_g\ {\scriptstyle\pm\sigma}\ (\%)\) at the 2000\,s checkpoint for the selected \textsc{SAGE} acceleration inequality family vs.\ \((|D_{adm}|, |D_{imp}|)\).}
    \label{tab:train_size}
    \setlength{\tabcolsep}{4pt}
    \begin{tabular}{@{}lccccc@{}}
    \toprule
    \((|D_{adm}|, |D_{imp}|)\) & \((2,10)\) & \((5,10)\) & \((10,10)\) & \((10,5)\) & \((10,2)\) \\
    \midrule
    \(\bar{\Delta}_g\ {\scriptstyle\pm\sigma}\ (\%)\) &
    37.3 {\scriptsize \(\pm\) 22.0} &
    37.3 {\scriptsize \(\pm\) 22.0} &
    40.4 {\scriptsize \(\pm\) 22.3} &
    37.3 {\scriptsize \(\pm\) 22.0} &
     7.1 {\scriptsize \(\pm\) 16.4} \\
    \bottomrule
    \end{tabular}
\end{table}
\noindent In Table~\ref{tab:train_size}, the \(|D_{adm}|=2\) and \(|D_{adm}|=5\) configurations both report 37.3\%, while \(|D_{adm}|=10\) reports 40.4\%. With \(|D_{adm}|=10\), the reported means for \(|D_{imp}|=2,5,10\) are 7.1\%, 37.3\%, and 40.4\%, respectively. The three settings that report 37.3\% are separate \textsc{SAGE} runs that reach the same test set mean. The default admissibility set is within 3.1 percentage points of the largest setting. None of the generated families under these configurations excludes a proven-optimal JSSP solution in the post-hoc check. All setups pass on 58/58 test instances.
\subsection{Evolutionary Diagnostics}
\label{sec:evo_diagnostics}

We report four diagnostics of the evolutionary process: (i) agent outcome statistics (Table~\ref{tab:agent_stat}), (ii) a traced ancestry graph for a JSSP evolutionary trajectory (Fig.~\ref{fig:evolutionary_trace} in Appendix~\ref{app:evolutionary_trace}), (iii) population score trajectories over generations (Fig.~\ref{fig:all_problems_impact_score_progress} in Appendix~\ref{app:ga_progress}), and (iv) a token matched initializer control on JSSP (Table~\ref{tab:evo_control_dev}, with the protocol and deployment results in Appendix~\ref{app:evo_control}).
\noindent For each generated offspring, we compute the relative improvement rate $\Delta_{imp} \,=\, (s_{imp,\text{child}} - s_{imp,\text{parent}}) / s_{imp,\text{parent}}$, where $s_{imp,\text{parent}}$ is the parent inequality impact score (a single value for mutation and the mean of two parents for crossover). For the initializer agent, the reference value is the neutral inequality impact score~10 that corresponds to no added acceleration inequality family. Table~\ref{tab:agent_stat} reports five statistics collected over the full run on all benchmarks: code generation failures, OSS failures, LP-separation rejections, overall success rate, and the average improvement $\bar{\Delta}_{imp}$ of successful offspring. The mutation and crossover agents have higher acceptance rates than the initializer agent across the three LLMs. The table reports mean inequality-impact-score gains \(\bar{\Delta}_{imp}\) that include both negative and positive values. The initializer acceptance rates are 25.4\%, 36.5\%, and 29.6\% for DeepSeek-R1, Gemini 3 Pro, and GPT-5.1, respectively. Across mutation agents, acceptance rates range from 63.3\% to 85.2\%. Across crossover agents, they range from 67.9\% to 84.4\%. The reported \(\bar{\Delta}_{imp}\) values range from \(-0.4\%\) to 19.1\%.

\begin{table}[t]
  \centering
  \footnotesize
  \caption{\textsc{SAGE} agent outcomes across three LLMs: code fail, OSS fail, LP-separation rejection, success rate, and mean inequality-impact-score gain $\bar{\Delta}_{imp} \pm \sigma$ (\%) (successful offspring only).}
  \label{tab:agent_stat}
  \setlength{\tabcolsep}{4pt}
  \begin{tabular}{
      l
      l
      c
      c
      c
      c
      c
    }
    \toprule
    Agent & LLM &
    Code Fail.\% &
    OSS Fail.\% &
    Not Useful \% &
    Success Rate \% &
    $\bar{\Delta}_{imp} \pm \sigma$ (\%) \\
    \midrule

    \multicolumn{7}{c}{Initializer} \\
    \midrule
    \multirow{3}{*}{Main}
      & DeepSeek      & 38.1 & 11.1 & 25.4 & 25.4 & 12.3 $\pm$ 20.8 \\
      & Gemini 3 Pro  & 5.6 & 12.8 & 45.1 & 36.5 & 14.8 $\pm$ 19.9 \\
      & GPT-5.1       & 8.9 & 14.7 & 46.8 & 29.6 & 10.4 $\pm$ 19.5 \\
    \midrule

    \multicolumn{7}{c}{Mutation} \\
    \midrule
    \multirow{3}{*}{General}
      & DeepSeek      & 17.6 & 13.2 & 2.9 & 66.2 & 15.5 $\pm$ 16.6 \\
      & Gemini 3 Pro  & 3.1 & 10.2 & 3.4 & 83.3 & 17.2 $\pm$ 15.1 \\
      & GPT-5.1       & 5.6 & 12.5 & 4.8 & 77.1 & 12.8 $\pm$ 15.3 \\
    \midrule
    \multirow{3}{*}{Exploratory}
      & DeepSeek      & 3.3 & 13.1 & 1.6 & 82.0 & 17.9 $\pm$ 20.8 \\
      & Gemini 3 Pro  & 1.4 & 11.3 & 2.1 & 85.2 & 19.1 $\pm$ 18.0 \\
      & GPT-5.1       & 3.2 & 13.4 & 2.9 & 80.5 & 13.7 $\pm$ 17.4 \\
    \midrule
    \multirow{3}{*}{Lifted}
      & DeepSeek      & 12.2 & 18.4 & 6.1 & 63.3 & 8.1 $\pm$ 13.1 \\
      & Gemini 3 Pro  & 2.8 & 15.9 & 6.4 & 74.9 & 9.6 $\pm$ 12.1 \\
      & GPT-5.1       & 4.9 & 18.2 & 8.1 & 68.8 & 6.1 $\pm$ 11.7 \\
    \midrule

    \multicolumn{7}{c}{Crossover} \\
    \midrule
    \multirow{3}{*}{Hybrid}
      & DeepSeek      & 13.2 & 15.1 & 3.8 & 67.9 & 3.0 $\pm$ 11.0 \\
      & Gemini 3 Pro  & 2.5 & 12.2 & 4.6 & 80.7 & 4.4 $\pm$ 10.2 \\
      & GPT-5.1       & 4.3 & 14.8 & 6.1 & 74.8 & 2.0 $\pm$ 9.5 \\
    \midrule
    \multirow{3}{*}{Intersect}
      & DeepSeek      & 7.7 & 9.9 & 5.5 & 76.9 & 1.3 $\pm$ 8.7 \\
      & Gemini 3 Pro  & 1.9 & 8.7 & 5.0 & 84.4 & 2.2 $\pm$ 8.1 \\
      & GPT-5.1       & 3.7 & 10.9 & 6.4 & 79.0 & 1.1 $\pm$ 7.8 \\
    \midrule
    \multirow{3}{*}{Complement}
      & DeepSeek      & 4.2 & 14.3 & 6.7 & 74.8 & 14.7 $\pm$ 13.9 \\
      & Gemini 3 Pro  & 1.6 & 11.4 & 6.9 & 80.1 & 15.6 $\pm$ 12.7 \\
      & GPT-5.1       & 3.3 & 14.2 & 8.1 & 74.4 & 10.8 $\pm$ 13.6 \\
    \midrule
    \multirow{3}{*}{Min Violation}
      & DeepSeek      & 10.1 & 11.9 & 9.2 & 68.8 & -0.4 $\pm$ 9.6 \\
      & Gemini 3 Pro  & 2.3 & 10.8 & 9.1 & 77.8 & 0.6 $\pm$ 8.7 \\
      & GPT-5.1       & 4.1 & 13.6 & 10.2 & 72.1 & -0.3 $\pm$ 9.0 \\
    \bottomrule
  \end{tabular}
\end{table}

\noindent Second, we trace one JSSP evolutionary trajectory. This trajectory is included to illustrate how \textsc{SAGE} mutates and recombines candidates. Fig.~\ref{fig:evolutionary_trace} in Appendix~\ref{app:evolutionary_trace} shows the traced graph for this trajectory. Third, we plot score curves at the population level. For each of the seven problems, we tracked the maximum and mean inequality impact scores of the population over 20 generations. Fig.~\ref{fig:all_problems_impact_score_progress} in Appendix~\ref{app:ga_progress} shows the trajectories of these values across generations. For each benchmark, the best score curve increases over the 20 generations. The initial population is the output of the initializer agent before any crossover or mutation. On every benchmark, the best score of the final generation exceeds the best score of the initial population, for example from 9.4 to 13.7 on TSP and from 10.2 to 14.7 on PDPTW (Appendix~\ref{app:ga_progress}).

\noindent Fourth, we compare five generations of \textsc{SAGE} with iterative initialization and feedback in one paired JSSP experiment using DeepSeek-R1. Both runs share the initial population and development data, with token usage after initialization matched to within 0.51\%. Table~\ref{tab:evo_control_dev} reports the best impact score and the corresponding mean development gap reduction, $-\bar d=\ln(s_{imp}/10)$. The mean development gap reduction is 64.6\% for \textsc{SAGE} and 56.5\% for the control. Appendix~\ref{app:evo_control} gives the protocol.

\begin{table}[t]
  \centering
  \small
  \caption{JSSP with DeepSeek-R1: development comparison between five generations of \textsc{SAGE} and the initializer with feedback, with token usage matched to within 0.51\% after the shared initialization. Appendix~\ref{app:evo_control} gives the protocol and the deployment results.}
  \label{tab:evo_control_dev}
  \setlength{\tabcolsep}{3pt}
  \begin{tabular}{lrrr}
    \toprule
    Run & Tokens & Best $s_{imp}$ & $-\bar d$ (\%) \\
    \midrule
    \textsc{SAGE} (5 generations) & 761,451 & 19.07 & 64.6 \\
    Initializer with feedback & 765,308 & 17.60 & 56.5 \\
    \bottomrule
  \end{tabular}
\end{table}

\noindent Table~\ref{tab:evo_control_dev} shows that \textsc{SAGE} raises the best score from 17.60 to 19.07 (a mean development gap reduction of 64.6\%), while the control remains at 17.60 (56.5\%). The selected \textsc{SAGE} family gives a strictly larger makespan bound on 38 test instances and an equal bound on the remaining 42, and both families preserve all stored proven optimal solutions (Appendix~\ref{app:evo_control}).

\noindent Appendix~\ref{app:evo_control}, Table~\ref{tab:evo_control}, evaluates both selected families against the same baseline on the JSSP test set. \textsc{SAGE} has a higher mean relative gap improvement at all seven checkpoints, with paired differences of 15.9 percentage points at 5\,s and 9.6 at 10\,s. The pointwise 95\% confidence intervals are above zero at 5, 10, 50, and 300\,s.

\subsection{Case Study: Post-Hoc Interpretation of the SHUC Acceleration Inequality}
\label{sec:shuc_case_study}

We next examine the selected acceleration inequality family for the sub-hour unit commitment problem (SHUC). This is a post-hoc case study for one acceleration inequality family found by \textsc{SAGE}, not a general validity guarantee for the method. \textsc{SAGE} selected this family through the empirical screens in Section~\ref{sec:solution_method}. The proof below was written after selection to interpret this particular output. The selected family combines several familiar unit-commitment strengthening motifs into one aggregate reachability inequality. The SHUC benchmark follows a reference unit commitment formulation with piecewise production costs, startup categories, ramping limits, minimum up/down times, demand balance, and spinning reserve requirements \citep{knueven2020mixed,powergridlib2019pglib}. For the SHUC formulation in Appendix~\ref{app:shuc}, this selected acceleration inequality family can be proved valid as a valid inequality family.

To avoid overloading notation, we denote the auxiliary capacity variable from Appendix~\ref{app:shuc} by $\bar P_{g,t}$. Define
\[
D_t^{\mathrm{net}} := L_t-\sum_{w\in W}p^w_{w,t}+R_t,
\qquad
a_g^{\mathrm{SU}} := \max\{P_g^{\max}-SU_g,0\},
\qquad
a_g^{\mathrm{SD}} := \max\{P_g^{\max}-SD_g,0\}.
\]
The selected SHUC acceleration inequality family is:
\begin{align}
b_{g,t} &\le v_{g,t},
&& \forall g\in G,\; t<T, \label{eq:shuc-b-linearization}\\
b_{g,t} &\le w_{g,t+1},
&& \forall g\in G,\; t<T, \notag\\
b_{g,t} &\ge v_{g,t}+w_{g,t+1}-1,
&& \forall g\in G,\; t<T, \notag\\
\bar P_{g,t}
&\le P_g^{\max}u_{g,t}-a_g^{\mathrm{SU}}v_{g,t},
&& \forall g\in G,\; t\in \mathcal{T}, \label{eq:shuc-su-cap}\\
\bar P_{g,t}
&\le P_g^{\max}u_{g,t}-a_g^{\mathrm{SD}}w_{g,t+1},
&& \forall g\in G,\; t<T, \label{eq:shuc-sd-cap}\\
\bar P_{g,t}
&\le P_g^{\max}u_{g,t}
-a_g^{\mathrm{SU}}v_{g,t}
-a_g^{\mathrm{SD}}w_{g,t+1}
+\min\{a_g^{\mathrm{SU}},a_g^{\mathrm{SD}}\}b_{g,t},
&& \forall g\in G,\; t<T, \label{eq:shuc-lifted-susd-cap}\\
\bar P_{g,t}
&\le P_g^{\min}u_{g,t}+p_{g,t-1}+RU_g
+(P_g^{\max}-P_g^{\min})(1-u_{g,t-1}),
&& \forall g\in G,\; t\ge 2, \label{eq:shuc-ramp-p}\\
\bar P_{g,t}
&\le P_g^{\min}u_{g,t}
+(P_g^{\max}-P_g^{\min})u_{g,t-1}
-a_g^{\mathrm{SU}}v_{g,t-1}+RU_g,
&& \forall g\in G,\; t\ge 2, \label{eq:shuc-ramp-su}\\
\bar P_{g,t}
&\le P_g^{\min}u_{g,t}
+(P_g^{\max}-P_g^{\min})u_{g,t-1}
-a_g^{\mathrm{SD}}w_{g,t}+RU_g,
&& \forall g\in G,\; t\ge 2, \label{eq:shuc-ramp-sd}\\
D_t^{\mathrm{net}} &\le \sum_{g\in G}\bar P_{g,t},
&& \forall t\in \mathcal{T}. \label{eq:shuc-aggregate-reachability}
\end{align}
The analogous $t=1$ ramp-reachability rows use the initial status and initial output fields from the instance data, as in the base SHUC model.

\begin{proposition}[Post-hoc validity of the selected SHUC acceleration inequality family by \textsc{SAGE}]
\label{prop:shuc_validity}
For the SHUC formulation in Appendix~\ref{app:shuc}, the acceleration inequality family from Eq.~\eqref{eq:shuc-b-linearization} through Eq.~\eqref{eq:shuc-aggregate-reachability} is valid as an extended inequality family: every feasible integer solution of the original SHUC formulation can be extended with auxiliary variables $(\bar P,b)$ so that all added constraints are satisfied.
\end{proposition}

\begin{proof}
Take any feasible integer solution of the original SHUC formulation. Set
\[
\widehat P_{g,t}:=P_g^{\min}u_{g,t}+p_{g,t}+r_{g,t},
\qquad
\widehat b_{g,t}:=v_{g,t}w_{g,t+1}.
\]
Because $v_{g,t}$ and $w_{g,t+1}$ are binary, the three inequalities in Eq.~\eqref{eq:shuc-b-linearization} are the standard linearization of the product $b_{g,t}=v_{g,t}w_{g,t+1}$.

The base SHUC startup and shutdown capacity rows give
\[
p_{g,t}+r_{g,t}
\le (P_g^{\max}-P_g^{\min})u_{g,t}-a_g^{\mathrm{SU}}v_{g,t}
\]
and
\[
p_{g,t}+r_{g,t}
\le (P_g^{\max}-P_g^{\min})u_{g,t}-a_g^{\mathrm{SD}}w_{g,t+1}.
\]
Adding $P_g^{\min}u_{g,t}$ to both inequalities proves Eq.~\eqref{eq:shuc-su-cap} and Eq.~\eqref{eq:shuc-sd-cap} for $\bar P_{g,t}=\widehat P_{g,t}$. Moreover, when $v_{g,t},w_{g,t+1}\in\{0,1\}$ and $b_{g,t}=v_{g,t}w_{g,t+1}$, the right-hand side of Eq.~\eqref{eq:shuc-lifted-susd-cap} equals the tighter of the two startup/shutdown upper envelopes. Hence Eq.~\eqref{eq:shuc-lifted-susd-cap} is also satisfied.

For the ramp-reachability rows, the base ramp-up constraint gives
\[
p_{g,t}+r_{g,t}-p_{g,t-1}\le RU_g.
\]
Adding $P_g^{\min}u_{g,t}$ yields
\[
\widehat P_{g,t}\le P_g^{\min}u_{g,t}+p_{g,t-1}+RU_g,
\]
which implies Eq.~\eqref{eq:shuc-ramp-p}. Combining the same ramp-up constraint with the startup-capacity upper bound at time $t-1$ gives
\[
p_{g,t-1}
\le (P_g^{\max}-P_g^{\min})u_{g,t-1}
-a_g^{\mathrm{SU}}v_{g,t-1},
\]
and therefore Eq.~\eqref{eq:shuc-ramp-su}. Combining the ramp-up constraint with the shutdown-capacity upper bound at time $t-1$ gives
\[
p_{g,t-1}
\le (P_g^{\max}-P_g^{\min})u_{g,t-1}
-a_g^{\mathrm{SD}}w_{g,t},
\]
and therefore Eq.~\eqref{eq:shuc-ramp-sd}.

Finally, the base power-balance constraint gives
\[
L_t-\sum_{w\in W}p^w_{w,t}
=
\sum_{g\in G}\left(P_g^{\min}u_{g,t}+p_{g,t}\right),
\]
and the base reserve constraint gives
\[
R_t\le \sum_{g\in G}r_{g,t}.
\]
Adding these two relations gives
\[
D_t^{\mathrm{net}}
=
L_t-\sum_{w\in W}p^w_{w,t}+R_t
\le
\sum_{g\in G}\left(P_g^{\min}u_{g,t}+p_{g,t}+r_{g,t}\right)
=
\sum_{g\in G}\widehat P_{g,t},
\]
which proves Eq.~\eqref{eq:shuc-aggregate-reachability}. Thus every feasible integer solution of the base SHUC model admits feasible auxiliary variables, so the added family is valid in the extended space.
\end{proof}

\begin{table}[H]
\centering
\small
\caption{Post-hoc source attribution for the selected SHUC acceleration inequality family. The table categorizes each component by whether it has a direct literature analog, follows from the base formulation, or appears to be a formulation-specific aggregate component.}
\label{tab:shuc_source_attribution}
\setlength{\tabcolsep}{4pt}
\begin{tabular}{p{0.18\linewidth}p{0.34\linewidth}p{0.42\linewidth}}
\toprule
Component & What it does & Source status \\
\midrule
\litcell Binary product rows Eq.~\eqref{eq:shuc-b-linearization}
&
\litcell Encode $b_{g,t}=v_{g,t}w_{g,t+1}$, i.e., whether unit $g$ starts at $t$ and shuts down at $t+1$.
&
\litcell Standard MILP linearization of binary products, not SHUC-specific \citep{glover1975improved}. \\
\midrule
\litcell Startup/shutdown capacity envelopes Eq.~\eqref{eq:shuc-su-cap} to Eq.~\eqref{eq:shuc-sd-cap}
&
\litcell Upper-bound reachable thermal capacity when the unit is starting up or shutting down.
&
\litcell Closely aligned with startup/shutdown ramping and strong UC production upper bounds \citep{moralesespana2013startup,knueven2020mixed}. \\
\midrule
\litcell Lifted adjacent start/shutdown envelope Eq.~\eqref{eq:shuc-lifted-susd-cap}
&
\litcell Combines the two derating envelopes and uses $b_{g,t}$ to avoid double-counting when both transition indicators are active.
&
\litcell Built from known lifting and UC derating ideas. Closest references are strong UC upper-bound and ramping formulations \citep{knueven2018ramping,knueven2020mixed}. \\
\midrule
\litcell Ramp-reachability envelopes Eq.~\eqref{eq:shuc-ramp-p} to Eq.~\eqref{eq:shuc-ramp-sd}
&
\litcell Bound what unit $g$ can contribute at $t$ using previous dispatch/status and ramp-up capability.
&
\litcell Related to ramping polytope and ramp-aware UC formulations \citep{knueven2018ramping,knueven2020mixed}. \\
\midrule
\sagecell Aggregate auxiliary reachability inequality Eq.~\eqref{eq:shuc-aggregate-reachability}
&
\sagecell Projects unit-level startup, shutdown, and ramp-reachability envelopes into one system-level condition: net load plus reserve must be coverable by reachable thermal capacity.
&
\sagecell In our targeted review, close analogs exist for the unit-level components in the UC formulation literature, including reserve and ramp-capability formulations \citep{moralesespana2014rampreserves,moralesespana2016powercapacity}, but we did not identify Eq.~\eqref{eq:shuc-aggregate-reachability} as this same static aggregate inequality added before solving for the SHUC formulation in Appendix~\ref{app:shuc}. \\
\bottomrule
\end{tabular}
\end{table}

Table~\ref{tab:shuc_source_attribution} shows that this selected SHUC acceleration inequality family is interpretable at two levels. At the unit level, \textsc{SAGE} recovers known modeling motifs: binary-product linearization, transition derating, and ramp-aware capacity envelopes. At the system level, it composes these motifs into the aggregate reachability inequality in Eq.~\eqref{eq:shuc-aggregate-reachability}. Our targeted review did not identify this exact static aggregate inequality for SHUC. Rather than adding a runtime separator or a full ramping polytope, this selected acceleration inequality family can be appended before the solver is called.

\section{Discussion}

\noindent Starting with RQ1, Table~\ref{tab:post_ga_results} and Appendix~\ref{app:gap_improvement_curves} show that \textsc{SAGE} can reduce the optimality gap at successive checkpoints under the same time budgets. The trajectory checks in Appendix~\ref{app:tsp_examples}, Appendix~\ref{app:tsp_best_bound_exp}, and Appendix~\ref{app:tsp_pdi_exp} show the same effect for individual instances. Table~\ref{tab:post_ga_results} also indicates that different LLMs lead to different families and levels of improvement. Therefore, choosing an LLM among the multiple LLM options of \textsc{SAGE} matters. The main LLM result is that Gemini 3 Pro has been generally the strongest backend in our experiments. This pattern justifies our use of the Gemini 3 Pro acceleration inequality family as the selected alternative in the rest of the experiments. Complementing this, Table~\ref{tab:post_ga_time} shows faster times to reach target gaps across benchmarks. Table~\ref{tab:selected_family_reporting} in Appendix~\ref{app:model_size_lp} shows that the selected families tighten the LP relaxation on nearly every test instance, so part of the deployment gain may come from a stronger initial relaxation, while the amount of tightening and the added model size vary by benchmark. \textsc{SAGE} often increases the number of instances that reach the target gap within the time limit. We have also observed on rare occasions that \textsc{SAGE} acceleration inequality families may have negative impact. When a discovered acceleration inequality family fails to improve the relaxed model for a particular instance, it increases model size and overhead nonetheless, leading to a slight deterioration of the overall performance (Appendix~\ref{app:mcnd_gap_failure}).

\noindent To test robustness beyond gap improvements, we reported on four complementary experiments. First, the post-hoc screening against proven-optimal solutions and the held-out LP separation check in Section~\ref{sec:deployment_time_sanity} indicate a reduced chance that the observed speedups come from excluding good integer solutions. Second, the cross-solver ablation in Section~\ref{sec:cross_solver_transfer} suggests that the same discovered families can transfer beyond one solver. Third, the internal-cut ablation in Section~\ref{sec:internal_cut_interaction} shows that \textsc{SAGE} remains useful when cut generation inside Gurobi changes. Fourth, the learned-policy factorial in Section~\ref{sec:ml_policy_complementarity} evaluates the same modeling claim against learned cut-selection and separator-configuration policies. The learned-policy results are consistent with complementarity between \textsc{SAGE}'s model-layer augmentation and the learned solver-layer policies. Together, these results suggest that SAGE provides a static model augmentation at deployment after a separate discovery stage. This augmentation is separate from solver cut policies and learned runtime policies because it changes the formulation before the solver is called, while those policies act during solving. We also observe that the magnitude of improvement depends on the interaction between the compact formulation, the selected acceleration inequality family, and the solver. Mature benchmarks can still show large gains when the compact baseline leaves useful model structure. Finally, we focus on improving a given symbolic formulation without callbacks or domain-specific separation code. Accordingly, our baselines use standard compact formulations (Appendix~\ref{app:benchmarks}) rather than a tailored branch-and-bound method for each benchmark problem.

\noindent The discovery and payback analysis in Section~\ref{sec:time_to_target} separates the search cost from repeated deployment. The discovery stage carries the cost that is paid once per problem class, while the added cost at each deployment is the time to instantiate the family for one instance. \textsc{SAGE} needs no callback, no separation routine, and no change to the solver, so the augmented model enters a standard solver pipeline. The results show that the search cost can be amortized across repeated deployments, but the required number varies by problem class. The same selected families also hold across the solvers in Section~\ref{sec:cross_solver_transfer} and the internal cut settings in Section~\ref{sec:internal_cut_interaction}, so one search covers more than one deployment configuration. The tested distribution, solver, hardware, target gap, time limit, and search budget determine the estimate of payback in compute time. API prices determine the separate monetary search cost.

\noindent The method sensitivity check in Section~\ref{sec:method_sensitivity} shows that the selected family improves solver progress under both methods, with the magnitude of the gain depending on the method.

\noindent The static literature augmentation comparison in Section~\ref{sec:literature_cut_comparison} and Appendix~\ref{app:literature_cuts} clarifies how \textsc{SAGE}'s model-layer gains compare with existing expert valid inequalities. Valid inequalities can be strong under the same static augmentation protocol, as shown in Table~\ref{tab:literature_cut_baselines} in Appendix~\ref{app:literature_cuts}. The \textsc{SAGE} acceleration inequality families, however, yield even larger mean speedups than the instantiated static literature valid inequalities on CWLP, JSSP, and TSP. On MCND both methods are essentially equivalent. Taken together, our analysis suggests that \textsc{SAGE} performs symbolic search for useful acceleration inequality families at the modeling level. The selected \textsc{SAGE} families include both familiar forms of inequalities and more specialized algebraic forms. This search can recover high-value structure that expert formulations of classic problems also exploit, including cut-set capacity, assignment and opening links, cover logic, ordering constraints, and workload bounds. For TSP, the smaller margin is consistent with the literature baseline already adding lifted MTZ ordering rows that overlap with the selected \textsc{SAGE} family. For the lesser explored benchmarks, such as IMO6 and SHUC, the same search can produce more specialized families tied to the symbolic formulation. Thus, \textsc{SAGE} is a symbolic search procedure whose empirical screening determines which acceleration inequality family is useful for the given compact formulation.

\noindent The anonymized RNTD control in Appendix~\ref{app:rntd_control} makes direct recall of a published formulation or its specific inequalities less plausible. Together, RNTD and IMO6 distinguish formulation specific exposure from general optimization knowledge. Their unpublished or newly formulated MILPs reduce cues for direct recall while leaving general optimization knowledge available to the LLM.

\noindent Turning to RQ2, Table~\ref{tab:train_size} suggests that small admissibility sets may suffice, while larger impact-assessment sets generally yield stronger families. The default admissibility set is the smallest one we tested, and on JSSP it still yields a family that performs close to the largest setting, so a small admissibility set is enough to produce a strong family. \textsc{SAGE} derives acceleration inequality families from underlying problem logic rather than patterns specific to a dataset. In our experiments, these families remain effective even when the test distribution or size differs from those used for admissibility screening and impact assessment. This is consistent with our choice of split: $D_{imp}$ and $D_{adm}$ are synthetic, while $D_t$ is a held-out public benchmark set. As a practical consideration, the sizes of $D_{imp}$ and $D_{adm}$ create a tradeoff between runtime and quality. Larger impact-assessment sets tend to improve deployment performance. Larger admissibility sets can also help, but to a lesser extent. Both increase \textsc{SAGE} development runtime because each candidate family is tested on more instances (Section~\ref{sec:sensitivity_training_size}).

\noindent For RQ3, the paired JSSP experiment in Section~\ref{sec:evo_diagnostics}, Table~\ref{tab:evo_control_dev}, shows that evolutionary search improves on the initializer's best development score at matched token usage. The bound comparison in Appendix~\ref{app:evo_control} shows how the evolved family strengthens the initializer's makespan bound. The deployment results in Table~\ref{tab:evo_control} show that this refinement also improves solver progress, especially at early checkpoints. The acceptance statistics in Table~\ref{tab:agent_stat}, the ancestry trace in Fig.~\ref{fig:evolutionary_trace}, and the score trajectories in Fig.~\ref{fig:all_problems_impact_score_progress} describe how the search develops these families.

\section{Conclusion and Future Direction}
\textsc{SAGE} automates acceleration inequality family generation, a step that augments a MILP model before solving and is typically driven by experts. Across seven MILP benchmarks, \textsc{SAGE} discovers reusable and interpretable families, ranging from expert strengthening patterns on mature benchmarks to more specialized families on less standard models. These families can be appended to MILPs in standard solver pipelines and often improve solver progress under time limits. The search runs once per problem class, and Section~\ref{sec:time_to_target} reports its cost and the number of deployment solves that recover it. We expect the approach to be particularly useful for custom or newly formulated MILPs, where established formulation ideas may be limited.

\noindent Two design choices define the present scope of \textsc{SAGE}. Empirical screening evaluates solution preservation and LP separation on sampled instances, while formal global guarantees require a separate proof. The current method instantiates acceleration inequality families before solving. Runtime separation and dynamic addition remain future extensions.

\noindent This scope motivates two directions for future work: (1) The current screening relies on recorded solutions on a small admissibility set. Integrating automated formal proof tooling could provide theoretical guarantees (e.g., discovering acceleration inequality families composed of provably valid inequalities or families that provably preserve optimality) \cite{yang2025position}. (2) \textsc{SAGE} currently inserts instantiated acceleration inequality families into the MILP model before the solution process begins. Dynamically separating and adding inequalities via callbacks during the solve could further improve solver performance, but requires generating an efficient separation algorithm for each discovered family.

\subsubsection*{Broader Impact Statement}
\textsc{SAGE} is a study of acceleration inequality families for MILP solving. It uses benchmark and synthetic MILP instances and no human subject data. By improving solver progress under fixed budgets, the method may reduce compute time and energy use, but it may also improve the performance of downstream optimization systems in high-stakes domains such as energy, logistics, scheduling, and public-sector resource allocation. The main technical risk is that empirically screened acceleration inequality families are not guaranteed to be globally valid or optimality preserving on all unseen instances. Users should validate any generated family in the target domain, keep a baseline fallback, monitor for regressions, and not treat empirical screening as proof of validity. Releasing reproducible evaluation code and synthetic generators helps others inspect the workflow and test generated families under their own assumptions.

\subsubsection*{Acknowledgments}
We thank the action editor and the reviewers for their detailed comments and suggestions.

\bibliography{refs}
\bibliographystyle{tmlr}

\newpage
\appendix
\makeatletter
\renewcommand*{\theHequation}{\theHsection.\arabic{equation}}
\makeatother
\onecolumn

\section{{\rmfamily\mdseries\textsc{SAGE}} Pseudocode}
\label{sec:llm_evocut}
Algorithm~\ref{alg:evocut} outlines the full \textsc{SAGE} procedure.

\begin{algorithm}
\caption{Pseudocode of \textsc{SAGE}}
\label{alg:evocut}
\begin{algorithmic}[1]

\REQUIRE (i) Baseline MILP Eq.~\eqref{eq:MILP}, (ii) MILP data instances for impact assessment and admissibility, and (iii) hyperparameters: population size $\mu$, crossover rate $\mathbb{P}_c$, mutation rate $\mathbb{P}_m=1-\mathbb{P}_c$, elitism ratio $r_e$, number of generations $T$, and maximum attempts for a failed generation or admissibility-screening cycle.

\ENSURE A set of discovered acceleration inequality families with corresponding inequality impact scores.

\STATE Phase 1: Data Preprocessing
\STATE Set up $D_{imp} = \{\,i \mapsto \mathrm{gap}_{\mathrm{ref}}(i)\,\}$, $D_{adm} = \{\,i \mapsto ((\hat x_{i}^{*}, \hat y_{i}^{*}), (x_{i}^{\mathrm{LP}},y_{i}^{\mathrm{LP}}), \mathrm{gap}_{\mathrm{ref}}(i))\,\}$.

\STATE Phase 2: Population Initialization
\STATE $\mathcal{P}_1 \gets \emptyset$.
\WHILE{$|\mathcal{P}_1| < \mu$}
  \STATE Call initializer LLM $\rightarrow$ propose candidate acceleration inequality family $C$.
  \STATE $s_{imp}(C) \gets \mathrm{ScreenAndScoreInequality}(C)$.
  \IF{$s_{imp}(C) \neq \texttt{reject}$}
    \STATE $\mathcal{P}_1 \gets \mathcal{P}_1 \cup \{C\}$.
  \ELSE
    \STATE Provide feedback prompt to LLM and retry (up to max attempts).
  \ENDIF
\ENDWHILE

\STATE Phase 3: Evolution
\FOR{$t \gets 1$ to $T$}
  \STATE Elitism: transfer top $\lceil r_e \mu \rceil$ from $\mathcal{P}_t$ to $\mathcal{P}_{t+1}$.
  \WHILE{$|\mathcal{P}_{t+1}| < \mu$}
    \STATE Select two parents $(C_p, C_q)$ from $\mathcal{P}_t$ using tournament selection based on inequality impact score.
    \STATE Draw $u \sim \mathrm{Uniform}(0,1)$.
    \IF{$u < \mathbb{P}_c$}
      \STATE \textit{Agent Selection:} randomly choose a crossover agent from the available pool.
      \STATE Call the selected Crossover LLM with $(C_p, C_q)$ $\rightarrow C_o$.
      \STATE $s_{imp}(C_o) \gets \mathrm{ScreenAndScoreInequality}(C_o)$.
      \IF{$s_{imp}(C_o) \neq \texttt{reject}$}
        \STATE $\mathcal{P}_{t+1} \gets \mathcal{P}_{t+1} \cup \{C_o\}$.
        \IF{$|\mathcal{P}_{t+1}| = \mu$}
          \STATE break
        \ENDIF
      \ELSE
        \STATE Provide feedback prompt to the selected crossover LLM and retry (up to max attempts).
      \ENDIF
    \ELSE
      \STATE \textit{Agent Selection:} randomly choose a mutation agent from the available pool.
      \STATE Call the selected Mutation LLM with one parent $C_p$ $\rightarrow C_m$.
      \STATE $s_{imp}(C_m) \gets \mathrm{ScreenAndScoreInequality}(C_m)$.
      \IF{$s_{imp}(C_m) \neq \texttt{reject}$}
        \STATE $\mathcal{P}_{t+1} \gets \mathcal{P}_{t+1} \cup \{C_m\}$.
      \ELSE
        \STATE Provide feedback prompt to the selected mutation LLM and retry (up to max attempts).
      \ENDIF
    \ENDIF
  \ENDWHILE
\ENDFOR

\RETURN final population $\mathcal{P}_{T+1}$.

\end{algorithmic}
\end{algorithm}

\noindent In Algorithm~\ref{alg:evocut}, \(\mathrm{ScreenAndScoreInequality}(C)\) denotes the local routine that first runs \(\mathrm{InequalityAdmissibilityScreen}(C)\) on \(D_{adm}\). If any admissibility check fails, the routine returns \texttt{reject}. Otherwise, \(\mathrm{InequalityImpactScorer}(C)\) computes and returns the inequality impact score \(s_{imp}(C)\) on \(D_{imp}\).

\noindent The reduced model fixes every original variable and leaves only the auxiliary variables of the candidate family free. The OSS check accepts a candidate when this model is feasible for every stored reference solution. The LP separation check accepts it when the corresponding model is infeasible for at least one stored LP solution. A code, OSS, or LP separation failure produces a diagnostic prompt for a retry.

\section{{\rmfamily\mdseries\textsc{SAGE}} Agents}
\label{app:agents_prompts}

This section outlines the agents used in the \textsc{SAGE} process and presents their prompt structures, including roles, tasks, requirements, and inputs/outputs. The agent library includes a single initializer and several Mutation and Crossover agents. Consistent with the framing in Section~\ref{sec:solution_method}, these prompts seek useful acceleration inequality families, while the Inequality Admissibility Screen later decides which candidates survive based on empirical admissibility checks. The prompt structure for the initializer agent is as follows:

\begin{tcolorbox}[breakable, title=initializer agent, colback=blue!5, colframe=blue!30!black]
Role: You are an MILP optimization expert with extensive knowledge in proposing acceleration inequality families for MILP models.

Task:
\begin{itemize}[leftmargin=1.5em]
    \item Propose a useful acceleration inequality family that is likely to improve solver performance.
    \item Provide a clear, concise explanation of the inequality intuition and its expected impact.
\end{itemize}

Requirements:
\begin{itemize}[leftmargin=1.5em]
    \item Do not assume a formal proof of global validity is available.
    \item The proposal should be broadly reusable across instances of the symbolic MILP formulation.
    \item Introduce new variables or constructs if needed.
    \item The idea should be conceptually distant from previous ideas.
\end{itemize}

Input:
\begin{itemize}[leftmargin=1.5em]
    \item < \textit{Baseline MILP (Pyomo model code)} >.
    \item < \textit{List of previous ideas (if applicable)} >.
\end{itemize}
Output:
\begin{lstlisting}[style=promptjson]
{"code": "<Only python code snippet for the added inequality family using Pyomo syntax>",
"idea": "Concise technical explanation of the inequality intuition and expected impact"}
\end{lstlisting}
\end{tcolorbox}

We employ three mutation agents, General Mutation, Lifted Mutation, and Exploratory Mutation, each of which follows the prompt structure described below.

\begin{tcolorbox}[breakable, title=Mutation Agents, colback=blue!5, colframe=blue!30!black]
Role: MILP optimization expert tasked with analyzing and modifying an individual's acceleration inequality family in an Evolutionary Algorithm for MILP acceleration inequality family generation.

Task:
\begin{itemize}[leftmargin=1.5em]
    \item Propose a useful acceleration inequality family by building on the idea behind the provided individual family.
    \item Provide a concise explanation of the inequality intuition and expected impact.
    \item <\textit{Agent specific instruction}>
\end{itemize}

Requirements:
\begin{itemize}[leftmargin=1.5em]
    \item Do not assume a formal proof of global validity is available.
    \item The new family should improve on the provided input family in spirit, structure, or empirical promise.
    \item Introduce new variables if needed.
\end{itemize}

Input:
\begin{itemize}[leftmargin=1.5em]
    \item < \textit{Baseline MILP (Pyomo model code)} >.
    \item < \textit{Input inequality family (code, idea, and score)} >.
\end{itemize}
Output:
\begin{lstlisting}[style=promptjson]
{"code": "<Only python code snippet for the added inequality family using Pyomo syntax>",
"idea": "Concise technical explanation of the inequality intuition and expected impact"}
\end{lstlisting}
\end{tcolorbox}

Agent specific instruction: Each mutation agent performs the same core task, with the following variations:
\begin{itemize}
    \item \textit{General Mutation}: Proposes a new family based on the individual family while modifying its structure or coefficients.
    \item \textit{Lifted Mutation}: Enhances the family by borrowing lifting-style ideas or auxiliary structure when useful.
    \item \textit{Exploratory Mutation}: Generates an exploratory family that diverges from the provided family to explore new regions of the design space.
\end{itemize}

Similarly, \textsc{SAGE} includes four crossover agents (Intersection, Complementary, Hybrid, and Min Violation) with the following prompt design.
\begin{tcolorbox}[breakable, title=Crossover Agents, colback=blue!5, colframe=blue!30!black]
Role: MILP optimization expert tasked with performing crossover between two parent acceleration inequality families.

Task:
\begin{itemize}[leftmargin=1.5em]
    \item Generate a useful acceleration inequality family by combining elements from the parent families.
    \item Provide a concise explanation of the new family's idea and expected impact.
    \item <\textit{Agent specific instruction}>
\end{itemize}

Requirements:
\begin{itemize}[leftmargin=1.5em]
    \item Do not assume a formal proof of global validity is available.
    \item Combine elements from both parent families in a distinct and effective way.
    \item Introduce new variables or auxiliary constructs if needed.
\end{itemize}

Input:
\begin{itemize}[leftmargin=1.5em]
    \item < \textit{Baseline MILP (Pyomo model code)} >.
    \item < \textit{First parent inequality family (code, idea, and score)} >.
    \item < \textit{Second parent inequality family (code, idea, and score)} >.
\end{itemize}

Output:
\begin{lstlisting}[style=promptjson]
{"code": "<Only python code snippet for the added inequality family using Pyomo syntax>",
"idea": "Concise technical explanation of the inequality intuition and expected impact"}
\end{lstlisting}
\end{tcolorbox}

Agent specific instruction: Each crossover agent performs the same core task, with the following variations:

\begin{itemize}
    \item Intersection Crossover: Combines elements of both parents to generate a family that preserves useful structural motifs from each.
    \item Complementary Crossover: Generates a family that complements both parents, creating a more distinct solution.
    \item Hybrid Crossover: Combines structural elements from one parent with numerical or conditional features from the other.
    \item Min Violation Crossover: Selects a crossover that minimizes the joint violation of both parents in previous runs.
\end{itemize}

\section{Detailed Evolutionary Diagnostics}
\label{app:evo}

\subsection{Evolutionary Generation of an Acceleration Inequality Family}
\label{app:evolutionary_trace}

\begin{figure}[H]
  \centering
  \includegraphics[width=0.95\textwidth]{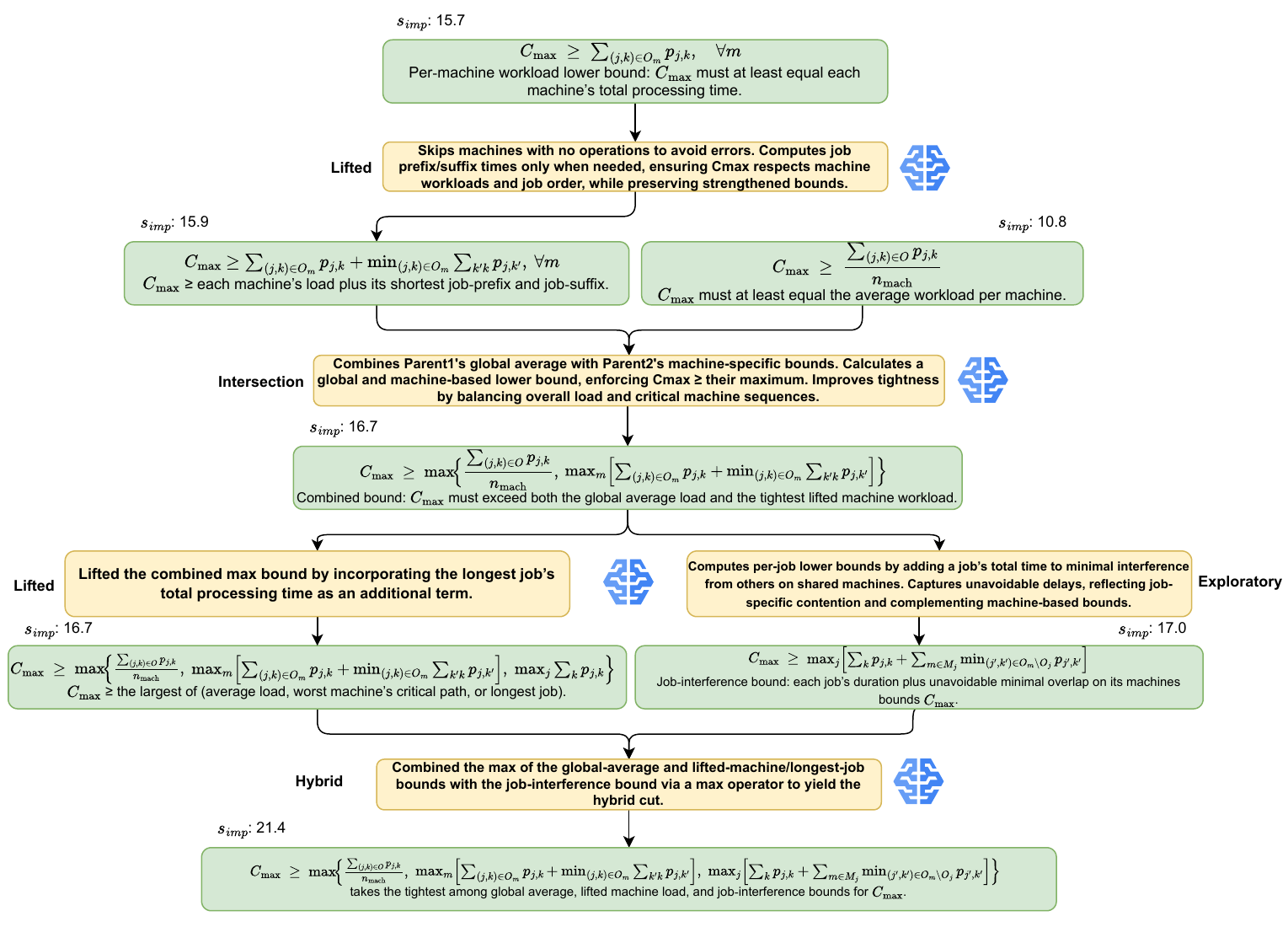}
\caption{Example JSSP evolutionary trajectory. The green panels show candidate families with annotated inequality impact scores, and the yellow panels show the agent instructions that produced each offspring.}
  \Description{An ancestry diagram showing JSSP candidate families, inequality impact scores, and agent instructions.}
  \label{fig:evolutionary_trace}
\end{figure}

Fig.~\ref{fig:evolutionary_trace} shows how \textsc{SAGE} with DeepSeek-R1 mutates and recombines candidate acceleration inequality families to obtain a strong family for the JSSP. Green boxes list candidate families with their inequality impact scores. Arrows connect parents to offspring, and yellow panels quote the agent instructions that produced the offspring.

\noindent Hybrid Crossover combines parents with impact scores of 16.7 and 17.0 and produces a family with a score of 21.4.

\subsection{Population-Level Inequality Impact Score Progress}
\label{app:ga_progress}

\begin{figure}[htbp]
    \centering
    \includegraphics[width=1.0\textwidth]{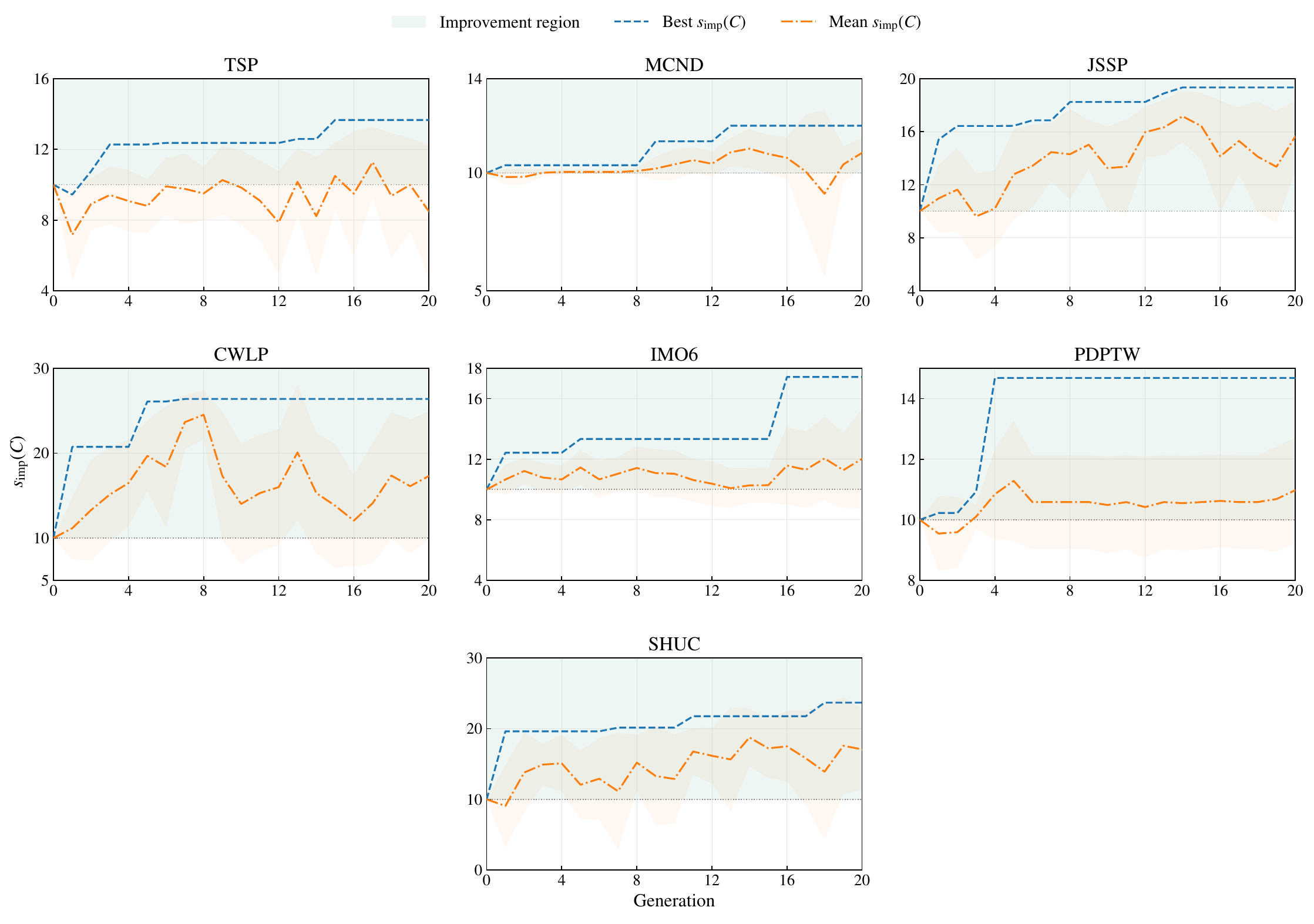}
    \caption{Best (blue, dashed) and mean (orange, dash-dotted) inequality impact scores across 20 generations for the TSP, MCND, CWLP, JSSP, IMO6, PDPTW, and SHUC benchmarks. Shaded regions show one standard deviation.}
    \Description{Line plots of best and mean inequality impact scores over 20 generations for seven benchmark problems.}
    \label{fig:all_problems_impact_score_progress}
\end{figure}

Fig.~\ref{fig:all_problems_impact_score_progress} plots the best (blue, dashed) and mean (orange, dash-dotted) inequality impact scores observed over 20 generations of \textsc{SAGE} on the TSP, MCND, CWLP, JSSP, IMO6, PDPTW, and SHUC benchmarks. Shaded bands indicate one standard deviation.

\noindent Generation 0 marks the neutral score of 10 for the baseline without an added family, and generation 1 is the initial population produced by the initializer agent. The best score rises from the initial population to the final generation on every benchmark. It rises from 9.4 to 13.7 on TSP, 10.3 to 12.0 on MCND, 15.4 to 19.3 on JSSP, 20.7 to 26.4 on CWLP, 12.4 to 17.4 on IMO6, 10.2 to 14.7 on PDPTW, and 19.6 to 23.7 on SHUC.

\subsection{Token Matched Initializer Control}
\label{app:evo_control}

We compare five generations of \textsc{SAGE} with iterative initialization and feedback on JSSP using DeepSeek-R1. We follow the setup of Section~\ref{sec:experimental_setup}, using 24 threads for development scoring and \texttt{Seed}=42 for test solves. Both runs start from the same initial population, with a best score of 17.60, and use the same frozen development instances. The control repeatedly calls the standard initializer with the baseline model code and feedback on recent ideas, scores, and rejections. It receives no code from earlier candidates and uses no parent selection, crossover, mutation, or elitism. The common retry procedure can reuse a candidate's own code within its round.

\noindent After shared initialization, \textsc{SAGE} uses 761,451 provider reported tokens and the control uses 765,308, or 0.51\% more. Each run selects its best family within the \textsc{SAGE} token budget, including the shared initial population. The control accepts 15 new families, of which 14 finish within this budget. Including its final candidate leaves the selected family unchanged. Table~\ref{tab:evo_control_dev} in Section~\ref{sec:evo_diagnostics} reports the development comparison.

\noindent After shared initialization, \textsc{SAGE} completes 35 impact scorer calls totaling 2328.54 recorded solver seconds. The control completes 15 impact scorer calls totaling 1150.56 solver seconds, with 14 calls and 1053.22 seconds within the target token budget.

\noindent The best control family is the machine critical path bound from the shared initialization, described in Appendix~\ref{app:jssp}. The best \textsc{SAGE} family, produced by Exploratory Mutation, sorts the operations of each machine by nonincreasing tail. It bounds the makespan by the largest prefix value of the minimum head plus the prefix processing time plus the minimum tail. The full prefix recovers the control bound. We compare these bounds on the JSSP test set and check both families against the stored proven optimal solutions using the protocol of Section~\ref{sec:deployment_time_sanity}. The evolved bound is strictly larger on 38 instances and equal on 42, and both families preserve all 58 stored solutions.

\noindent Table~\ref{tab:evo_control} evaluates the two families selected in Table~\ref{tab:evo_control_dev} against the same baseline trace for each of the JSSP test instances. At checkpoints without an incumbent, we assign a gap of 1.

\noindent Paired differences give \textsc{SAGE} minus control improvements. Positive differences favor \textsc{SAGE}. Pointwise 95\% bootstrap intervals use 10,000 resamples of the instance pairs at each checkpoint. Wins, losses, and ties use strict comparisons of the recorded gaps.

\begin{table}[H]
  \centering
  \small
  \caption{JSSP deployment results for the families selected in Table~\ref{tab:evo_control_dev}. $\bar{\Delta}_g$ is reported as mean with standard deviation (\%). Difference and CI give the mean paired difference and its pointwise confidence interval in percentage points. W/L/T gives win, loss, and tie counts for \textsc{SAGE} against the initializer.}
  \label{tab:evo_control}
  \setlength{\tabcolsep}{3pt}
  \begin{tabular}{rrrrcr}
    \toprule
    $t$ (s) & \textsc{SAGE} & Initializer & Difference & 95\% CI & W/L/T \\
    \midrule
    5 & \meanstd{33.8}{22.6} & \meanstd{17.9}{28.1} & 15.9 & $[11.7,20.1]$ & 44 / 9 / 27 \\
    10 & \meanstd{34.7}{23.1} & \meanstd{25.1}{27.8} & 9.6 & $[6.0,13.2]$ & 41 / 11 / 28 \\
    50 & \meanstd{26.8}{26.4} & \meanstd{22.8}{31.2} & 4.0 & $[0.9,7.1]$ & 35 / 15 / 30 \\
    150 & \meanstd{37.2}{28.3} & \meanstd{34.9}{33.4} & 2.3 & $[-0.7,5.3]$ & 34 / 23 / 23 \\
    300 & \meanstd{28.4}{27.1} & \meanstd{23.5}{34.5} & 4.9 & $[1.1,8.7]$ & 38 / 17 / 25 \\
    1000 & \meanstd{38.6}{26.8} & \meanstd{34.5}{35.2} & 4.1 & $[-0.4,8.6]$ & 35 / 22 / 23 \\
    2000 & \meanstd{40.5}{25.9} & \meanstd{38.2}{37.1} & 2.3 & $[-2.8,7.4]$ & 33 / 24 / 23 \\
    \bottomrule
  \end{tabular}

\end{table}

\noindent Table~\ref{tab:evo_control} reports higher mean relative gap improvement for \textsc{SAGE} at all seven checkpoints, with paired differences from 2.3 to 15.9 percentage points. The pointwise 95\% confidence intervals are above zero at 5, 10, 50, and 300\,s and include zero at 150, 1000, and 2000\,s.

\section{Supplementary Mathematical Preliminaries}
\label{app:preliminaries}

\subsection{Geometric intuition for valid inequalities}
\label{app:geometric_intuition}
\begin{figure}[htbp]
    \centering
    \includegraphics[width=0.5\textwidth]{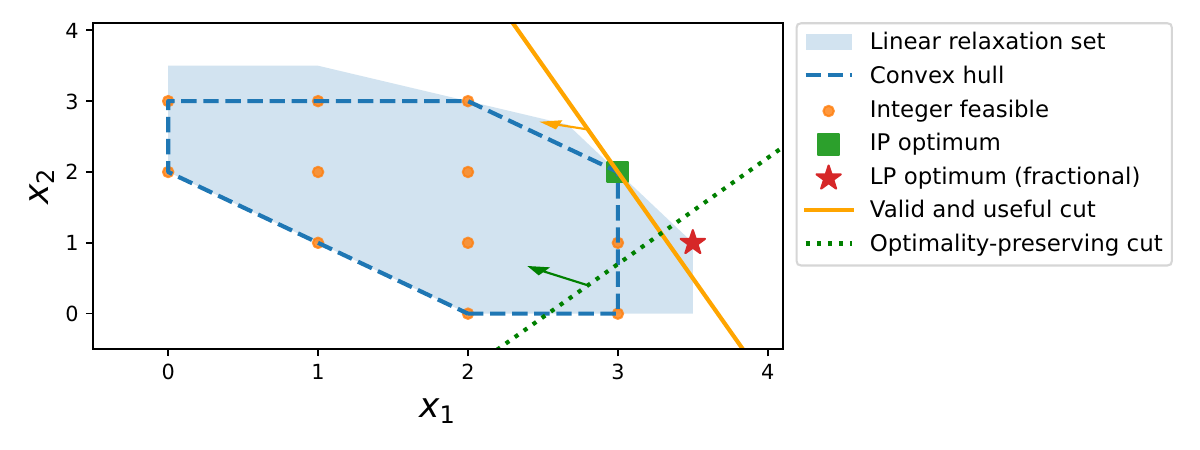}
    \caption{Relationship between the integer feasible set \(S\), its LP relaxation \(P\), the set \(\operatorname{conv}(S)\), a valid inequality that removes the fractional LP optimum (useful), and an optimality-preserving inequality that may remove some feasible points while keeping the optimal value.}
    \Description{A geometric sketch comparing an integer feasible set, its LP relaxation, the convex hull, and valid and optimality-preserving inequalities.}
    \label{fig:ip-lp-convex-cut}
\end{figure}

Let \(\operatorname{conv}(S)\) denote the convex hull of~\(S\), i.e., the smallest convex set that contains~\(S\). Valid inequalities tighten \(P\) toward \(\operatorname{conv}(S)\), narrowing the gap between the LP optimum and the true MILP optimum. Adding inequalities that are valid for \(\operatorname{conv}(S)\) can separate some fractional extreme points of~\(P\) and can reduce the integrality gap. Optimality-preserving inequalities need not be valid for \(\operatorname{conv}(S)\) but can still speed up branch-and-bound by pruning provably suboptimal regions. Fig.~\ref{fig:ip-lp-convex-cut} provides a schematic view of the integer feasible set \(S\), its LP relaxation \(P\), and the convex hull \(\operatorname{conv}(S)\). It also contrasts (i) valid inequalities that remove fractional LP optima while preserving all points in \(\operatorname{conv}(S)\), and (ii) optimality-preserving inequalities that may remove some feasible points but keep the optimal objective value unchanged. \textsc{SAGE} instead generates acceleration inequality families that are empirically checked on an admissibility set (Section~\ref{sec:solution_method}).

\subsection{Polyhedra and Facets}
\label{app:polyhedra_facets}
A polyhedron is any set $Q$ in \(\mathbb{R}^n\) that can be described by finitely many linear inequalities. The constraints $Ax \le b$ are the linear inequalities that define $Q$.
\[
Q \;=\; \{\,x \in \mathbb{R}^{n} : Ax  \le b\}.
\]
Polyhedra are convex, and if \(Q\) is bounded, it is called a polytope.

A face of a polyhedron $Q$ is a set of the form $F:=Q \cap \{x \in \mathbb{R}^{n} : cx=\beta\}$ where $cx \le \beta$ is a valid inequality for $Q$. We say that the valid inequality $cx \le \beta$ defines the face $F$ of the polyhedron $Q$. The equation $cx=\beta$ defines a supporting hyperplane that intersects $Q$. A facet is a face of maximum dimension (i.e., one dimension less than that of the polyhedron).

For example, consider a cube as a 3D polytope in $\mathbb{R}^{3}$ and a plane as the supporting hyperplane. Intersecting a cube with a supporting hyperplane may result in a 0D corner point, a 1D straight line, or a 2D plane. The corner point, the line (edge of a cube), and the 2D plane are all faces of the 3D cube. However, only the 2D plane is a facet of the 3D cube.

\subsection{Facet-Defining Valid Inequalities}
\label{app:facet_defining}
In integer programming, valid inequalities that are facet-defining for $\operatorname{conv}(S)$ are especially important because they are necessary and sufficient for characterizing $\operatorname{conv}(S)$. Once $\operatorname{conv}(S)$ is characterized, solving the MILP over $S$ reduces to the linear program of maximizing $c^\top x + h^\top y$ over $\operatorname{conv}(S)$, which is polynomially solvable.

\section{Additional Experiments}
\label{app:additional_experiments}

\subsection{Aggregate gap improvement curves}
\label{app:gap_improvement_curves}

\begin{figure}[htbp]
  \centering
  \includegraphics[width=0.9\textwidth]{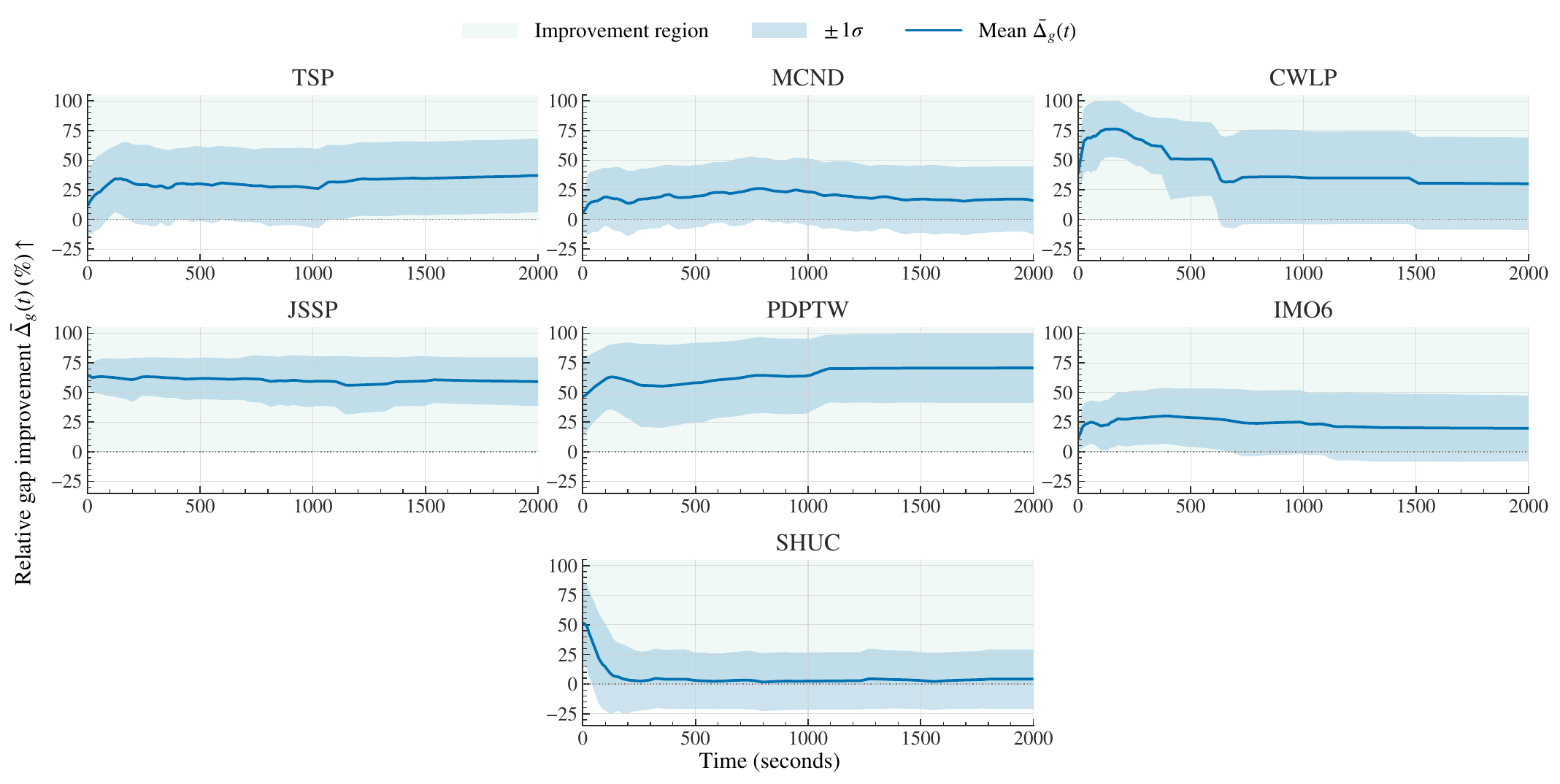}
			  \caption{Change over time (up to 2000\,s) of the mean stabilized relative gap improvement \(\bar{\Delta}_g(t)\) after adding the selected \textsc{SAGE} acceleration inequality family to the baseline MILP for TSP, MCND, CWLP, JSSP, PDPTW, IMO6, and SHUC.  The solid line is the mean across test instances at each wall-clock time and the shaded band indicates one standard deviation (\(\sigma\)).  Positive values mean the augmented model has a smaller optimality gap at that wall-clock time.}
  \Description{Line plots of mean stabilized relative gap improvement over time with shaded standard deviation bands.}
  \label{fig:gap_rel_panel}
\end{figure}

Fig.~\ref{fig:gap_rel_panel} reports the mean stabilized relative gap improvement \(\bar{\Delta}_g(t)\) as a function of time after augmenting the baseline MILP with the selected \textsc{SAGE} acceleration inequality family, averaged across held-out test instances (TSP, MCND, CWLP, JSSP, PDPTW, IMO6, and SHUC). This plot complements the checkpoint summaries in Table~\ref{tab:post_ga_results} by showing the full 2000\,s profile rather than only selected checkpoints.

\subsection{Held-Out LP Separation Rates}
\label{app:heldout_lp_separation}

As a robustness and admissibility check, we also test whether the selected \textsc{SAGE} acceleration inequality family separates the baseline LP optimum on held-out test instances. This repeats the LP-separation check from Section~\ref{sec:solution_method}, but uses \(D_t\) instead of the small admissibility set \(D_{adm}\). For each test instance, we fix the original decision variables to the baseline LP optimum, append the selected acceleration inequality family, and solve only for any auxiliary variables introduced by the family. If this reduced model is infeasible, the family separates that LP optimum. The resulting rate measures one baseline LP optimum per test instance.

\begin{table}[htbp]
  \centering
  \small
  \caption{Held-out test-set LP separation rate of the selected \textsc{SAGE} acceleration inequality family.}
  \label{tab:heldout_lp_separation}
  \begin{tabular}{lcc}
    \toprule
    Problem & Separated / Total & Rate (\%) \\
    \midrule
    TSP & 54/54 & 100.0 \\
    MCND & 72/81 & 88.9 \\
    CWLP & 36/36 & 100.0 \\
    JSSP & 80/80 & 100.0 \\
    PDPTW & 165/165 & 100.0 \\
    IMO6 & 34/34 & 100.0 \\
    SHUC & 48/56 & 85.7 \\
    \bottomrule
  \end{tabular}
\end{table}
\noindent Table~\ref{tab:heldout_lp_separation} shows that the selected acceleration inequality family separates the LP optimum on all TSP, CWLP, JSSP, PDPTW, and IMO6 test instances. For MCND and SHUC, the measured separation rates are 88.9\% and 85.7\%, respectively.

\subsection{Discovery Cost and Deployment Payback}
\label{app:payback}

This experiment measures the cost of one \textsc{SAGE} search, deployment setup, and payback on 450 reserved test instances from TSP, MCND, CWLP, JSSP, PDPTW, and IMO6. We charge the full search budget from Section~\ref{sec:experimental_setup} as the discovery cost. One Gemini 3 Pro search costs about USD 30 in API calls. Appendix~\ref{app:llm_params} reports API settings and total costs. In the JSSP control run of Appendix~\ref{app:evo_control}, LLM replies account for 83.3\% of the elapsed wall clock time. Deployment setup is the added time for model construction, family instantiation, and loading the persistent solver model. Payback uses target $g\le10^{-3}$ and caps solver time to target at $T_{\max}$. The Baseline and \textsc{SAGE} columns report the mean capped solver times. The Solves column divides the search budget in seconds by the unrounded difference between these means. The last column adds setup time to the \textsc{SAGE} time before computing payback.

\begin{table}[H]
  \centering
  \small
  \caption{Discovery payback at $g\le10^{-3}$ on reserved test instances. Setup is the mean added deployment setup time. Solves is the number of deployments whose solver time savings recover the search budget. The last column charges the added setup time to \textsc{SAGE} as well.}
  \label{tab:payback}
  \setlength{\tabcolsep}{4pt}
  \begin{tabular}{@{}lrrrrr@{}}
    \toprule
    Problem & Setup (s) & Baseline (s) & SAGE (s) & Solves & Solves with setup \\
    \midrule
    TSP   & 0.94  & 1063.8 & 760.3  & 238  & 238 \\
    MCND  & 0.006 & 640.7  & 627.1  & 5276 & 5278 \\
    CWLP  & 0.97  & 220.5  & 11.4   & 345  & 347 \\
    JSSP  & 0.46  & 1886.5 & 1846.1 & 1782 & 1803 \\
    PDPTW & 0.08  & 2000.0 & 1962.7 & 1933 & 1935 \\
    IMO6  & 10.55 & 1594.5 & 1555.1 & 1828 & 2496 \\
    \bottomrule
  \end{tabular}
\end{table}
\noindent The mean added setup time ranges from 0.006 seconds on MCND to 10.55 seconds on IMO6. The estimated payback in solver time ranges from 238 solves on TSP to 5276 solves on MCND. Charging the setup time as well leaves five benchmarks within 21 solves of these values and raises IMO6 from 1828 to 2496 solves.

\subsection{Sensitivity to the Continuous Solution Method}
\label{app:method_sensitivity}

This experiment fixes the selected TSP family and compares the automatic method with the deterministic concurrent method on TSP test instances. We use a 300 second limit.

\begin{table}[H]
  \centering
  \small
  \caption{TSP sensitivity to the Gurobi continuous solution method.}
  \label{tab:method_sensitivity}
  \begin{tabular}{cc}
    \toprule
    Method & Mean stabilized relative gap improvement (\%) \\
    \midrule
    Automatic ($-1$) & $34.3\pm29.3$ \\
    Deterministic concurrent (4) & $27.6\pm33.1$ \\
    \bottomrule
  \end{tabular}
\end{table}
\noindent Both methods give positive mean improvement. The mean is $34.3\%$ with the automatic method and $27.6\%$ with the deterministic concurrent method.

\subsection{Anonymized Formulation Control}
\label{app:rntd_control}

This experiment tests SAGE on an unpublished problem and formulation for which no inequality family had been published. Appendix~\ref{app:rntd} gives the RNTD description, formulation, selected family, generator, and instance details. We compare the baseline formulation with the same formulation augmented by the family selected on the development set.

For this control, the baseline model input in Appendix~\ref{app:agents_prompts} was the Pyomo implementation of Eq.~\eqref{eq:rntd_milp}. Its identifiers and comments referred only to nodes, links, commodities, capacity, degree, hops, and a reference vector. The prompt did not contain the RNTD name or an application description.

\begin{table}[H]
  \centering
  \small
  \caption{RNTD test mean stabilized relative gap improvement $\bar{\Delta}_g\ (\%)$. Mean $\pm\sigma$ is reported at each checkpoint.}
  \label{tab:rntd_gap}
  \setlength{\tabcolsep}{5pt}
  \begin{tabular}{lrrrrrrrr}
    \toprule
    Problem & 5 s & 10 s & 50 s & 150 s & 300 s & 1000 s & 2000 s & Average \\
    \midrule
    RNTD & $18\pm35$ & $-50\pm197$ & $24\pm17$ & $24\pm19$ & $25\pm20$ & $28\pm20$ & $23\pm19$ & 13 \\
    \bottomrule
  \end{tabular}
\end{table}
\noindent The mean improvement is positive at six of seven checkpoints and is $23\pm19\%$ at 2000 seconds. At 10 seconds, the mean is $-50\%$, while the augmented model has a smaller gap on 36 test instances.

\begin{table}[H]
  \centering
  \small
  \caption{RNTD time to target results. B/A counts the baseline and augmented runs that reach the target.}
  \label{tab:rntd_time}
  \begin{tabular}{ccc}
    \toprule
    Target gap & B/A & SGM speedup \\
    \midrule
    $10^{-5}$ & 9/12  & $4.2\times$ \\
    $10^{-4}$ & 9/12  & $4.2\times$ \\
    $10^{-3}$ & 9/13  & $4.2\times$ \\
    $10^{-2}$ & 44/44 & $2.7\times$ \\
    $10^{-1}$ & 44/44 & $1.0\times$ \\
    Average       & 23.0/25.0 & $3.3\times$ \\
    \bottomrule
  \end{tabular}
\end{table}
\noindent The SGM speedup exceeds one at four of five target gaps and averages $3.3\times$.

\subsection{Model Size and LP Relaxation}
\label{app:model_size_lp}
We count the baseline model and the increments from the complete selected family on every test instance before solver presolve. For each instance with a proven integer objective, we also compute
\[
g_{LP}=\frac{|z_{IP}-z_{LP}|}{\max\{|z_{IP}|,10^{-8}\}}.
\]
Gap closure is computed per instance as $(z_{LP}^{aug}-z_{LP}^{base})/(z_{IP}-z_{LP}^{base})$. Table~\ref{tab:selected_family_reporting} reports means across each test set. Ranges follow the model size means in brackets. The LP objective improvement count uses every test instance.

\begin{table}[H]
\centering
\scriptsize
\caption{Model size and LP relaxation results for the baseline and the model augmented by the selected \textsc{SAGE} family. Auxiliary variables and added rows are increments from the selected family.}
\label{tab:selected_family_reporting}
\textit{(a) Model size.}\\[2pt]
\resizebox{\textwidth}{!}{%
\begin{tabular}{@{}lrrrr@{}}
\toprule
    Benchmark & Baseline variables & Auxiliary variables & Baseline rows & Added rows \\
\midrule
    TSP & 15,821.6 [196, 52,441] & 0 & 15,931.0 [211, 52,671] & 31,319.2 [351, 104,196] \\
    MCND & 24,991.9 [4,920, 63,315] & 0 & 2,485.2 [920, 4,315] & 38.2 [36, 40] \\
    CWLP & 34,398.0 [816, 100,100] & 0 & 420.2 [66, 1,100] & 34,343.3 [793, 100,009] \\
    JSSP & 21,382.2 [1,801, 101,001] & 1,431.2 [450, 4,000] & 42,046.9 [3,375, 200,000] & 1,613.1 [246, 6,081] \\
    PDPTW & 11,090 & 0 & 4,565 & 4,161 \\
    IMO6 & 20,759.5 [63, 73,224] & 0 & 7,524.5 [39, 26,010] & 19,368.0 [41, 69,407] \\
    SHUC & 276,709.7 [38,928, 426,240] & 61,105.4 [6,935, 92,910] & 369,301.0 [41,998, 561,468] & 336,793.7 [38,927, 511,562] \\
\bottomrule
\end{tabular}}

\vspace{5pt}
\textit{(b) LP relaxation.}\\[2pt]
\begin{tabular}{@{}lrrrrr@{}}
\toprule
    Benchmark & Proven $N$ & Baseline LP gap & Augmented LP gap & Gap closed & LP objective improved \\
\midrule
    TSP & 54 & 21.90\% & 8.96\% & 62.51\% & 53/54 \\
    MCND & 77 & 25.45\% & 24.62\% & 2.81\% & 72/81 \\
    CWLP & 36 & 24.93\% & 0.38\% & 98.59\% & 36/36 \\
    JSSP & 58 & 48.34\% & 2.94\% & 88.85\% & 80/80 \\
    PDPTW & 0 & n/a & n/a & n/a & 165/165 \\
    IMO6 & 9 & 90.01\% & 26.23\% & 70.88\% & 34/34 \\
    SHUC & 53 & 0.57\% & 0.43\% & 26.44\% & 48/56 \\
\bottomrule
\end{tabular}
\end{table}

\noindent Table~\ref{tab:selected_family_reporting} shows that only JSSP and SHUC introduce auxiliary variables. Among benchmarks with proven objectives, the mean per-instance gap closure is smallest on MCND (2.81\%, mean LP gap 25.45\% to 24.62\%) and largest on CWLP (98.59\%, mean LP gap 24.93\% to 0.38\%), while PDPTW has no proven objective and its mean LP objective increases from 22,013.02 to 504,451.72 on its test instances. On TSP, the family separates the baseline LP optimum on all test instances (Table~\ref{tab:heldout_lp_separation}) but leaves the LP objective unchanged on instance ts225, so the LP objective improves on 53 instances.

\subsection{Static Literature Augmentation Baselines}
\label{app:literature_cuts}

This experiment compares the selected static \textsc{SAGE} acceleration inequality families with a static set of valid inequalities from the literature. We keep the same base formulations, test sets, solver budget, and time-to-target evaluation protocol used in Section~\ref{sec:time_to_target}. The only change is that the \textsc{SAGE} acceleration inequality families are replaced by the corresponding published valid inequalities. Since several of these inequalities are indexed by node subsets, this is a static model augmentation comparison rather than a dynamic separation implementation.

Let \(\mathcal{G}=\{10^{-5},10^{-4},10^{-3},10^{-2},10^{-1}\}\). For an instantiated static literature valid inequality set \(L\), Table~\ref{tab:literature_cut_baselines} reports
\[
\bar{s}_{L}
=\frac{1}{|\mathcal{G}|}\sum_{g\in\mathcal{G}}
\frac{\mathrm{SGM}(t_{\mathrm{ref}}(g))}{\mathrm{SGM}(t_{L}(g))},
\]
using the same shifted geometric mean convention as Table~\ref{tab:post_ga_time}. Values larger than one mean that the model augmented with static literature valid inequalities reaches the target gaps faster than the baseline MILP on average.

\begin{table}[htbp]
  \centering
  \small
  \caption{Mean SGM speedup for static literature valid inequalities and \textsc{SAGE} acceleration inequality families. Each value is averaged over the same five target gaps used in Table~\ref{tab:post_ga_time}.}
  \label{tab:literature_cut_baselines}
  \setlength{\tabcolsep}{6pt}
  \begin{tabular}{@{}lcc@{}}
    \toprule
    Problem & Static literature valid inequalities & \textsc{SAGE} families \\
    \midrule
    CWLP & \(3.01\times\) & \(4.02\times\) \\
    JSSP & \(1.91\times\) & \(3.33\times\) \\
    MCND & \(1.01\times\) & \(1.01\times\) \\
    TSP & \(2.57\times\) & \(2.87\times\) \\
    \bottomrule
  \end{tabular}
\end{table}
\noindent Table~\ref{tab:literature_cut_baselines} shows that the \textsc{SAGE} acceleration inequality families yield a larger mean speedup than the instantiated static literature valid inequalities on CWLP, JSSP, and TSP, and are essentially tied on MCND. The average over these four rows is \(2.81\times\) for \textsc{SAGE} and \(2.13\times\) for the static literature valid inequalities. The largest absolute gap is on JSSP (\(3.33\times\) vs.\ \(1.91\times\), a \(1.42\times\) margin), followed by CWLP (\(4.02\times\) vs.\ \(3.01\times\), a \(1.01\times\) margin). TSP is close (\(2.87\times\) vs.\ \(2.57\times\), a \(0.30\times\) margin), and MCND is tied (\(1.01\times\) vs.\ \(1.01\times\)).

\paragraph{CWLP inequalities.}
For each warehouse \(j\), the capacity row in Eq.~\eqref{eq:cwlp_milp} induces a 0--1 knapsack structure over customer-assignment variables \(x_{ij}\). Following the 0--1 knapsack inequality in Balas~\cite{balas1975facets} and the facility-location inequalities in Aardal et al.~\cite{aardal1995capacitated}, any set \(C\subseteq I\) with \(\sum_{i\in C}d_i>u_j\) gives
{\scriptsize
\begin{equation}
\left\{
\begin{aligned}
\sum_{i\in C} x_{ij}
&\le (|C|-1)y_j
&&\forall j\in J,\; C\subseteq I:\sum_{i\in C}d_i>u_j,\\[2pt]
\sum_{i\in I:\; d_i>u_j/(q+1)}x_{ij}
&\le q\,y_j
&&\forall j\in J,\; q=1,\dots,|I|-1 .
\end{aligned}
\right.
\tag{LIT-CWLP}\label{eq:lit_cwlp_cover}
\end{equation}
}
The second row is the threshold form of the same inequality: if each selected customer has demand greater than \(u_j/(q+1)\), then no feasible integer solution can assign \(q+1\) such customers to warehouse \(j\). This is the same local capacity logic present in the CWLP \textsc{SAGE} acceleration inequality family, but the \textsc{SAGE} version also includes global demand-threshold slot bounds.

\paragraph{JSSP inequalities.}
Using the JSSP notation from Appendix~\ref{app:jssp}, write an operation as \(a=(j,k)\), its start time as \(S_a:=S_{j,k}\), and its processing time as \(p_a:=p_{j,k}\). For \(Q\subseteq O_m\), define
\[
P_Q:=\sum_{a\in Q}p_a,\qquad
H_Q:=\min_{a=(j,k)\in Q}\sum_{t<k}p_{j,t},\qquad
T_Q:=\min_{a=(j,k)\in Q}\sum_{t>k}p_{j,t},
\]
and \(\Phi_Q:=\sum_{\{a,b\}\subseteq Q}p_a p_b\). Applegate and Cook~\cite{applegate1991computational} give the two inequalities
{\scriptsize
\begin{equation}
\left\{
\begin{aligned}
\sum_{a\in Q}p_a S_a
&\ge H_Q P_Q + \Phi_Q
&&\forall m,\; Q\subseteq O_m,\\[2pt]
\sum_{a\in Q}p_a(C_{\max}-S_a)
&\ge T_Q P_Q + \sum_{a\in Q}p_a^2 + \Phi_Q
&&\forall m,\; Q\subseteq O_m .
\end{aligned}
\right.
\tag{LIT-JSSP}\label{eq:lit_jssp_basic}
\end{equation}
}
These inequalities use job heads and tails to bound how early a set of operations on one machine can start and how late they can finish. In our experiments they yield a meaningful mean speedup (Table~\ref{tab:literature_cut_baselines}), while the \textsc{SAGE} acceleration inequality family in Appendix~\ref{app:jssp} acts directly as a makespan lower-bound family and produces a larger speedup on the same benchmark.

\paragraph{MCND inequalities.}
For a node subset \(S\subseteq N\), let \(\delta^{+}(S)\), \(K(S)\), \(B\subseteq K(S)\), and \(D_B\) be defined as in Appendix~\ref{app:mcnd}. Chouman et al.~\cite{chouman2017commodity} define \(b^B_{ij}:=\min\{u_{ij},D_B\}\) for this node subset relaxation and include the following two inequalities:
{\scriptsize
\begin{equation}
\left\{
\begin{aligned}
\sum_{(i,j)\in\delta^+(S)}\min\{u_{ij},D_B\}\,y_{ij}
&\ge D_B
&&\forall S\subseteq N,\; B\subseteq K(S),\\[2pt]
\sum_{(i,j)\in\delta^+(S)} y_{ij}
&\ge q_{S,B}
&&\forall S\subseteq N,\; B\subseteq K(S),
\end{aligned}
\right.
\tag{LIT-MCND}\label{eq:lit_mcnd_cutset}
\end{equation}
}
where \(q_{S,B}\) is the smallest number of largest-capacity arcs in \(\delta^+(S)\) whose total capacity can cover \(D_B\). The selected \textsc{SAGE} MCND rows in Eq.~\eqref{eq:evocut_mcnd} use the same algebraic form, but they are added only for the generated pairs used by the implementation.

\paragraph{TSP inequalities.}
The compact TSP model in Eq.~\eqref{eq:tsp_mtz_compact} uses the Miller--Tucker--Zemlin ordering variables~\cite{miller1960integer}. Desrochers and Laporte~\cite{desrochers1991improvements} give the following lifted ordering rows in the notation of Appendix~\ref{app:tsp}:
{\scriptsize
\begin{equation}
\left\{
\begin{aligned}
u_i-u_j+(n-1)x_{ij}+(n-3)x_{ji}
&\le n-2
&&\forall i,j\in V\setminus\{1\},\; i\neq j,\\[2pt]
u_i
&\ge 3-x_{1i}+(n-3)x_{i1}
&&\forall i\in V\setminus\{1\},\\[2pt]
u_i
&\le (n-1)+x_{i1}-(n-3)x_{1i}
&&\forall i\in V\setminus\{1\}.
\end{aligned}
\right.
\tag{LIT-TSP}\label{eq:lit_tsp_mtz}
\end{equation}
}
The TSP \textsc{SAGE} acceleration inequality family in Eq.~\eqref{eq:evocut_ec} contains these rows and augments them with explicit two-cycle and depot-anchored short-cycle eliminators.

\subsection{Solver Default Parameters}
\label{app:solver_default_parameters}
Table~\ref{tab:solver_default_params} lists important solver defaults that were not set by our experiments but can affect MILP search. We omit \texttt{MIPGap}, \texttt{Seed}, \texttt{Method}, \texttt{WorkLimit}, \texttt{TimeLimit}, and the stated Gurobi thread setting because these are already given in Section~\ref{sec:experimental_setup}. The listed values are documented defaults, not tuned settings~\citep{gurobi2026parameters,scip2026parameters,highs2026options}.

\begin{table}[htbp]
  \centering
  \scriptsize
  \caption{Important default solver parameters not otherwise listed in the experimental setup.}
  \label{tab:solver_default_params}
  \setlength{\tabcolsep}{4pt}
  \renewcommand{\arraystretch}{1.12}
  \begin{tabular}{@{}>{\raggedright\arraybackslash}p{0.10\textwidth}
                  >{\raggedright\arraybackslash}p{0.28\textwidth}
                  >{\raggedright\arraybackslash}p{0.12\textwidth}
                  >{\raggedright\arraybackslash}p{0.41\textwidth}@{}}
    \toprule
    Solver & Parameter & Default & Role \\
    \midrule
    Gurobi & \nolinkurl{MIPGapAbs} & \texttt{1e-10} & Absolute optimality gap tolerance. \\
    Gurobi & \nolinkurl{Presolve} & \texttt{-1} & Automatic presolve control. \\
    Gurobi & \nolinkurl{Cuts} & \texttt{-1} & Automatic global cut generation control. \\
    Gurobi & \nolinkurl{Heuristics} & \texttt{0.05} & Target fraction of runtime for primal heuristics. \\
    Gurobi & \nolinkurl{MIPFocus} & \texttt{0} & Balanced MIP search focus. \\
    Gurobi & \nolinkurl{Symmetry} & \texttt{-1} & Automatic symmetry detection. \\
    \midrule
    SCIP 10.0 & \nolinkurl{limits/gap} & \texttt{0} & Relative optimality gap limit. \\
    SCIP 10.0 & \nolinkurl{limits/absgap} & \texttt{0} & Absolute optimality gap limit. \\
    SCIP 10.0 & \nolinkurl{presolving/maxrounds} & \texttt{-1} & Automatic number of presolving rounds. \\
    SCIP 10.0 & \nolinkurl{separating/maxrounds} & \texttt{-1} & Automatic separation rounds after root. \\
    SCIP 10.0 & \nolinkurl{separating/maxroundsroot} & \texttt{-1} & Automatic separation rounds at the root node. \\
    SCIP 10.0 & \nolinkurl{separating/maxcuts} & \texttt{100} & Cut pool limit outside the root node. \\
    SCIP 10.0 & \nolinkurl{separating/maxcutsroot} & \texttt{2000} & Cut pool limit at the root node. \\
    SCIP 10.0 & \nolinkurl{misc/usesymmetry} & \texttt{7} & Symmetry handling mode. \\
    SCIP 10.0 & \nolinkurl{parallel/maxnthreads} & \texttt{8} & Maximum number of parallel threads. \\
    \midrule
    HiGHS & \nolinkurl{mip_rel_gap} & \texttt{1e-4} & Relative optimality gap tolerance. \\
    HiGHS & \nolinkurl{mip_abs_gap} & \texttt{1e-6} & Absolute optimality gap tolerance. \\
    HiGHS & \nolinkurl{presolve} & \texttt{choose} & Automatic presolve control. \\
    HiGHS & \nolinkurl{mip_heuristic_effort} & \texttt{0.05} & Effort level for MIP heuristics. \\
    HiGHS & \nolinkurl{mip_allow_cut_separation_at_nodes} & \texttt{true} & Enables cut separation at branch-and-bound nodes. \\
    HiGHS & \nolinkurl{mip_detect_symmetry} & \texttt{true} & Enables MIP symmetry detection. \\
    HiGHS & \nolinkurl{threads} & \texttt{0} & Automatic thread count. \\
    HiGHS & \nolinkurl{parallel} & \texttt{choose} & Automatic parallel execution mode. \\
    \bottomrule
  \end{tabular}
\end{table}

\subsection{Illustrative TSP Instance Trajectories}
\label{app:tsp_examples}
\begin{figure}[H]
  \centering
  \includegraphics[width=0.6\textwidth]{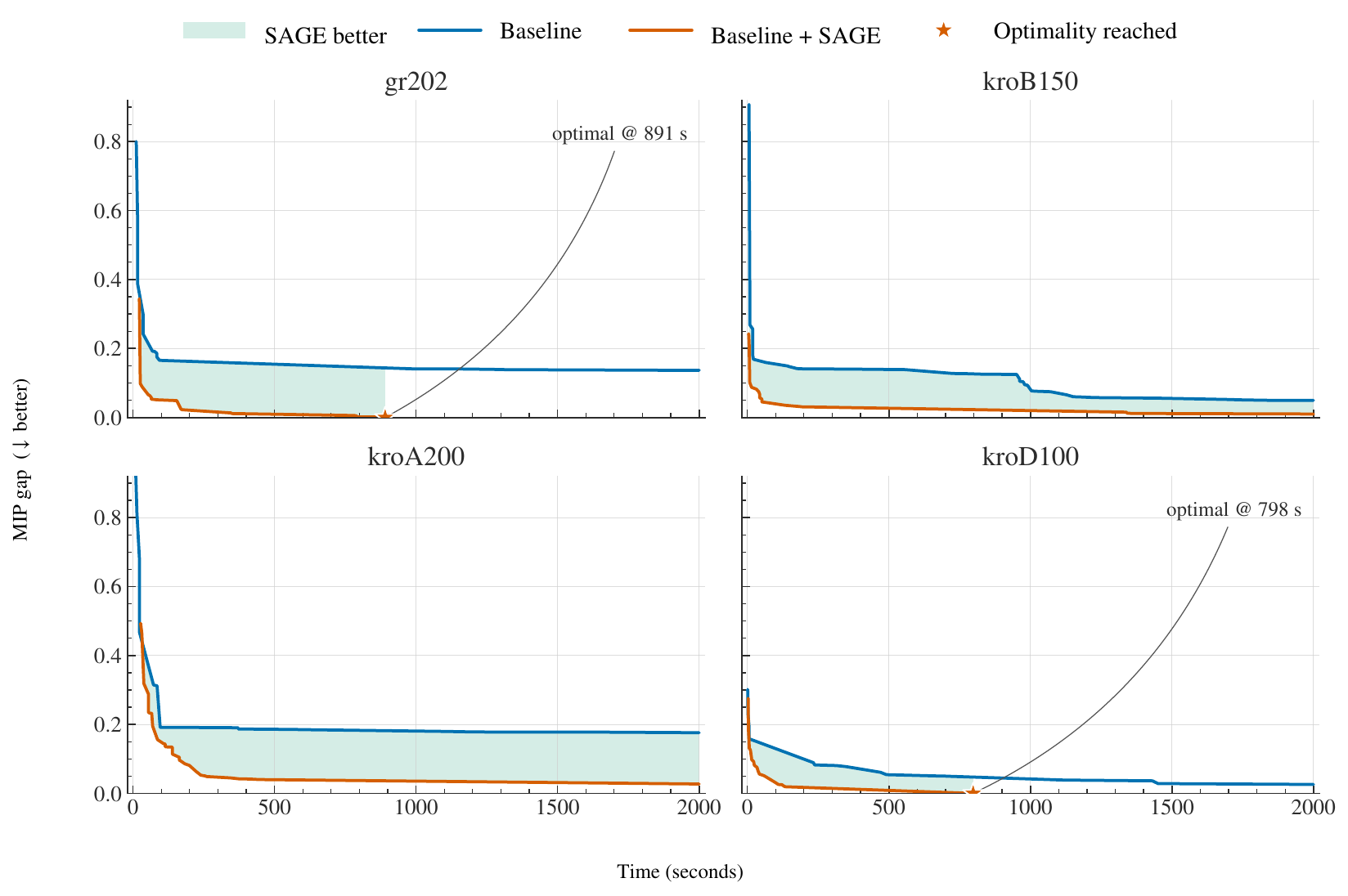}
  \caption{Gap trajectories over time for four representative TSP instances (\texttt{gr202}, \texttt{kroB150}, \texttt{kroA200}, and \texttt{kroD100}) from \cite{reinelt1991tsplib}. Orange: baseline MILP augmented with the selected \textsc{SAGE} acceleration inequality family. Blue: baseline MILP.}
  \Description{Four TSP gap trajectory plots comparing the baseline MILP with the augmented MILP over time.}
  \label{fig:tsp_examples}
\end{figure}

Fig.~\ref{fig:tsp_examples} plots the full gap traces over time for four instances from the Traveling Salesman Problem Library (\textsc{TSPLIB}). Each panel compares the baseline MILP (blue line) with the same model augmented by the selected \textsc{SAGE} acceleration inequality family found during evolution (orange line). The acceleration inequality family used in Fig.~\ref{fig:tsp_examples} for the TSP model is described in Appendix~\ref{app:tsp}.

\noindent The augmented model reaches optimality on \texttt{gr202} at 891\,s and \texttt{kroD100} at 798\,s. Its gap curve lies below the baseline curve for most of the plotted horizon in all four panels.

\subsection{Best Bound Trajectories for Representative TSP Instances}
\label{app:tsp_best_bound_exp}

\begin{figure}[H]
  \centering
  \includegraphics[width=0.6\textwidth]{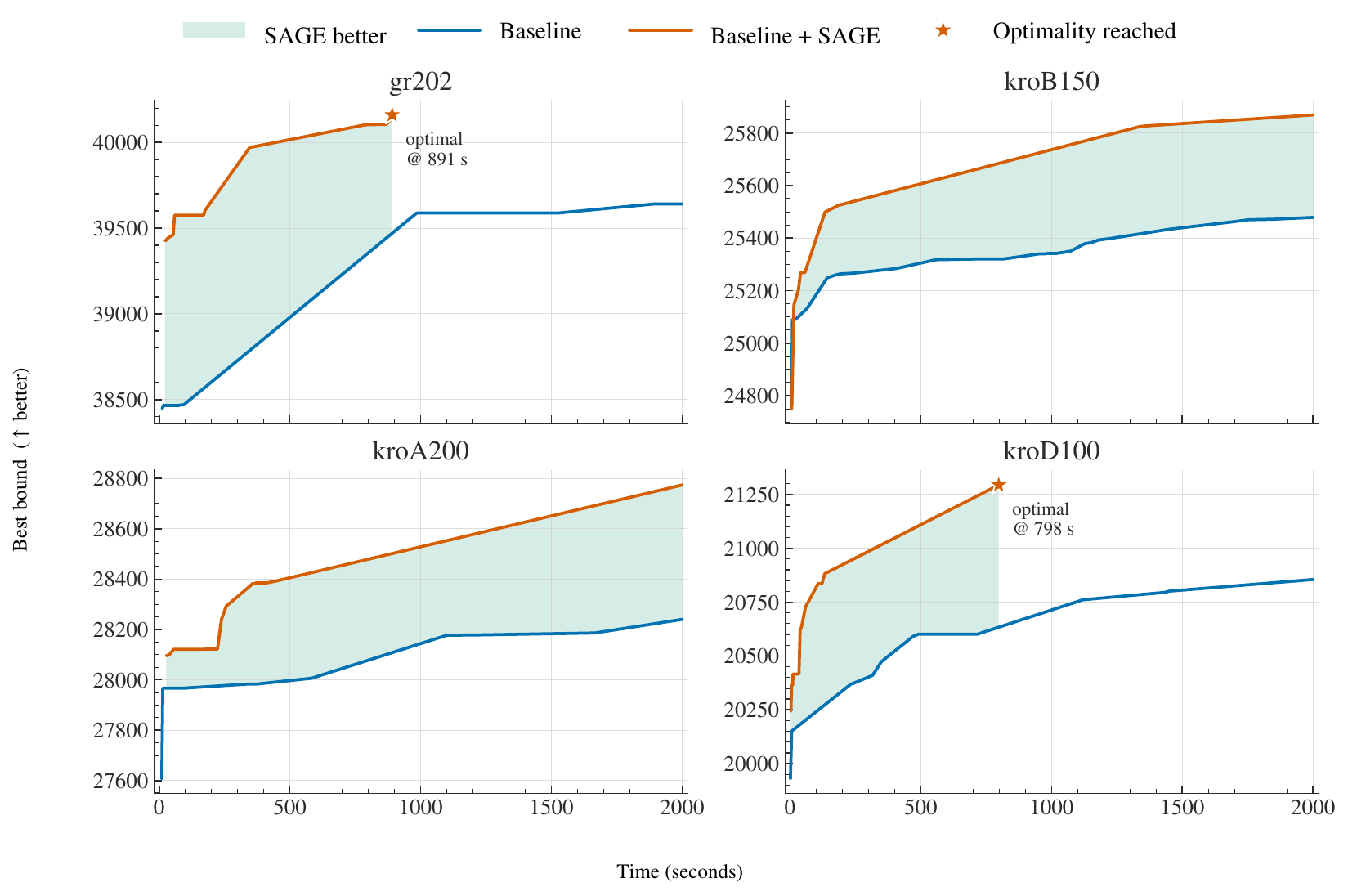}
  \caption{Evolution of the solver's best bound for four representative TSP instances. Blue lines: baseline MILP (reference). Orange lines: MILP with the selected \textsc{SAGE} acceleration inequality family.}
  \Description{Four TSP best-bound plots comparing the baseline MILP with the augmented MILP over time.}
  \label{fig:tsp_best_bound_panel}
\end{figure}

Fig.~\ref{fig:tsp_best_bound_panel} complements the gap results of Section~\ref{sec:post_ga_generalization} by plotting the solver's reported best bound over time for four representative TSP test instances drawn from \(D_{t}\). For each instance, we show two curves: the baseline MILP (reference) and the MILP augmented with the selected \textsc{SAGE} acceleration inequality family. Solver logs are sampled when the solver finds an MILP incumbent, up to the same 2000\,s wall-clock limit. Because TSP is formulated as a minimization problem, higher curves correspond to tighter lower bounds.

\subsection{Primal-Dual Integral (PDI) Trajectories for Representative TSP Instances}
\label{app:tsp_pdi_exp}

\begin{figure}[htbp]
  \centering
  \includegraphics[width=0.6\textwidth]{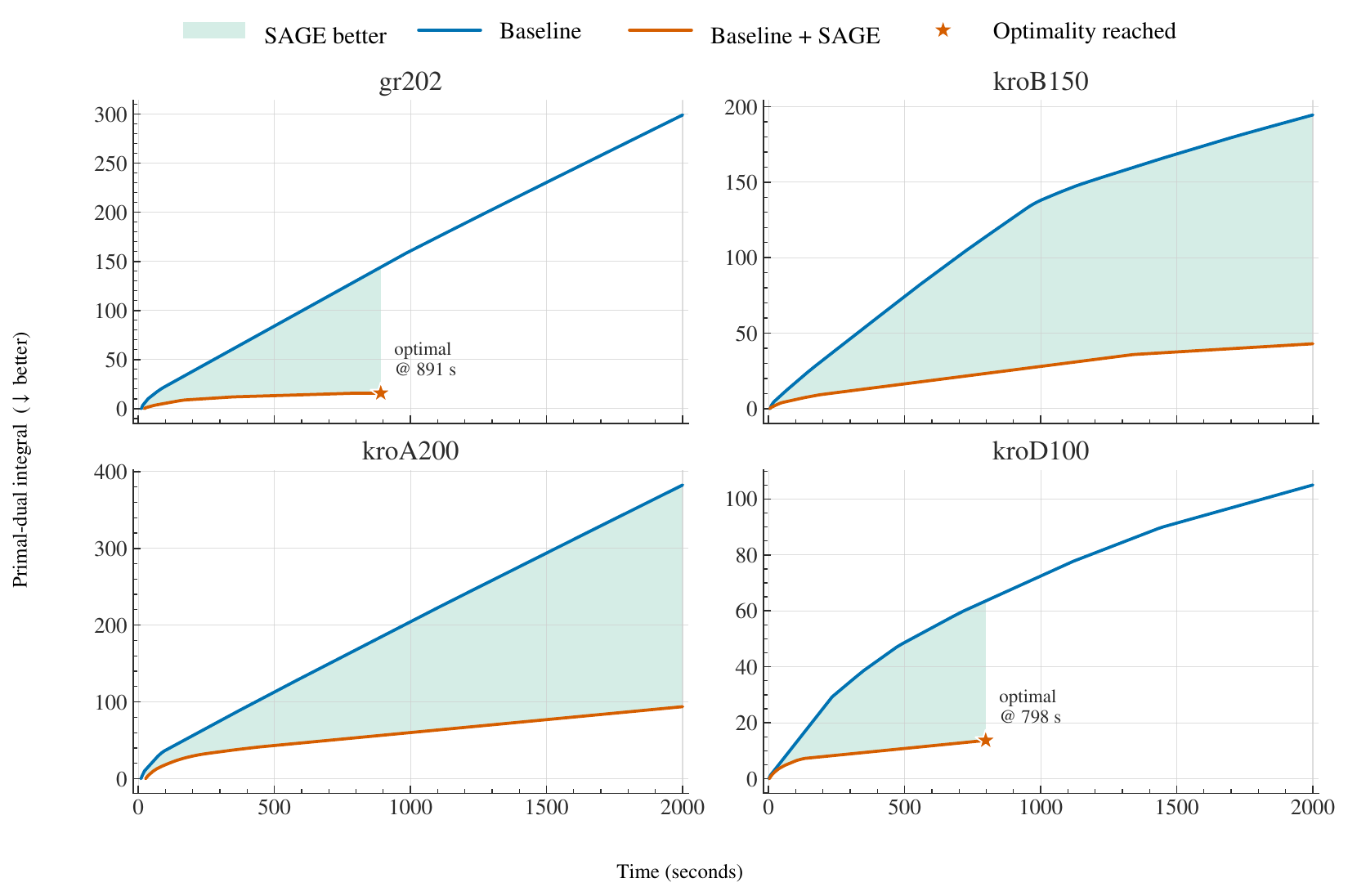}
  \caption{PDI trajectories for the same four \textsc{TSPLIB} instances used in Figs.~\ref{fig:tsp_examples}-\ref{fig:tsp_best_bound_panel}. Lower curves are better.  Orange: baseline + selected \textsc{SAGE} acceleration inequality family. Blue: baseline MILP.}
  \Description{Four TSP primal-dual integral plots comparing the baseline MILP with the augmented MILP over time.}
  \label{fig:tsp_pdi_panel}
\end{figure}

Fig.~\ref{fig:tsp_examples} reports gap trajectories, and Fig.~\ref{fig:tsp_best_bound_panel} reports solver best bounds. Fig.~\ref{fig:tsp_pdi_panel} complements both views by plotting the PDI, a cumulative metric that rewards closing the gap early and penalizes every second the gap remains open.\footnote{PDI is the time integral of the optimality gap, hence lower curves are better and an earlier flat curve indicates faster convergence.} As before, the blue curve shows the baseline MILP. The orange curve adds the selected \textsc{SAGE} acceleration inequality family. Across all four cases, the model augmented by \textsc{SAGE} reduces the total PDI by roughly \(75\!-\!95\%\), showing that acceleration inequality families can improve search progress relative to the baseline.

\subsection{Failure Analysis: MCND Cases with Slight Gap Increase}
\label{app:mcnd_gap_failure}
\begin{figure}[htbp]
  \centering
  \includegraphics[width=0.6\textwidth]{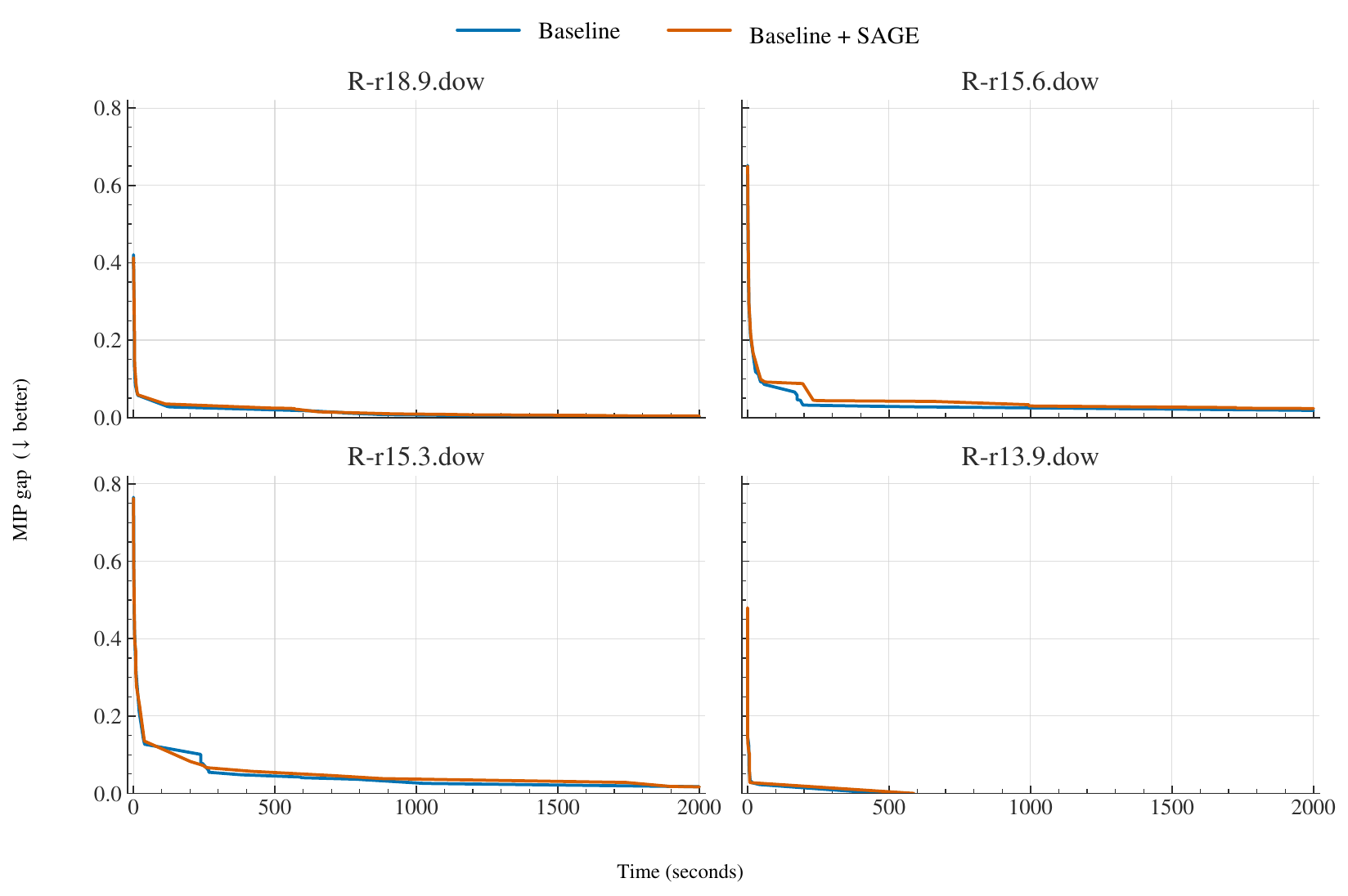}
  \caption{Gap trajectories for four MCND instances where the selected acceleration inequality family slightly hurts the gap (\texttt{R-r18.9.dow}, \texttt{R-r15.6.dow}, \texttt{R-r15.3.dow}, and \texttt{R-r13.9.dow}). Orange: baseline MILP augmented with the selected \textsc{SAGE} acceleration inequality family. Blue: baseline MILP.}
  \Description{Four MCND gap trajectory plots where the augmented MILP has slightly higher gaps than the baseline.}
  \label{fig:mcnd_gap_failure}
\end{figure}

In a small number of MCND instances, the selected acceleration inequality family does not help. Fig.~\ref{fig:mcnd_gap_failure} shows four such cases. The gap curves have similar shapes, but the model with the selected acceleration inequality family stays slightly higher for most of the horizon. This indicates that the larger model can slightly slow gap reduction on some instances.

\section{Benchmark MILP Problems and Selected {\rmfamily\mdseries\textsc{SAGE}} Families}
\label{app:benchmarks}

Table~\ref{tab:benchmark_scale} reports the scale of the seven main held-out benchmark sets. For each of these benchmarks, we show the number of test instances and the mean and standard deviation of the variable and constraint counts after model construction. The selected acceleration inequality families below show the symbolic output of \textsc{SAGE}: on mature benchmarks they sometimes recover expert strengthening structure, while on less standard models they can produce more specialized families tied to the formulation.

\begin{table}[ht]
\centering
\footnotesize
\caption{Scale of held out benchmarks before solver presolve. Variable and constraint counts are reported as mean \(\pm\) sample standard deviation for the baseline formulations and exclude the \textsc{SAGE} families.}
\label{tab:benchmark_scale}
\setlength{\tabcolsep}{6pt}
\begin{tabular}{@{}lccc@{}}
\toprule
Benchmark & \# Instances & Variables & Constraints \\
\midrule
TSP & 54 & $15821.65 \pm 15593.23$ & $15931.00 \pm 15655.59$ \\
MCND & 81 & $24991.89 \pm 17962.83$ & $2485.22 \pm 1330.64$ \\
CWLP & 36 & $34398.00 \pm 47121.26$ & $420.22 \pm 487.64$ \\
JSSP & 80 & $21382.25 \pm 31271.97$ & $42046.88 \pm 62013.04$ \\
PDPTW & 165 & $11090.00 \pm 0.00$ & $4565.00 \pm 0.00$ \\
IMO6 & 34 & $20759.50 \pm 22078.88$ & $7524.50 \pm 7842.23$ \\
SHUC & 56 & $276709.71 \pm 147182.35$ & $369301.00 \pm 194012.45$ \\
\bottomrule
\end{tabular}
\end{table}

Table~\ref{tab:dev_set_scale} reports the corresponding per-instance scale for the synthetic sources used to form \(D_{imp}\) and \(D_{adm}\). For a given problem, \(D_{imp}\) and \(D_{adm}\) are drawn from the same generator distribution, so their per-instance scale distribution is the same. The default difference is only cardinality.

\begin{table}[ht]
\centering
\footnotesize
\caption{Synthetic development source scale for \(D_{imp}\) and \(D_{adm}\). Variable and constraint counts are reported as mean \(\pm\) population standard deviation over the size pool of each generator.}
\label{tab:dev_set_scale}
\setlength{\tabcolsep}{5pt}
\begin{tabular}{@{}lcc@{}}
\toprule
Synthetic source for \(D_{imp}\) / \(D_{adm}\) & Variables & Constraints \\
\midrule
TSP synthetic & $16737.80 \pm 15413.50$ & $16852.61 \pm 15473.23$ \\
MCND synthetic & $24988.19 \pm 17868.66$ & $2485.22 \pm 1322.35$ \\
CWLP synthetic & $19690.59 \pm 31363.09$ & $420.22 \pm 449.19$ \\
JSSP synthetic & $21382.25 \pm 31075.91$ & $42046.88 \pm 61624.24$ \\
PDPTW synthetic & $8645.00 \pm 1038.60$ & $3627.50 \pm 548.92$ \\
IMO6 synthetic & $41740.75 \pm 44521.34$ & $14863.25 \pm 15606.55$ \\
SHUC synthetic & $48944.00 \pm 30308.70$ & $65436.33 \pm 40031.25$ \\
\bottomrule
\end{tabular}
\end{table}

\subsection{Traveling Salesman Problem}
\label{app:tsp}

\paragraph{Compact MILP formulation for TSP.}
{\scriptsize
\begin{equation}
\left\{
\begin{aligned}
\min\;&\sum_{(i,j)\in A} c_{ij}\,x_{ij}\\[3pt]
\text{s.t.}\;&\sum_{j\in V\setminus\{i\}} x_{ij}=1               &&\forall i\in V\\
            &\sum_{i\in V\setminus\{j\}} x_{ij}=1               &&\forall j\in V\\
            &u_i-u_j+n\,x_{ij}\le n-1                           &&\forall i,j\in V\setminus\{1\},\;i\neq j\\
            &x_{ij}\in\{0,1\}                                   &&\forall(i,j)\in A\\
            &1\le u_i\le n                                      &&\forall i\in V,\quad u_1=1
\end{aligned}
\,\right.\tag{MTZ}\label{eq:tsp_mtz_compact}
\end{equation}
}

\paragraph{Notation glossary.}
\begin{itemize}[itemsep=2pt,leftmargin=*]
  \item $V$ (set): cities to be visited (index $i,j$).
  \item $A\subseteq V\times V$ (set): directed arcs $(i,j)$ that can be used.
  \item $c_{ij}$ (parameter): travel cost from city $i$ to city $j$.
  \item $n=|V|$ (parameter): number of cities.
  \item $x_{ij}\in\{0,1\}$ (variable): equals 1 iff the tour goes directly from city $i$ to city $j$.
  \item $u_i\in[1,n]$ (variable): MTZ position of city $i$ in the tour. $u_1$ is fixed to 1 to anchor the numbering.
\end{itemize}

The acceleration inequality family in Eq.~\eqref{eq:evocut_ec} combines short-cycle elimination, a lifted Desrochers--Laporte variant of MTZ ordering, and depot-anchored envelope bounds on the MTZ position variables. This acceleration inequality family had the highest inequality impact score selected by \textsc{SAGE} under our gap-reduction metric.

\paragraph{Selected {\mdseries\textsc{SAGE}} acceleration inequality family.}
{\scriptsize
\begin{equation}
\left\{
\begin{aligned}
x_{ij} + x_{ji}
    &\le 1
      &&\forall i,j\in V,\; i<j\\[2pt]
x_{1i}+x_{i1}+x_{1j}+x_{j1}+x_{ij}+x_{ji}
    &\le 2
      &&\forall i,j\in V\setminus\{1\},\; i<j\\[2pt]
u_i - u_j + (n-1)\,x_{ij} + (n-3)\,x_{ji}
    &\le n-2
      &&\forall i,j\in V\setminus\{1\},\;i\neq j\\[2pt]
u_i
    &\ge 3 - x_{1i} + (n-3)\,x_{i1}
      &&\forall i\in V\setminus\{1\}\\[2pt]
u_i
    &\le (n-1) + x_{i1} - (n-3)\,x_{1i}
      &&\forall i\in V\setminus\{1\}
\end{aligned}
\,\right.\tag{SAGE-TSP}\label{eq:evocut_ec}
\end{equation}
}

\paragraph{Benchmark sizes.}
We considered $54$ \textsc{TSPLIB} instances with $n<250$ cities.

\subsection{Multi-Commodity Network Design (MCND)}
\label{app:mcnd}
The MCND problem involves selecting a set of network links and assigning multiple flow demands (commodities) at minimal cost. Each commodity must be routed from its source to destination without exceeding link capacities, and activating a link incurs a fixed cost. The MCND problem is an NP-hard combinatorial problem~\cite{gendron1999multicommodity} that can be formulated as an MILP. We used a public set of MCND instances (group \texttt{R} from the CommaLAB dataset)~\cite{commalab2021multicommodity}, which provides a range of network sizes, cost structures, and capacity settings. We used the MILP formulation in Eq.~\eqref{eq:mcnd_milp}.

\paragraph{Compact MILP formulation for MCND.}

{\scriptsize
\begin{equation}
\left\{
\begin{aligned}
\min\;&
      \sum_{(i,j)\in A}\sum_{k\in K} c_{ij}\,x_{ijk}
      +\sum_{(i,j)\in A} f_{ij}\,y_{ij} \\[6pt]
\text{s.t.}\;
&\sum_{j:(i,j)\in A} x_{ijk}=d_k
   &&\forall k\in K,\; i=O_k \\[4pt]
&\sum_{j:(j,i)\in A} x_{jik}=d_k
   &&\forall k\in K,\; i=D_k \\[4pt]
&\sum_{j:(i,j)\in A} x_{ijk}-\sum_{j:(j,i)\in A} x_{jik}=0
   &&\forall k\in K,\; i\in N\setminus\{O_k,D_k\} \\[6pt]
&\sum_{k\in K} x_{ijk}\le u_{ij}\,y_{ij}
   &&\forall(i,j)\in A \\[4pt]
&x_{ijk}\ge 0
   &&\forall(i,j)\in A,\;k\in K \\[2pt]
&y_{ij}\in\{0,1\}
   &&\forall(i,j)\in A
\end{aligned}
\,\right.\tag{MCND}\label{eq:mcnd_milp}
\end{equation}
}

\paragraph{Notation glossary.}
\begin{itemize}[itemsep=2pt,leftmargin=*]
  \item $N$ (set): all network nodes (index $i$).
  \item $A\subseteq N\times N$ (set): candidate directed arcs $(i,j)$.
  \item $K$ (set): commodities to be routed (index $k$).
  \item $O_k$ / $D_k$ (nodes): origin and destination of commodity $k$.
  \item $d_k$ (parameter): demand (quantity) of commodity $k$ that must be shipped from $O_k$ to $D_k$.
  \item $c_{ij}$ (parameter): unit transportation cost on arc $(i,j)$.
  \item $f_{ij}$ (parameter): fixed cost to activate arc $(i,j)$.
  \item $u_{ij}$ (parameter): capacity of arc $(i,j)$.
  \item $x_{ijk}\ge 0$ (variable): flow of commodity $k$ on arc $(i,j)$.
  \item $y_{ij}\in\{0,1\}$ (variable): equals 1 iff arc $(i,j)$ is activated.
\end{itemize}

\paragraph{Selected {\mdseries\textsc{SAGE}} acceleration inequality family.}
Let $\mathcal{I}_{\mathrm{MCND}}$ be the finite set of generated pairs $(S,B)$ used in the static augmentation. Each pair has $S\subseteq N$ and $B\subseteq K(S)$. The set $S$ ranges over concentric breadth-first search zones expanded from commodity origins. The bundles $B$ are sampled small bundles in $K(S)$.
{\scriptsize
\begin{equation}
\left.
\begin{aligned}
\sum_{(i,j)\in \delta^{+}(S)} \min\{u_{ij},\,D_B\}\,y_{ij}
&\;\ge\; D_B
&&\forall (S,B)\in\mathcal{I}_{\mathrm{MCND}} \\[2pt]
\sum_{(i,j)\in \delta^{+}(S)} y_{ij}
&\;\ge\; q_{S,B}
&&\forall (S,B)\in\mathcal{I}_{\mathrm{MCND}}
\end{aligned}
\,\right.\tag{SAGE-MCND}\label{eq:evocut_mcnd}
\end{equation}
}
Here, for each generated pair $(S,B)\in\mathcal{I}_{\mathrm{MCND}}$, $\delta^{+}(S):=\{(i,j)\in A : i\in S,\; j\notin S\}$ denotes the arcs leaving $S$, and $K(S):=\{k\in K : O_k\in S,\; D_k\notin S\}$ are the commodities that must cross this cut. For a generated bundle $B$, let $D_B:=\sum_{k\in B} d_k$. The first row in Eq.~\eqref{eq:evocut_mcnd} is a knapsack-cover variant of the standard graph cut-set capacity inequality. The truncation $\min\{u_{ij},D_B\}$ prevents the LP relaxation from satisfying the capacity constraint using tiny fractional $y_{ij}$ on very high-capacity arcs. The second row is a cardinality companion, where $q_{S,B}$ is the smallest integer such that the sum of the $q_{S,B}$ largest capacities in $\delta^{+}(S)$ reaches $D_B$. These rows have the same form as the full cut-set family over all $S\subseteq N$ and $B\subseteq K(S)$. The MILP receives only the finite generated subset $\mathcal{I}_{\mathrm{MCND}}$.

\paragraph{Benchmark sizes.}

We used the group~\texttt{R} instances r10 to r18 from the CommaLAB dataset ($81$ instances), which cover diverse network topologies and demand settings. Each instance is identified as \texttt{r$x.y$}, where $x$ determines the network size and number of commodities, and $y$ encodes the fixed cost and capacity regime. The following table summarizes the key characteristics of the $x$ configurations used in our benchmark dataset.

\begin{center}
\begin{tabular}{@{}cccc@{}}
\toprule
$x$ & \# Nodes & \# Arcs & \# Commodities \\
\midrule
10-12 & 20 & 120     & 40-200 \\
13-15 & 20 & 220     & 40-200 \\
16-18 & 20 & 314-318 & 40-200 \\
\bottomrule
\end{tabular}
\end{center}

The second index $y$ in \texttt{r$x.y$} controls the fixed cost and capacity setting. Higher fixed costs increase the relative importance of arc selection costs, while higher capacities relax arc flow constraints.

\subsection{Capacitated Warehouse Location Problem (CWLP)}
\label{app:cwlp}
The CWLP is an NP-hard combinatorial optimization problem, formulated as an MILP~\cite{beasley1988algorithm}. It involves selecting a subset of candidate warehouse locations to open and assigning each customer to one open warehouse. The problem has capacity constraints at each facility, with the goal of minimizing total fixed opening and transportation costs. We used a public set~\cite{beasley1990or} of CWLP instances. For our experiments, we use the formulation in Eq.~\eqref{eq:cwlp_milp}.

\paragraph{Compact MILP formulation.}
{\scriptsize
\begin{equation}
\left\{
\begin{aligned}
\min \; & \sum_{j\in J} f_j\,y_{j} \;+\; \sum_{i\in I}\sum_{j\in J} c_{ij}\,x_{ij} \\[4pt]
\text{s.t. } &
\sum_{j\in J} x_{ij} \;=\; 1 &&\forall i\in I \quad\text{(each customer is assigned)} \\[4pt]
&\sum_{i\in I} d_i\,x_{ij} \;\le\; u_j\,y_{j} &&\forall j\in J \quad\text{(capacity)}\\[4pt]
&x_{ij} \in \{0,1\} &&\forall i\in I,\, j\in J\\
&y_{j} \in \{0,1\} &&\forall j\in J
\end{aligned}
\right.
\tag{CWLP}\label{eq:cwlp_milp}
\end{equation}
}
\paragraph{Glossary of notation.}
\begin{itemize}[itemsep=2pt,leftmargin=*]
  \item $I$ (index $i$): set of customers.
  \item $J$ (index $j$): set of candidate warehouses.
  \item $d_i$: demand of customer~$i$.
  \item $u_j$: capacity of warehouse~$j$.
  \item $f_j$: fixed cost to open warehouse~$j$.
  \item $c_{ij}$: total transportation cost incurred if all of customer~$i$'s demand is served by warehouse~$j$ (so $c_{ij}$ already accounts for $d_i$).
  \item $y_j\in\{0,1\}$: decision variable that equals 1 iff warehouse~$j$ is opened (0 otherwise).
  \item $x_{ij}\in\{0,1\}$: decision variable that equals 1 iff customer~$i$ is assigned to warehouse~$j$ (0 otherwise).
\end{itemize}

\paragraph{Selected {\mdseries\textsc{SAGE}} acceleration inequality family.}
The selected CWLP acceleration inequality family found by \textsc{SAGE} is a hybrid of global bin-packing bounds and local lifted capacity covers. It combines (i) global slot count lower bounds (at multiple demand thresholds) and (ii) per warehouse clique/cover inequalities that limit how many large customers can be assigned to a single facility, together with strong links between assignment and opening.

{\scriptsize
\begin{equation}
\left\{
\begin{aligned}
\sum_{j\in J} u_j\,y_j
    &\ge \sum_{i\in I} d_i \\[2pt]
\sum_{j\in J} \Bigl\lfloor \frac{u_j}{v}\Bigr\rfloor\,y_j
    &\ge \bigl|\{i\in I : d_i\ge v\}\bigr|
    &&\forall v\in \mathcal{V} \\[2pt]
x_{ij}
    &\le y_j
    &&\forall i\in I,\;\forall j\in J \\[2pt]
x_{ij}
    &\le 0
    &&\forall i\in I,\;\forall j\in J\ \text{with } d_i>u_j \\[2pt]
\sum_{i\in C_j} x_{ij}
    &\le y_j
    &&\forall j\in J \\[2pt]
\sum_{i\in I : d_i>u_j/3} x_{ij}
    &\le 2\,y_j
    &&\forall j\in J
\end{aligned}
\right.\tag{SAGE-CWLP}
\end{equation}
}
Here, $\mathcal{V}\subseteq \{d_i:i\in I\}$ is a small set of demand thresholds (we use up to the 8 largest distinct demand values in each instance) to instantiate the global slot count bounds. For the local clique inequality, $C_j\subseteq I$ is a lifted conflict set constructed by starting from $\{i\in I: d_i>u_j/2\}$ and sequentially adding smaller customers that still conflict with every customer already in the set (so $d_i+d_{i'}>u_j$ for all distinct $i,i'\in C_j$).

\paragraph{Benchmark sizes.}
We use the CWLP instances from the OR-Library benchmark. Eight instances have 16 facility locations and 50 customers, eight have 25 facility locations and 50 customers, eight have 50 facility locations and 50 customers, and twelve have 100 facility locations and 1,000 customers.
The corresponding baseline models before presolve have 816, 1,275, 2,550, and 100,100 variables and 66, 75, 100, and 1,100 constraints, in the same order. These counts exclude the selected \textsc{SAGE} family.

\subsection{Job Shop Scheduling Problem (JSSP)}
\label{app:jssp}
The JSSP is a classic NP-hard combinatorial optimization problem~\cite{garey1976complexity} and admits a MILP formulation~\cite{manne1960job}. It involves scheduling a set of jobs on multiple machines, where each job has a sequence of operations that must be processed in a specified order on specified machines. The objective is to minimize the makespan (the completion time of the last operation), while ensuring that each machine handles at most one operation at a time. We use the formulation in Eq.~\eqref{eq:jssp_milp} and test \textsc{SAGE} on the widely used Taillard benchmark~\cite{taillard1993benchmarks}.

\paragraph{Compact MILP formulation.}
{\scriptsize
\begin{equation}
\left\{
\begin{aligned}
\min\;& C_{\max} \\[4pt]
\text{s.t.}\quad
& S_{j,k+1} \ge S_{j,k} + p_{j,k}
      &&\forall j,\; k = 1,\dots,n_{\text{mach}}-1 \\[6pt]
& S_{j_1,k_1} + p_{j_1,k_1} \le \\
  &\qquad S_{j_2,k_2} + M(1 - y_{j_1,k_1,j_2,k_2})
      &&\forall\, \bigl((j_1,k_1),(j_2,k_2)\bigr)\in\mathcal{E} \\[6pt]
& S_{j_2,k_2} + p_{j_2,k_2} \le \\
  &\qquad S_{j_1,k_1} + M\,y_{j_1,k_1,j_2,k_2}
      &&\forall\, \bigl((j_1,k_1),(j_2,k_2)\bigr)\in\mathcal{E} \\[6pt]
& C_{\max} \ge S_{j,n_{\text{mach}}} + p_{j,n_{\text{mach}}}
      &&\forall j \\[6pt]
& S_{j,k} \ge 0
      &&\forall j,k \\[2pt]
& y_{j_1,k_1,j_2,k_2}\in\{0,1\}
      &&\forall\, \bigl((j_1,k_1),(j_2,k_2)\bigr)\in\mathcal{E}
\end{aligned}
\right.\tag{JSSP}\label{eq:jssp_milp}
\end{equation}
}
\paragraph{Variable and set glossary.}
\begin{itemize}[itemsep=2pt,leftmargin=*]
  \item $n_{\text{mach}}$ (parameter): number of machines (and operations per job), with $k\in\{1,\dots,n_{\text{mach}}\}$.
  \item $m\in\{1,\dots,n_{\text{mach}}\}$ (index): machine index.
  \item $O$ (set): set of all operations $(j,k)$.
  \item $O_m$ (set): set of operations processed on machine~$m$.
  \item $S_{j,k}\in\mathbb{R}_{\ge 0}$ (variable): start time of the $k$-th operation of job~$j$.
  \item $p_{j,k}$ (parameter): processing time of operation $(j,k)$.
  \item $\mathcal{E}$ (set): ordered pairs $\bigl((j_1,k_1),(j_2,k_2)\bigr)$ of distinct operations that require the same machine. For every such pair, exactly one of the two order enforcing inequalities becomes active.
  \item $y_{j_1,k_1,j_2,k_2}\in\{0,1\}$: binary variable that equals 1 if operation $(j_1,k_1)$ is scheduled before $(j_2,k_2)$ on their shared machine, 0 otherwise.
  \item $C_{\max}\in\mathbb{R}_{\ge 0}$: continuous makespan (completion time of the last operation). The objective minimizes this value.
  \item $M=\sum_{j,k}p_{j,k}$: the total processing time, which is a valid schedule horizon and deactivates the unselected sequencing inequality.
\end{itemize}

\paragraph{Selected {\mdseries\textsc{SAGE}} acceleration inequality family.}
{\scriptsize
\begin{equation}
\left.
\begin{aligned}
C_{\max}
&\ge \frac{1}{n_{\text{mach}}} \sum_{(j,k)\in O} p_{j,k} \\[2pt]
C_{\max}
&\ge \texttt{CP}_m
&&\forall m\in\{1,\dots,n_{\text{mach}}\} \\[2pt]
C_{\max}
&\ge \sum_{k=1}^{n_{\text{mach}}} p_{j,k}
&&\forall j
\end{aligned}
\,\right.\tag{SAGE-JSSP}
\end{equation}
}
This hybrid family combines three lower bounds on the makespan:

\begin{itemize}[itemsep=2pt,leftmargin=*]
\item \textbf{Average load bound}: $\frac{1}{n_{\text{mach}}} \sum_{(j,k)\in O} p_{j,k}$ represents the average total processing time per machine, so the makespan is at least the average workload.

\item \textbf{Machine-level critical path bound} ($\texttt{CP}_m$): For each machine $m$, let $O_m$ be the set of operations processed on $m$. For any operation $(j,k)$, define the job prefix (head) and job suffix (tail)
\[
h_{j,k} := \sum_{t<k} p_{j,t}, \qquad
t_{j,k} := \sum_{t>k} p_{j,t}.
\]
Then we take
\[
\texttt{CP}_m \;:=\; \sum_{(j,k)\in O_m} p_{j,k}
\;+\; \min_{(j,k)\in O_m} h_{j,k}
\;+\; \min_{(j,k)\in O_m} t_{j,k},
\]
which lower bounds the completion time of any schedule by accounting for the total load on $m$ plus the earliest possible release (head) of the first operation on $m$ and the smallest remaining tail after the last operation on $m$.

\item \textbf{Longest job bound}: $\max_{j}\sum_{k} p_{j,k}$ lower bounds the makespan by the total processing time of the longest job chain.
\end{itemize}

These three bounds together use machine workloads and job chain precedence (heads/tails). The hybrid acceleration inequality family was selected by \textsc{SAGE} based on optimality gap reduction.

\noindent Table~\ref{tab:selected_family_reporting} counts the family as instantiated on each test instance. The instantiated form strengthens the machine bound with two continuous variables in $[0,1]$ for every operation, which indicate whether the operation is first or last on its machine. On each machine whose load is at least 85\% of the largest machine load, it adds the rows that link these indicators to the sequencing variables and the start times, and a bound that adds the head of the first operation and the tail of the last operation to the machine load. These variables and rows give the JSSP counts in the table.

\paragraph{Benchmark sizes.}
We evaluated \textsc{SAGE} on the Taillard JSSP benchmark (\texttt{ta01} to \texttt{ta80}, 80 instances). This includes instances with the following job and machine counts:
\begin{itemize}[itemsep=2pt,leftmargin=*]
\item 15 jobs $\times$ 15 machines (e.g., ta01-ta10),
\item 20 jobs $\times$ 15 machines (e.g., ta11-ta20),
\item 20 jobs $\times$ 20 machines (e.g., ta21-ta30),
\item 30 jobs $\times$ 15 machines (e.g., ta31-ta40),
\item 30 jobs $\times$ 20 machines (e.g., ta41-ta50),
\item 50 jobs $\times$ 15 machines (e.g., ta51-ta60),
\item 50 jobs $\times$ 20 machines (e.g., ta61-ta70),
\item 100 jobs $\times$ 20 machines (e.g., ta71-ta80).
\end{itemize}
These instances are widely used in the literature to assess scheduling algorithms on problems with different sizes and difficulty levels.

\subsection{Pickup and Delivery Problem with Time Windows (PDPTW)}
\label{app:pdptw}
The PDPTW is a vehicle routing problem in which each request consists of a pickup location and a delivery location that must be served in precedence order, subject to time window and vehicle capacity constraints. The objective is to route vehicles to serve all requests while minimizing total travel cost. PDPTW is NP-hard and admits MILP formulations. It comes from practice and has many constraints, and some benchmark instances used in our suite are included in MIPLIB~\cite{koch2011miplib}. We use the truckload PDPTW instances derived from a real-world drayage case at the Port of Rotterdam~\cite{srour2010miplib}. We obtained the compressed MPS files from the MIPLIB 2010 contribution page~\cite{srour2010miplib}. We convert them into our parameterized model by reading each MPS file in Gurobi and recovering \((K,N,d_{0i}^k,d_{ij},d_{iH}^k,v^k,\tau_i^{-},\tau_i^{+})\) from the instance coefficients.

\paragraph{Compact MILP formulation.}
We use the MILP formulation of~\cite{srour2010miplib}, which models trucks and jobs as nodes in a complete directed graph and permits explicit job rejection via self-loops \(x_{K+i,K+i}=1\) with penalty \(d_{ii}\):
{\scriptsize
\begin{equation}
\left\{
\begin{aligned}
\min\;&\sum_{k=1}^{K}\sum_{i=1}^{N} d_{0i}^k\,x_{k,K+i}
      +\sum_{i=1}^{N}\sum_{j=1}^{N} d_{ij}\,x_{K+i,K+j}
      +\sum_{i=1}^{N}\sum_{k=1}^{K} d_{iH}^k\,x_{K+i,k}\\[4pt]
\text{s.t.}\;&\sum_{v=1}^{K+N} x_{uv} = 1
      &&\forall u=1,\dots,K+N \\[4pt]
&\sum_{v=1}^{K+N} x_{vu} = 1
      &&\forall u=1,\dots,K+N \\[4pt]
&\delta_i - \sum_{k=1}^{K} (d_{0i}^k + v^k)\,x_{k,K+i} \ge 0
      &&\forall i=1,\dots,N \\[4pt]
&\delta_j - \delta_i - M\,x_{K+i,K+j} + (d_{ii}+d_{ij})\,x_{K+i,K+i} \ge d_{ii}+d_{ij}-M
      &&\forall i,j=1,\dots,N \\[4pt]
&\tau_i^{-} \le \delta_i \le \tau_i^{+}
      &&\forall i=1,\dots,N \\[4pt]
&\delta_i \in \mathbb{R}_{\ge 0}
      &&\forall i=1,\dots,N \\[4pt]
&x_{uv} \in \{0,1\}
      &&\forall u,v=1,\dots,K+N
\end{aligned}
\right.\tag{PDPTW}\label{eq:pdptw_milp}
\end{equation}
}

\paragraph{Notation glossary.}
\begin{itemize}[itemsep=2pt,leftmargin=*]
  \item $K$ (parameter): number of trucks/vehicles in the fleet (truck nodes $k\in\{1,\dots,K\}$).
  \item $N$ (parameter): number of pickup and delivery demands/jobs (job nodes $K+i$ for $i\in\{1,\dots,N\}$).
  \item $u,v\in\{1,\dots,K+N\}$ (indices): nodes in the complete directed graph.
  \item $d_{ij}$ (parameter): travel time from demand $i$'s return terminal to demand $j$'s pickup terminal. For $i=j$, $d_{ii}$ is the loaded time of job $i$ and is used as the rejection penalty.
  \item $d_{0i}^k$ (parameter): travel time from the home terminal of truck $k$ to demand $i$'s pickup terminal.
  \item $d_{iH}^k$ (parameter): travel time from demand $i$'s return terminal to the home terminal of truck $k$.
  \item $v^k$ (parameter): time at which truck $k$ becomes available.
  \item $\tau_i^{-},\tau_i^{+}$ (parameters): earliest/latest permissible arrival times at demand $i$'s pickup terminal.
  \item $M$ (parameter): big-$M$ constant that deactivates the sequencing inequality on an unselected arc. All source MPS files~\cite{srour2010miplib} use $M=60{,}000$, which we retain in both formulations.
  \item $x_{uv}\in\{0,1\}$ (variable): equals 1 iff arc $(u,v)$ is selected in the routing cycle cover.
  \item $\delta_i\in\mathbb{R}_{\ge 0}$ (variable): arrival time at demand $i$'s pickup terminal.
\end{itemize}

\paragraph{Selected {\mdseries\textsc{SAGE}} acceleration inequality family.}
The selected PDPTW acceleration inequality family adds four time-window inequalities. Let
\[
\mathrm{EST}_i := \max\{\tau_i^{-}, \min_{k} (v^k + d_{0i}^k)\}
\]
and
\[
\mathcal{A}^{-} := \{(i,j): \mathrm{EST}_i + d_{ii} + d_{ij} > \tau_j^{+}\}.
\]
We then add the following four inequalities:
{\scriptsize
\begin{equation}
\left\{
\begin{aligned}
&x_{K+i,K+j} \le 0 && \forall (i,j)\in \mathcal{A}^{-} \\
&x_{K+i,K+j} + x_{K+j,K+i} + x_{K+i,K+i} \le 1 && \forall (i,j)\in \mathcal{F}_2 \\
&\sum_{i\in C} x_{K+i,K+i} \ge |C|-|K(C)| && \forall C\in \mathcal{C} \\
&x_{K+i,K+k} + x_{K+k,K+j} + x_{K+k,K+k} \le 1 && \forall (i,k,j)\in \mathcal{Q}
\end{aligned}
\right.\tag{SAGE-PDPTW}
\end{equation}
}
Here, \(\mathcal{F}_2\) is the set of mutually feasible pairs:
\[
\mathcal{F}_2 := \{(i,j): i\neq j,\; (i,j)\notin \mathcal{A}^{-},\; (j,i)\notin \mathcal{A}^{-}\}.
\]
\(\mathcal{C}\) is the set of cliques with \(|C|>|K(C)|\) in the conflict graph that connects pairs with both directions infeasible, where
\[
K(C):=\{k\in\{1,\dots,K\}:\exists i\in C\ \text{s.t.}\ v^k+d_{0i}^k\le\tau_i^{+}\}
\]
is the set of trucks that can reach at least one job in $C$ by its latest pickup time. Finally,
\[
\mathcal{Q}:=\{(i,k,j): (i,k)\notin \mathcal{A}^{-},\; (k,j)\notin \mathcal{A}^{-},\; \max\{\mathrm{EST}_k,\mathrm{EST}_i + d_{ii}+d_{ik}\}+d_{kk}+d_{kj} > \tau_j^{+}\}.
\]

\paragraph{Benchmark sizes.}
We use the five MIPLIB truckload PDPTW groups \texttt{R0}, \texttt{R25}, \texttt{R50}, \texttt{R75}, and \texttt{R100}~\cite{srour2010miplib}. Each group contains 33 instances (one per day) for a total of 165 instances. Each instance has $K=40$ trucks and $N=65$ jobs. The groups differ by the fraction of jobs that are released after the start of the day (0\%, 25\%, 50\%, 75\%, 100\%, respectively).

\subsection{International Mathematical Olympiad 2025 Problem~6 (IMO6)}
\label{app:imo6}
This benchmark is based on IMO 2025 Problem~6: given an $N\times N$ grid of unit squares, place axis-aligned rectangular tiles (no overlaps) so that each row and each column contains exactly one uncovered unit square (a hole). The objective is to minimize the number of tiles used~\cite{imo2025international}. We include IMO6 as a newly formulated MILP benchmark (derived from a math olympiad problem), so it is not drawn from standard IP benchmark literature and LLMs are unlikely to have seen established formulation ideas for it. We model the problem by defining fixed-length strips and tracking their vertical continuation over a 2D board. The following IP model matches the Pyomo code used in our experiments.

\paragraph{Compact MILP formulation.}
Let $R=\{1,\dots,N\}$ be the rows, $C=\{1,\dots,N\}$ the columns, and $I=\{(a,b)\in C\times C:\;a\le b\}$ the set of horizontal strips. Binary variables: $h_{ij}$ indicate the hole position, and $x_i^{ab}$ indicates that strip $(a,b)$ is active on row $i$. The variables $s_i^{ab}, \, t_i^{ab} \in \{0, 1\}$ mark the start and end rows of strip $(a,b)$. The objective counts rectangles, because each rectangle contributes exactly one start. The flow equalities ensure each active strip $(a,b)$ has one start and one end, so summation over start variables equals the number of rectangles.
{\scriptsize
\begin{equation}
\left\{
\begin{aligned}
\min\;&\sum_{i\in R}\;\sum_{(a,b)\in I} s_i^{ab} \\[4pt]
\text{s.t.}\;&\sum_{j\in C} h_{ij} = 1
 &&\forall i\in R\\[-2pt]
& &&\text{(one hole per row)}\\[4pt]
&\sum_{i\in R} h_{ij} = 1
 &&\forall j\in C\\[-2pt]
& &&\text{(one hole per column)}\\[4pt]
&\sum_{\substack{(a,b)\in I\\ a\le j\le b}} x_i^{ab} + h_{ij} = 1
 &&\forall i\in R,\;\forall j\in C\\[-2pt]
& &&\text{(each cell is covered by exactly one strip or a hole)}\\[4pt]
&x_{1}^{ab} - s_{1}^{ab} = 0
 &&\forall (a,b)\in I\\[-2pt]
& &&\text{(top flow)}\\[4pt]
&x_{i}^{ab} - x_{i-1}^{ab} - s_{i}^{ab} + t_{i-1}^{ab} = 0
 &&\forall i=2,\dots,N,\;\forall(a,b)\in I\\[-2pt]
& &&\text{(mid flow)}\\[4pt]
&x_{N}^{ab} - t_{N}^{ab} = 0
 &&\forall (a,b)\in I\\[-2pt]
& &&\text{(bottom flow)}\\[4pt]
&h_{ij}\in\{0,1\}
 &&\forall i\in R,\;\forall j\in C\\
&x_{i}^{ab},\,s_{i}^{ab},\,t_{i}^{ab}\in\{0,1\}
 &&\forall i\in R,\;\forall(a,b)\in I
\end{aligned}
\,\right.\tag{RT-2DFlow}\label{eq:rect_2dflow_compact}
\end{equation}
}

\paragraph{Notation glossary.}
\begin{itemize}[itemsep=2pt,leftmargin=*]
  \item $R=\{1,\dots,N\}$, $C=\{1,\dots,N\}$: row and column index sets ($i\in R$, $j\in C$).
  \item $I=\{(a,b)\in C^2:\ a\le b\}$: all contiguous column intervals (index $(a,b)$).
  \item $h_{ij}\in\{0,1\}$: $=1$ iff $(i,j)$ is the unique hole in row $i$ and in column $j$.
  \item $x_i^{ab}\in\{0,1\}$: $=1$ iff on row $i$ the columns $a,a{+}1,\dots,b$ are covered by the same tile and strip $(a,b)$ is active.
  \item $s_i^{ab},t_i^{ab}\in\{0,1\}$: start/end flags of the vertical strip for interval $(a,b)$ at row $i$. The number of rectangles equals $\sum_{i,(a,b)} s_i^{ab}$.
\end{itemize}

\paragraph{Selected {\mdseries\textsc{SAGE}} acceleration inequality family.}
The selected IMO6 acceleration inequality family uses both uniqueness conditions on holes. It forbids an active strip from containing the row hole, tightens row activity to the number of connected covered components in the row, and enforces mandatory new starts when a hole in the previous row is vacated. The last two inequalities also restrict the new start to the same side of the new hole as the covered column.
{\scriptsize
\begin{equation}
\left\{
\begin{aligned}
x_i^{ab}+\sum_{\ell=a}^{b}h_{i\ell}
&\le 1
&&\forall i\in R,\ \forall (a,b)\in I\\[4pt]
\sum_{(a,b)\in I} x_i^{ab}
&\ge 2-h_{i1}-h_{iN}
&&\forall i\in R\\[4pt]
\sum_{(a,b)\in I} s_i^{ab}
&\ge 1
&&\forall i\in\{2,\dots,N\}\\[4pt]
\sum_{\substack{(a,b)\in I\\ a\le j\le b}} s_i^{ab}
&\ge h_{i-1,j}
&&\forall i\in\{2,\dots,N\},\ \forall j\in C\\[4pt]
\sum_{\substack{(a,b)\in I\\ a\le j\le b<k}} s_i^{ab}
&\ge \sum_{\substack{(a,b)\in I\\ a\le j\\ b\ge k}} x_{i-1}^{ab}+h_{ik}-1
&&\forall i\in\{2,\dots,N\},\ \forall j,k\in C,\ j<k\\[4pt]
\sum_{\substack{(a,b)\in I\\ k<a\le j\le b}} s_i^{ab}
&\ge \sum_{\substack{(a,b)\in I\\ a\le k\\ b\ge j}} x_{i-1}^{ab}+h_{ik}-1
&&\forall i\in\{2,\dots,N\},\ \forall j,k\in C,\ j>k
\end{aligned}
\,\right.\tag{SAGE-RT-Disrupt}\label{eq:evocut_rect_hybrid}
\end{equation}
}
\noindent The first inequality prevents any active strip from crossing the row's unique hole. The second inequality counts row components: if the hole is in an interior column, the covered cells split into two disconnected intervals. Otherwise they form one interval. The third inequality follows from column uniqueness: the column that was a hole in row $i{-}1$ must be covered in row $i$, so at least one rectangle starts in every row after the first. The fourth tightens the vacated-column inequality to require a start covering the exact column vacated by the previous row's hole. The final two inequalities handle intervals broken by the new hole: if the new hole at $k$ breaks an interval from row $i{-}1$, the new start covering $j$ cannot cross $k$. Together with the top-flow equality, these inequalities give the LP relaxation the objective lower bound $\sum_{i,(a,b)}s_i^{ab}\ge N$.

\paragraph{Benchmark sizes.}
We generate one instance per $N\in\{3,\dots,46\}$ (44 instances total). We use 10 instances as the development pool for \textsc{SAGE}. In this pool, $D_{imp}$ contains all 10 instances and $D_{adm}$ is a 2-instance subset used for admissibility screening. The instances with $N\in\{3,\dots,36\}$ form the test set.

\subsection{Sub-Hour Unit Commitment (SHUC)}
\label{app:shuc}
The SHUC benchmark is a unit commitment problem with a horizon under one hour that decides generator commitment (on/off) and dispatch levels to meet demand and reserve. SHUC is a problem motivated by practice with many constraints. Some benchmark instances used in our suite are included in MIPLIB~\cite{koch2011miplib}. We use the PGLib-UC v19.08 JavaScript Object Notation (JSON) benchmark~\cite{powergridlib2019pglib} and follow its reference formulation with piecewise production costs, startup categories, ramping limits, and minimum up and down times~\cite{knueven2020mixed}. For source attribution, we use the recommended documents for the \texttt{ca} cases~\citep{knueven2020mixed}, the \texttt{ferc} cases~\citep{knueven2020mixed,krall2012rto}, and the \texttt{rts\_gmlc} cases~\citep{barrows2020ieee}.

\paragraph{Compact MILP formulation.}
{\scriptsize
\begin{equation}
\left\{
\begin{aligned}
\min\;&\sum_{g\in G}\sum_{t\in \mathcal{T}} \bigl(c_{g,t} + C_{g,1}\,u_{g,t}\bigr)
      + \sum_{g\in G}\sum_{s\in S_g}\sum_{t\in \mathcal{T}} C^{su}_{g,s}\,d_{g,s,t} \\[4pt]
\text{s.t.}\;&\sum_{g\in G} \bigl(p_{g,t} + P^{\min}_g\,u_{g,t}\bigr)
            + \sum_{w\in W} p^w_{w,t} = L_t
      &&\forall t\in \mathcal{T} \\[4pt]
&\sum_{g\in G} r_{g,t} \ge R_t
      &&\forall t\in \mathcal{T} \\[4pt]
&u_{g,t} - u_{g,t-1} = v_{g,t} - w_{g,t}
      &&\forall g\in G,\ t=2,\dots,T \\[4pt]
&\sum_{\tau=t-U_g+1}^{t} v_{g,\tau} \le u_{g,t}
      &&\forall g\in G,\ t\ge U_g \\[4pt]
&\sum_{\tau=t-D_g+1}^{t} w_{g,\tau} \le 1 - u_{g,t}
      &&\forall g\in G,\ t\ge D_g \\[4pt]
&v_{g,t} = \sum_{s\in S_g} d_{g,s,t}
      &&\forall g\in G,\ t\in \mathcal{T} \\[4pt]
&d_{g,s,t} \le \sum_{i=\ell_{g,s}}^{\ell_{g,s+1}-1} w_{g,t-i}
      &&\forall g\in G,\ s<|S_g|,\ t\ge \ell_{g,s+1} \\[4pt]
&u_{g,t} \ge \mathrm{MR}_g
      &&\forall g\in G,\ t\in \mathcal{T} \\[4pt]
&p_{g,t} + r_{g,t}
 \le (P^{\max}_g - P^{\min}_g)\,u_{g,t} - \max(P^{\max}_g - SU_g,0)\, v_{g,t}
      &&\forall g\in G,\ t\in \mathcal{T} \\[4pt]
&p_{g,t} + r_{g,t}
 \le (P^{\max}_g - P^{\min}_g)\,u_{g,t} - \max(P^{\max}_g - SD_g,0)\, w_{g,t+1}
      &&\forall g\in G,\ t<T \\[4pt]
&p_{g,t} + r_{g,t} - p_{g,t-1} \le RU_g
      &&\forall g\in G,\ t=2,\dots,T \\[4pt]
&p_{g,t-1} - p_{g,t} \le RD_g
      &&\forall g\in G,\ t=2,\dots,T \\[4pt]
&p_{g,t} = \sum_{l\in L_g} (P_{g,l} - P_{g,1})\,\lambda_{g,l,t}
      &&\forall g\in G,\ t\in \mathcal{T} \\[4pt]
&c_{g,t} = \sum_{l\in L_g} (C_{g,l} - C_{g,1})\,\lambda_{g,l,t}
      &&\forall g\in G,\ t\in \mathcal{T} \\[4pt]
&u_{g,t} = \sum_{l\in L_g} \lambda_{g,l,t}
      &&\forall g\in G,\ t\in \mathcal{T} \\[4pt]
&P^{w,\min}_{w,t} \le p^w_{w,t} \le P^{w,\max}_{w,t}
      &&\forall w\in W,\ t\in \mathcal{T} \\[4pt]
&u_{g,t}, v_{g,t}, w_{g,t}, d_{g,s,t} \in \{0,1\},\ \lambda_{g,l,t}\in[0,1] \\[-1pt]
&p_{g,t}, r_{g,t}, c_{g,t}, p^w_{w,t} \ge 0
\end{aligned}
\,\right.\tag{SHUC}\label{eq:shuc_milp}
\end{equation}
}
\noindent We also enforce initial on status, initial output, and remaining up or down time at $t=1$ using the dataset fields \texttt{unit\_on\_t0}, \texttt{power\_output\_t0}, \texttt{time\_up\_t0}, and \texttt{time\_down\_t0}, matching the PGLib-UC reference model.

\paragraph{Notation glossary.}
\begin{itemize}[itemsep=2pt,leftmargin=*]
  \item $\mathcal{T}=\{1,\dots,T\}$: time periods (index $t$).
  \item $G$: thermal generators, $W$: renewable generators.
  \item $S_g$: startup categories for generator $g$, with lag $\ell_{g,s}$ and cost $C^{su}_{g,s}$.
  \item $L_g$: piecewise production points $(P_{g,l}, C_{g,l})$, with $C_{g,1}$ the fixed on cost per period.
  \item $L_t$, $R_t$: demand and reserve requirements at time $t$.
  \item $P^{\min}_g$, $P^{\max}_g$: minimum and maximum thermal output, and $P^{w,\min}_{w,t}$, $P^{w,\max}_{w,t}$ are renewable bounds.
  \item $RU_g$, $RD_g$: ramp up and ramp down limits, and $SU_g$, $SD_g$: startup and shutdown ramp limits.
  \item $U_g$, $D_g$: minimum up and minimum down times, and $\mathrm{MR}_g$: must run flag.
  \item $u_{g,t}$: on status, $v_{g,t}$: startup, and $w_{g,t}$: shutdown.
  \item $p_{g,t}$: thermal output above minimum, $r_{g,t}$: spinning reserve, and $p^w_{w,t}$: renewable output.
  \item $d_{g,s,t}$: startup category selection, $\lambda_{g,l,t}$: piecewise weights, and $c_{g,t}$: variable production cost above the base cost.
\end{itemize}

\paragraph{Selected {\mdseries\textsc{SAGE}} acceleration inequality family.}
{\scriptsize
\begin{equation}
\left\{
\begin{aligned}
&b_{g,t} \le v_{g,t}
&&\forall g\in G,\ t<T \\[2pt]
&b_{g,t} \le w_{g,t+1}
&&\forall g\in G,\ t<T \\[2pt]
&b_{g,t} \ge v_{g,t} + w_{g,t+1} - 1
&&\forall g\in G,\ t<T \\[6pt]
&\overline{P}_{g,t} \le P^{\max}_g\,u_{g,t} - \max(P^{\max}_g - SU_g,0)\,v_{g,t}
&&\forall g\in G,\ t\in \mathcal{T} \\[2pt]
&\overline{P}_{g,t} \le P^{\max}_g\,u_{g,t} - \max(P^{\max}_g - SD_g,0)\,w_{g,t+1}
&&\forall g\in G,\ t<T \\[2pt]
&\overline{P}_{g,t} \le P^{\max}_g\,u_{g,t}
 - \max(P^{\max}_g - SU_g,0)\,v_{g,t}
 - \max(P^{\max}_g - SD_g,0)\,w_{g,t+1}
&&\forall g\in G,\ t<T \\[-1pt]
&\qquad + \min\!\bigl(\max(P^{\max}_g - SU_g,0),\max(P^{\max}_g - SD_g,0)\bigr)\,b_{g,t}
&& \\[6pt]
&\overline{P}_{g,t} \le P^{\min}_g\,u_{g,t} + p_{g,t-1} + RU_g + (P^{\max}_g - P^{\min}_g)\,(1-u_{g,t-1})
&&\forall g\in G,\ t=2,\dots,T \\[2pt]
&\overline{P}_{g,t} \le P^{\min}_g\,u_{g,t} + (P^{\max}_g - P^{\min}_g)\,u_{g,t-1} - \max(P^{\max}_g - SU_g,0)\,v_{g,t-1} + RU_g
&&\forall g\in G,\ t=2,\dots,T \\[2pt]
&\overline{P}_{g,t} \le P^{\min}_g\,u_{g,t} + (P^{\max}_g - P^{\min}_g)\,u_{g,t-1} - \max(P^{\max}_g - SD_g,0)\,w_{g,t} + RU_g
&&\forall g\in G,\ t=2,\dots,T \\[6pt]
&L_t - \sum_{w\in W} p^w_{w,t} + R_t \le \sum_{g\in G} \overline{P}_{g,t}
&&\forall t\in \mathcal{T}
\end{aligned}
\,\right.\tag{SAGE-SHUC}\label{eq:evocut_shuc}
\end{equation}
}
This acceleration inequality family requires that demand minus renewable output plus reserve can be met by thermal capacity that is reachable under ramp limits, with derating during startups and shutdowns. We also apply the $t=1$ form of the ramp reachability bounds using the dataset $t0$ fields. Here, $\overline{P}_{g,t}$ and $b_{g,t}$ are auxiliary variables introduced by the selected acceleration inequality family.

\noindent Table~\ref{tab:selected_family_reporting} counts the family as instantiated on each test instance. Beyond the displayed block, the instantiated form adds ramp rows on the actual output that account for the commitment status and the startup and shutdown limits, a lower bound on the number of online units in each period, and cover rows that require the committed, starting, and stopping units to cover the residual demand and reserve. Section~\ref{sec:shuc_case_study} proves the validity of the displayed block.

\paragraph{Benchmark sizes.}
We use all $56$ PGLib-UC v19.08 instances: 20 \texttt{ca}, 24 \texttt{ferc}, and 12 \texttt{rts\_gmlc} cases. Each instance has $T=48$ periods, $73$ to $978$ thermal generators, and $0$ to $81$ renewable generators. We set $|D_{imp}|=10$ and $|D_{adm}|=2$ from our synthetic generator, and use the full 56-instance PGLib-UC benchmark as $D_t$.

\subsection{Reconfigurable Network Topology Design (RNTD)}
\label{app:rntd}

RNTD installs capacity on undirected links and routes commodities through the resulting network. It limits the number of capacity units incident to each node, the number of hops in each route, and the deviation from a reference design. One installed capacity variable is shared by flow in both directions of a link.

Let $P$ be the nodes, $E$ the undirected candidate links, and $K$ the commodities. Commodity $k$ has volume $w_k$, source $s_k$, and destination $t_k$. The set $A_k$ contains the directed arcs that can occur in a route of at most $L$ links. For each link $e=\{i,j\}$, $\Gamma_e$ is its fixed cost, $\chi_{ke}$ is its routing cost, $u_e$ is the capacity of one installed unit, and $C_i$ is the capacity unit limit at node $i$. The reference design is $\hat y$, its deviation budget is $R$, and $y^{\max}$ bounds the units on one link. The variables are commodity flow $x_{kij}\ge0$, installed units $y_e\in\mathbb Z_{\ge0}$, and absolute deviation $q_e\ge0$. Define $b_{ki}=w_k$ at $s_k$, $b_{ki}=-w_k$ at $t_k$, and $b_{ki}=0$ otherwise. The RNTD MILP is
\begin{align}
\min_{x,y,q}\quad
& \sum_{e\in E}\Gamma_e y_e
  +\sum_{k\in K}\sum_{(i,j)\in A_k}\chi_{k,\{i,j\}}x_{kij} \label{eq:rntd_milp}\\
\text{s.t.}\quad
& \sum_{j:(i,j)\in A_k}x_{kij}-\sum_{j:(j,i)\in A_k}x_{kji}=b_{ki}
&& \forall k\in K,\ i\in P, \notag\\
& \sum_{k:(i,j)\in A_k}x_{kij}+\sum_{k:(j,i)\in A_k}x_{kji}
  \le u_{\{i,j\}}y_{\{i,j\}}
&& \forall \{i,j\}\in E, \notag\\
& \sum_{e\in\delta(i)}y_e\le C_i
&& \forall i\in P, \notag\\
& q_e\ge y_e-\hat y_e,\qquad q_e\ge \hat y_e-y_e
&& \forall e\in E, \notag\\
& \sum_{e\in E}q_e\le R,\qquad
  0\le y_{\{i,j\}}\le \min\{y^{\max},C_i,C_j\}
&& \forall \{i,j\}\in E. \notag
\end{align}
All generated instances use $L=2$, so the restricted arc sets enforce the hop limit.

The selected family combines demand across a node cut with the node capacity limits. For $S\subset P$, let $\delta(S)$ be its cut links, $E(S)$ its internal links, $W_S$ the total commodity volume whose source and destination lie on opposite sides, and $B_S=\lceil W_S/\max_{e\in\delta(S)}u_e\rceil$. For every set with $1\le |S|\le2$, the family adds
\begin{align}
\sum_{e\in\delta(S)}y_e &\ge B_S, \label{eq:rntd_family}\\
\sum_{e\in E(S)}y_e
&\le \left\lfloor\frac{\sum_{i\in S}C_i-B_S}{2}\right\rfloor, \notag\\
\sum_{e\in E(P\setminus S)}y_e
&\le \left\lfloor\frac{\sum_{i\in P\setminus S}C_i-B_S}{2}\right\rfloor. \notag
\end{align}

The generator samples nodes in a square and uses the complete undirected candidate graph. It samples ordered source and destination pairs, volumes, distance based link costs, and one of three routing cost profiles. Each volume is between 10 and 60, one capacity unit carries 100, and two sampled hubs define a feasible reference design. Node limits equal the incident reference units plus three, and the deviation budget is sampled between 0.8 and 1.2 times the total reference capacity. We used independent seeds for 12 development instances and 44 reserved test instances. The selected family was fixed before any test solve. The generator and all instances are released with the code.

\section{LLM API Configuration}
\label{app:llm_params}

\subsection{Parameters and token usage}
Table~\ref{tab:llm_params} lists the exact parameters used in every call to the three LLM APIs (\texttt{deepseek-reasoner}, \texttt{google/gemini-3-pro-preview}, and \texttt{gpt-5.1}) throughout our \textsc{SAGE} pipeline, together with approximate token usage.
At the listed prices, the approximate total API cost across the three LLM runs is USD 400.

\begin{table}[ht]
\centering
\footnotesize
\caption{LLM API settings and estimated token costs used for all \textsc{SAGE} experiments. Costs use Jan 2026 OpenRouter list prices. The Gemini 3 Pro API identifier and pricing follow OpenRouter~\cite{openrouter2025google}.}
\label{tab:llm_params}
\setlength{\tabcolsep}{4pt}
\begin{tabular}{lccc}
\toprule
Parameter
    & DeepSeek
    & Gemini 3 Pro
    & GPT-5.1 \\
\midrule
Model                         & \texttt{deepseek-reasoner} & \texttt{google/gemini-3-pro-preview} & \texttt{gpt-5.1} \\
Maximum output tokens         & 10,000 & 10,000 & 10,000 \\
Temperature                   & 1.0 & 1.0 & 1.0 \\
Frequency penalty             & 0.0 & 0.0 & 0.0 \\
Presence penalty              & 0.0 & 0.0 & 0.0 \\
Total input tokens            & $\sim$26 M & $\sim$19 M & $\sim$28 M \\
Total output tokens           & $\sim$13 M & $\sim$14 M & $\sim$17 M \\
Total tokens                  & $\sim$39 M & $\sim$33 M & $\sim$45 M \\
Total cost (Jan 2026 pricing) & $\sim$\$50 & $\sim$\$200 & $\sim$\$150 \\
\bottomrule
\end{tabular}
\end{table}

\section{Inequality Impact Score}
\label{app:impact_score}
This appendix gives the range and the implementation of the inequality impact score defined in Section~\ref{sec:solution_method}. Because $\mathrm{gap}_{\mathrm{aug}}(i)\ge 0$, every term satisfies $d(i)\ge -1$, so $s_{imp}(C)\le 10e\approx 27.2$, with equality only when the family closes the gap on every impact assessment instance. A score above 10 indicates a mean gap reduction, and the mean relative gap reduction $-\bar d$ equals $\ln(s_{imp}(C)/10)$. For example, the best scores of 17.60 and 19.07 in Appendix~\ref{app:evo_control} correspond to mean development gap reductions of 56.5\% and 64.6\%. In the implementation, $\mathrm{gap}_{\mathrm{aug}}(i)$ is the relative gap that Gurobi reports when the work limit is reached. If the augmented run finds no incumbent within the work limit, the solver reports no gap and the instance is left out of $\bar d$.

\end{document}